THÈSE de DOCTORAT DE L'UNIVERSITÉ DE LYON

Opérée au sein de :

**l'Université Claude Bernard Lyon 1**

**Ecole Doctorale** 512
InfoMath

**Spécialité de doctorat :** Informatique
**Discipline :** Informatique

Soutenue publiquement le 15/03/2023, par :

## Miguel Palencia-Olivar

# A Topical Approach to Capturing Customer Insight In Social Media

Devant le jury composé de :

| | |
|---|---|
| **Julyan ARBEL**<br>Chargé de recherche, INRIA | **Rapporteur** |
| **Alexandre AUSSEM**<br>Professeur des Universités, Université Claude Bernard Lyon 1 | **Co-Directeur de thèse** |
| **Stéphane BONNEVAY**<br>Maître de Conférences, Université Claude Bernard Lyon 1 | **Directeur de thèse** |
| **Bruno CANITIA**<br>Directeur R&D, Lizeo IT | **Invité** |
| **Marianne CLAUSEL**<br>Professeure des Universités, Université de Lorraine | **Rapporteure** |
| **Eric DUCHENE**<br>Professeur des Universités, Université Claude Bernard Lyon 1 | **Examinateur** |
| **Christine LARGERON**<br>Professeure des Universités, Université Jean Monnet | **Examinatrice** |
| **Hubert NAACKE**<br>Maître de Conférences, Sorbonne Université | **Examinateur** |
| **Lynda TAMINE-LECHANI**<br>Professeure des Universités, Université Paul Sabatier | **Examinatrice** |

*To my parents, for their everyday love and sacrifice for us.*

**Abstract**

The age of social media has opened new opportunities for businesses. This flourishing wealth of information is outside traditional channels and frameworks of classical marketing research, including that of Marketing Mix Modeling (MMM). Textual data, in particular, poses many challenges that data analysis practitioners must tackle. Social media constitute massive, heterogeneous, and noisy document sources. Industrial data acquisition processes include some amount of ETL. However, the variability of noise in the data and the heterogeneity induced by different sources create the need for ad-hoc tools. Put otherwise, customer insight extraction in fully unsupervised, noisy contexts is an arduous task.

This research addresses the challenge of fully unsupervised topic extraction in noisy, Big Data contexts. We present three approaches we built on the Variational Autoencoder framework: the Embedded Dirichlet Process, the Embedded Hierarchical Dirichlet Process, and the time-aware Dynamic Embedded Dirichlet Process. These nonparametric approaches concerning topics present the particularity of determining word embeddings and topic embeddings. These embeddings do not require transfer learning, but knowledge transfer remains possible. We test these approaches on benchmark and automotive industry-related datasets from a real-world use case. We show that our models achieve equal to better performance than state-of-the-art methods and that the field of topic modeling would benefit from improved evaluation metrics.

Lastly, we leverage the Autoencoding Variational Bayes framework and Deep Learning to design a toolkit suitable for industrial practice. This toolkit allows for fast and scalable training and development of new models, thus bridging the gap between statistical modeling and software development and allowing for working with iterative project management methods and domain knowledge updates.

**Résumé**

L'ère des médias sociaux a ouvert de nouvelles perspectives aux entreprises. Cette richesse florissante d'informations se situe en dehors des canaux et des cadres traditionnels de la recherche marketing classique, y compris celui du Marketing Mix Modeling (MMM). Les données textuelles, en particulier, posent de nombreux défis que les praticiens de l'analyse de données doivent relever. Les médias sociaux constituent des sources de documents massives, hétérogènes et bruitées. Les processus industriels d'acquisition de données comprennent une certaine quantité d'ETL, cependant, la variabilité du bruit dans les données et l'hétérogénéité induite par les différentes sources créent le besoin d'outils ad hoc. En d'autres termes, l'extraction d'insight client dans des contextes bruités et totalement non supervisés est une tâche ardue.

Nous nous intéressons ici à l'extraction de thématiques entièrement non supervisée dans des contextes Big Data bruités. Nous présentons trois approches construites suivant le framework de l'autoencodeur variationnel : l'Embedded Dirichlet Process, l'Embedded Hierarchical Dirichlet Process et le Dynamic Embedded Dirichlet Process. Ces approches non paramétriques concernant les thèmes présentent la particularité de déterminer des embeddings de mots et des embeddings de sujets. Ces embeddings ne nécessitent pas d'apprentissage par transfert, mais le transfert de connaissances reste possible. Nous testons ces approches sur des jeux de données de référence et des jeux de données liés à l'industrie automobile, issus d'un cas d'utilisation réel. Nous montrons que nos modèles atteignent des performances égales ou supérieures à celles de l'état de l'art et que le domaine du topic modeling bénéficierait de meilleures mesures d'évaluation.

Enfin, nous tirons parti du cadre Autoencoding Variational Bayes et du Deep Learning pour concevoir une boîte à outils adaptée à la pratique industrielle. Cette boîte à outils permet un entraînement et un développement rapides et évolutifs de nouveaux modèles, comblant ainsi le fossé entre la modélisation statistique et le développement de logiciels et permettant de travailler à la fois avec les méthodes de gestion de projet itératives et les mises à jour de connaissances métier.

Here we are, after three years of doctoral studies, and more than a decade of learning at school and other intellectual and not-so-intellectual pursuits. This work is their culminating point, yet not an ending point. It is an additional step and a new starting point. Should I summarize, the most important lesson I have learned so far is that the only thing I am sure about is that I know very little, considering all that is known and - probably - what is unknown. I cannot wait to dig into all of this. For now, the time is for expressing my gratitude.

My first thoughts go to my thesis advisors, Stéphane BONNEVAY, Alexandre AUSSEM, and Bruno CANITIA. Thank you for your trust from the beginning to the end of this project. Also, thank you for the total liberty you granted me. I thank Xavier GIGNOUX from the Lizeo Group for welcoming me into his company.

I thank Julyan ARBEL, Marianne CLAUSEL, Eric DUCHENE, Christine LARGERON, Hubert NAACKE, and Lynda TAMINE-LECHANI for reviewing my work and being part of my thesis committee.

I want to thank the people I met before starting my doctoral studies, without whom I would not have even been able to reach this stage. Many thanks to my Bachelor and Master's Degree in Applied Mathematics and Social Sciences (MIASHS) teachers and colleagues at the University of Lyon 2. Thank you, also, for allowing me to teach the next generations of students and to be part of this passionating project again, this time as a teacher.

Finally, I thank my family and Sébastien WILLMANN for their unwavering support and kind words during hardships. Mom and Dad, thank you for everything. I am now plainly aware of all the sacrifices you went through and how much you and my brother love me. I will always love you too.

# Contents

# List of Figures

# List of Tables

# List of Algorithms

# Chapter 1

# Introduction

## 1.1 General considerations

The age of social media has opened new opportunities for businesses. Customers are no longer the final link of a linear value chain; they have also become informants and influencers as they review goods and services, talk about their buying interests and share their opinion about brands, manufacturers, and retailers. This flourishing wealth of information is outside traditional channels and frameworks of classical marketing research - including that of MMM [1] - and poses many challenges. Data analysis practitioners must tackle these challenges when testing the viability of a business idea or capturing the whole picture and the latest trends in consumers' opinions. Social media constitute massive, heterogeneous, and noisy document sources often accessed through web scraping when no API [2] is available. Data acquisition processes include some amount of ETL [3] or ELT [4]. However, the variability of noise in the data and the heterogeneity induced by different sources create the need for ad-hoc tools. In other words, even if large quantities of data are accessible for virtually free, customer insight extraction is arduous.
Additionally, documents' structure is frequently more complex than in classical applications, as documents can exhibit some linking in a graph (e.g., Twitter) or in the form of a nested

1. Marketing Mix Modeling.
2. Application Programming Interface.
3. Extract-Transform-Load.
4. Extract-Load-Transform.

hierarchy (e.g., Reddit). Linking between words and linking between topics are also crucial for efficient meaning extraction and better interpretation. Finally, customer trends tend to evolve through time, thus causing data drifts that create the need for Machine Learning models' updates – not to mention already existing data that businesses have not yet integrated into their information systems. This industrial context is paramount to understanding the choices and implications of this piece of research.

## 1.2 Industrial context

Our research originates from industrial needs for customer insight extraction from massive streams of texts from social media in a broad sense: technical reports, blogs, microblogs, and forum posts. The data sources are carefully selected; consequently, these documents all display technical details about the products or customer insights. The needs, however, solely cover the contents of these media and not their emitters. Lizeo IT provides our experimental material and harvests data daily. The company uses web scraping techniques on over a thousand websites in 6 different languages: English, French, Spanish, Italian, German, and Dutch. The data acquisition pipeline includes parsing and basic data-cleaning steps. However, the noise remains, e.g., markup languages, misspells, and documents in a given language that comes from a source supposedly in another language. Due to this noise and its variability, off-the-shelf tools seldom work. Plus, tools are only available in some domains, such as the tire industry, which is Lizeo IT's field of expertise. In-house data dictionaries and ontologies about the tire industry exist, but they rely on manual, expert knowledge-backed labeling that does not apply to other products. The company's intent for this project is to extract information without any background information - objectively observable elements inherent to data set aside - to work with data related to other industries. The aimed use case is *pure exploratory data analysis.* Unfortunately, data opaqueness, lack of data background, heterogeneity, and noisiness are all hindrances to the practitioner.

## 1.3 Desiderata

We pursue several goals :

**Explicit modeling** The primary purpose of the tools is exploratory data analysis. The latent structures or topics must enable practitioners to explore several dimensions with precise definitions.

**Nonparametric topic extraction** The data volumes are massive and come from diverse sources. We cannot anticipate the topics or their number, especially without prior information on new products.

**Integrative extraction** We need to extract customers' insights in a way that preserves document, topic, and word structures and relationships while considering temporal dependencies, languages, and noise.

**Generic Extraction without prior knowledge** Customers tend to *focus on their product experience*. They also discuss product characteristics. Without prior knowledge injection, we need to extract these insights to the fullest extent.

**Data cleaning processes improvement** We need our approaches to cope with noise directly, either by isolating or filtering it.

**Scalability** We must adapt to Big Data-like settings.

**Fast development** As practitioners extract information, knowledge reinjection to future iterations of the data analysis cycle becomes desirable. Moreover, other data properties (graphs, Etc.) may be available in the future. To benefit from this knowledge faster, we need a framework that enables model builders to add a hypothesis in the most seamless possible way.

## 1.4 Document outline

This document's organization follows the chronological order of our contributions. Chapter 1 lays the Bayesian foundations regarding modeling, parameter inference, and criticism, particularly in applying these concepts to topic modeling. We give an overview of the state of the art of topic modeling and its application to media mining. We also link classical probabilistic graphical modeling with the latest advances in deep generative modeling. Chapter 2 presents our theoretical framework and how it enables moving quickly between modeling phases. This framework is the first contribution to this project. In Chapter 3, we introduce two novel topic models we call Embedded Dirichlet Processes. These models can efficiently capture the number of topics along with their contents. They can also generate topic and word embeddings, thus enabling practitioners to see the correlation patterns in a given corpus. We also compare our approaches to the state-of-the-art on data proceeding from our industrial context. Finally, we present a novel dynamic extension to our Embedded Dirichlet Processes in Chapter 4. We test our models on two benchmark datasets and two others in different languages proceeding from our industrial context.

# Chapter 2

# State-of-the-art

In this Chapter, we lay the Bayesian foundations for probabilistic graphical modeling. We show its link with the exponential family of density distributions, parameter inference, model criticism, and how it relates to topic modeling. Among the different parameter inference methods, we insist on variational inference due to its ability to scale efficiently to massive text streams. We then detail the link between deep neural networks and probabilistic graphical modeling. In particular, we review autoencoders and their subsequent variants concerning probabilistic settings. We show that this kind of architecture leverages the advantages and cons of both probabilistic graphical modeling and neural networks. Finally, we review the state of the art of topic modeling. We present an overview of the latest wave of topic models, i.e., neural topic models. We also show how contextualization helps in getting meaningful topic representations. The Chapter ends with a presentation of a few applications of topic modeling to social media mining.

## 2.1 Bayesian foundations

Our work finds its roots in the field of Bayesian Statistics. In this section, we present some principles of probabilistic graphical modeling. In particular, we review the principles of model building, parameter inference, and model criticism.

### 2.1.1 Probabilistic Graphical Modeling

Probabilistic topic models are latent variable models at their very core, i.e., models where a specific data structure is assumed. They are a language whose building blocks are distributions and express dependencies between hidden and observed variables, thus forming a unifying framework. These variables often follow distributions that belong to the exponential family. A family of probability density functions (PDF) $\mathcal{P} = \{p_\theta : \theta \in \Theta\}$ on a measure space $(\chi, \mathcal{B}, \nu)$ forms an exponential family if:

$$Pr_\theta(\mathbf{x}) = \exp\left(\eta(\theta)^T t(\mathbf{x}) - A(\eta(\theta))\right) \tag{2.1}$$

$$A(\eta(\theta)) = \log \int \exp\left(\eta(\theta)^T t(\mathbf{x})\right) \nu(d\mathbf{x}) \tag{2.2}$$

In Eqn. 2.1 and Eqn. 2.2, $A(\eta(\theta))$ is the *log-normalizer*, $\eta(\theta)$ is the *natural parameter*, and $t(\mathbf{x})$ is the vector of sufficient statistics. These expressions are known and depend on the considered probability distribution. The exponential family of probability functions includes usual distributions, such as the Gaussian, the Dirichlet, the Gamma, the Beta, or the Poisson distributions. During the fitting step, practitioners try to unveil the structure's parameter values in terms of distributions of the latent variables, thus inscribing them in a Bayesian probabilistic frame. This kind of modeling is often referred to as graphical due to the representation of model structures.

These models are also generative by essence:

$$Pr(h, x \mid \eta) = Pr(h \mid \eta)\ Pr(x \mid h) \tag{2.3}$$

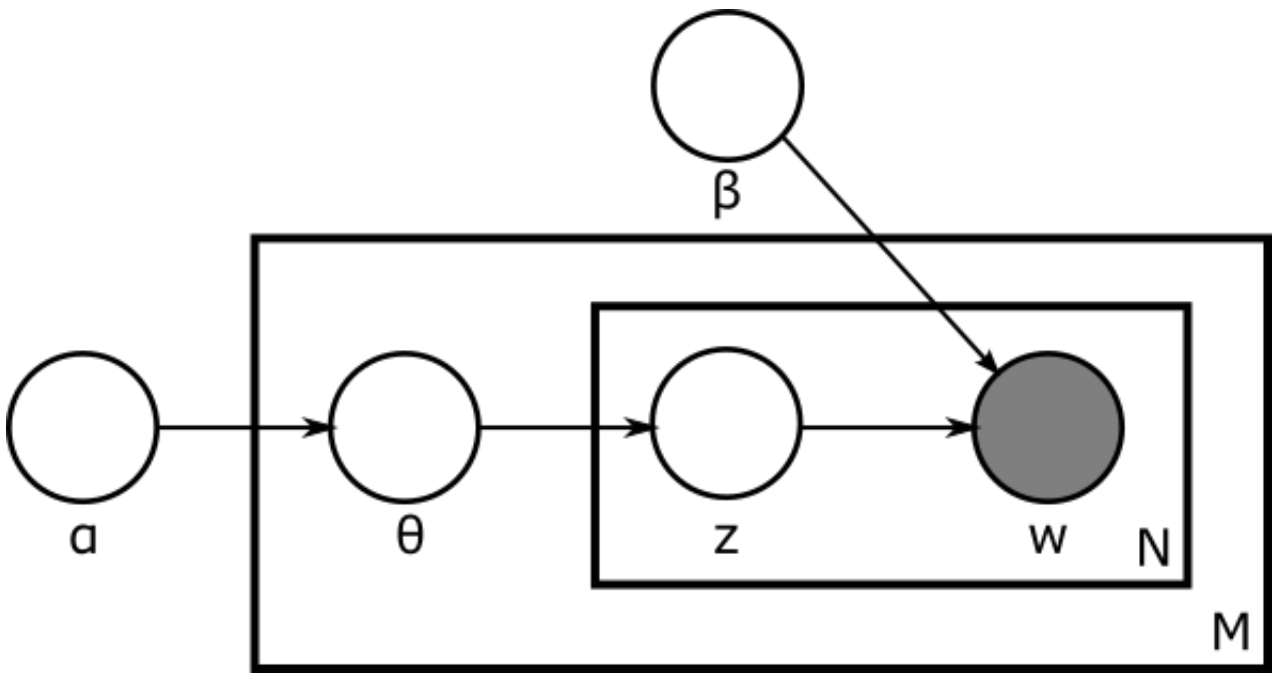

Figure 2.1 – LDA's graphical model

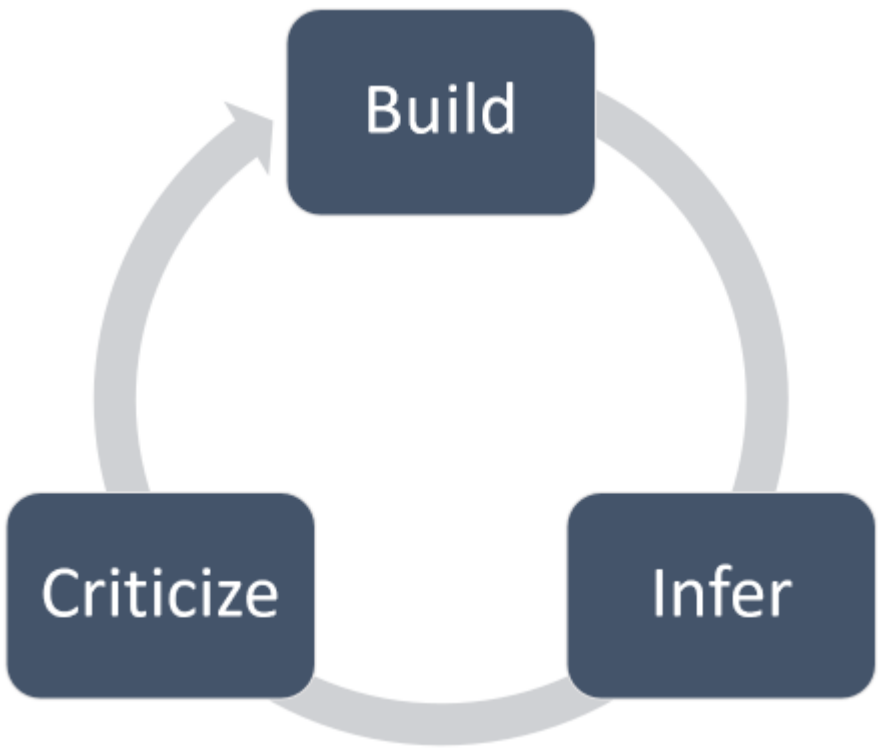

Figure 2.2 – A modeling framework: G. Box's loop

In Eqn. 2.3, $x$ denotes an *observation*, $h$ denotes a *latent variable*, and $\eta$ is a *fixed random parameter* or *hyperparameter*. As it turns out, and thanks to the Bayesian setting, latent variables models are helpful for *descriptive*, *exploratory*, and *predictive* purposes. As the models are also generative, they can theoretically serve as methods for data augmentation when data is too scarce to use with other techniques.

G. Box & al.'s framework ([Box76], Fig. 2.2) is a particularly pertinent and valuable framework for data modeling. The framework consists of three steps, from model building to model assessment, and includes the faculty of further iterations through the process without losing the benefit of the previous actions. The approach enables practitioners to build parsimonious solutions by adding variables stepwise, thus getting the minimal set of variables that best model the data in their view. However, it does not apply the fullest, orthodox Bayesian approach, as it does not encode all the uncertainty about the data structure but only the *variables of interest*. In the following sections, we detail each step and show how they apply to a probabilistic topic model through the example of the Latent Dirichlet Allocation (LDA) [BNJ03].

#### 2.1.1.1 Model building

The first step of the model building step consists of setting a generative process, i.e., to form the setting of underlying assumptions about the data. This process distinguishes between the *global latent variables* and the *local latent variables*. Let us consider LDA's document-wise generating process (Alg. 1).

**Algorithm 1** Generative process for the LDA

1: Choose $N_d \sim \text{Poisson}(\lambda)$
2: Choose $\theta_d \sim \text{Dirichlet}(\alpha)$
3: **for all** word $w_n$ in document $d$ **do**
4: Choose a topic $z_{dn} \sim \text{Multinomial}(\theta)$
5: Choose a word $w_{dn}$ from $\Pr(w_n | z_{dn}, \beta)$, a multinomial probability conditioned on the topic $z_{dn}$
6: **end for**

In the generative process (Alg. 1), $w_{dn}$ denotes the observations; $\alpha$ is a hyperparameter for the Dirichlet distribution; $\theta$ is a global latent variable that parameterizes a multinomial and that represents the document-wise topic mixture; $z_{dn}$ indicates a multinomial-generated topic; $\beta$ is a local latent variable that represents the word-wise topic mixture, and that also parameterizes a multinomial to generate a word. Considering the LDA's generative process, we can see what its building blocks are and that the model is hierarchical in the sense that these basic building blocks interact together. As such, and in principle, this apparent atomicity eases the task of creating new models by changing distributional assumptions or exchanging components from different models.

$$\Pr(\theta,\ \mathbf{z},\ \mathbf{w} \mid \alpha,\ \beta) = \Pr(\theta \mid \alpha) \times \prod_{n=1}^{N} \Pr(z_n \mid \theta) \Pr\left(w_n \mid z_n, \beta\right) \tag{2.4}$$

$$\Pr(\theta,\ z \mid w,\ \alpha,\ \beta) = \frac{\Pr(\theta,\ z,\ w \mid \alpha,\ \beta)}{\Pr(w \mid \alpha,\ \beta)} \tag{2.5}$$

$$\Pr(w \mid w_{obs}) = \int \sum_{z} \Pr(w \mid z) \Pr(z \mid \theta) \times \Pr(\theta \mid w_{obs}) d\theta \tag{2.6}$$

Formalizing this setting helps to form the joint distribution and the graphical model that fits the variables. The joint distribution (Eq. 2.4), in turn, is used to define the posterior (Eq. 2.5), thanks to Bayes' theorem. Last but not least, marginalizing the posterior over the global latent variables enables deducing the predictive distribution (Eq. 2.6) given the observations.

In this Section, we have seen that the formalization step is essential to leveraging the model and setting how variables interact together. One of the most significant difficulties in using latent variable models is to reverse the data generating process in the inference step, as the latent quantities are unknown by definition and design.

| **Property** | **Method** | |
|---|---|---|
| | *MCMC* | *VI* |
| *Engine* | Sampling | Optimization |
| *Core differences* | A tool for simulating densities | A tool for approximating densities |
| *Theoretical guarantees* | Asymptotically, computes exact samples from the target density | Moderate certainty of the results only |

Table 2.1 – Approximate inference methods' properties

| **Criterion** | **Method** | |
|---|---|---|
| | *MCMC* | *VI* |
| *Problem scale* | Small (expensive to compute) | Large (generally fast) |
| *Need for precision* | Great | Moderate |
| *Certainty of model specification* | Great | Moderate |
| *Fidelity to the geometry of the posterior* | Gibbs sampling can help, but the methods generally fall short in complex settings | Is not guaranteed |
| *Relative accuracy* | Asymptotically perfect | Generally underestimates the variance of the posterior |

Table 2.2 – Inference method selection guide

#### 2.1.1.2 Parameter inference

Parameter inference is not only necessary to understand the data but also one of the keys to the journey to model *scalability*. As exact inference is seldom possible due to the usual evidence's intractability in the posterior distribution, practitioners usually fall back on approximate inference. Among the existing methods, three are fit to determine the value of the latent quantities. These approaches are *Laplace approximation*, *Markov Chain Monte Carlo* (MCMC), and *Variational Inference* (VI).

Laplace approximation represents the posterior as a Gaussian distribution, derived from *Taylor's theorem*. However, it is not a convenient analytical tool to handle data from other distributions, and therefore, the literature is much more focused on MCMC and VI methods. While MCMC forms a Markov chain over the hidden variables whose stationary state is the posterior, VI posits a variational objective whose optimization offers a reasonable approximation of the desired quantities [BKM16]. Both methods are equivalent to choosing between two different engines, as they both have valuable properties depending on their use case. Before choosing, and besides different foundations (Tab. 2.1), one has to assess the needs of the model application (Tab. 2.2). The criteria to consider are the problem scale, the need for precision, and certainty.

MCMC is a tool for simulating densities at its very core. It offers *asymptotical* certainty of exact computing samples from the target density. These methods are helpful when the need for precision is predominant and the model specification is neat and clear, like in conjugate distributional contexts. However, it falls short when the setting is too complicated or when the data to analyze is massive, as in Big Data contexts. VI, on the other hand, can succeed in settings where MCMC falls short, including in massive scale, non-conjugate distributional settings, and even non-convex settings [1]. This basis also is VI's weak spot, as this technique is prone to underestimating the variance of the posterior. Our work relies on VI for three reasons. The first one is that despite the theoretical guarantees it offers,

1. As this Section's contents are mathematically dense, we refer the reader to [HBB10] for an application of online variational inference to the LDA model to avoid clutter.

most classical MCMC methods such as Gibbs sampling are too slow for parameter inference in our industrial setting. The second reason is that despite the existence of online Gibbs sampling variants [DB16], adding a new modeling hypothesis to an existing model requires devising an entirely new inference process, thus making model updates slightly tricky. The third reason is that VI is able to handle nonconjugate settings [BL06; WB13].

VI approximates the posterior distribution using optimization. To achieve this goal, it posits the existence of a family of distributions that match the posterior. This family is called variational. The Kullback-Leibler divergence (KLD) defines this closeness between the approximate distribution and the true posterior. Thus, the optimization procedure aims to find the parameters that minimize the KLD between the variational family and the true posterior. Let $\lambda$ be this set of parameters. The optimization objective is the following:

$$\lambda^* = \arg\min_{\lambda} \mathrm{KLD}\left(q\left(\theta, z; \lambda\right) \parallel \Pr\left(\theta, z \mid w\right)\right) \tag{2.7}$$

Eqn 2.7 is intractable due to the posterior. It is, however, possible to re-express it:

$$\mathrm{KLD}\left(q\left(\theta, z; \lambda\right) \parallel \Pr\left(\theta, z \mid w\right)\right) = \log \Pr\left(w\right) - \mathbb{E}_{q(\theta,z;\lambda)}\left[\log \frac{\Pr\left(w, z, \theta\right)}{q\left(\theta, z; \lambda\right)}\right] \tag{2.8}$$

As the first term in Eqn. 2.8 does not depend on the set of parameters to optimize, our goal is equivalent to minimizing the expectation. Doing so enables defining a lower bound on the expectation, hence the name of Evidence Lower BOund (ELBO).

$$\mathrm{ELBO} = \mathbb{E}_{q(\theta,z;\lambda)}\left[\log \frac{\Pr\left(w, z, \theta\right)}{q\left(\theta, z; \lambda\right)}\right] \leq \log \Pr\left(w\right) \tag{2.9}$$

It is possible to (tractably) approximate the ELBO provided a tractable variational density exists. The following sections review two ways of performing VI: mean-field VI and black-box VI. We close this section by showing how Deep Learning can leverage these techniques for parameter inference.

**2.1.1.2.1 Mean-field variational inference** As its name states, Mean-field VI (MFVI) makes the mean-field assumption, i.e., we can write the variational as follows:

$$q(\theta, z; \lambda) = q(\theta; \lambda_\theta) \prod_{i=1}^{N} q(z_i; \lambda_i) \tag{2.10}$$

with $\lambda = (\lambda_\theta, \lambda_1, \ldots, \lambda_N)$. The ELBO becomes the following:

$$\text{ELBO}(w, \lambda) = \mathbb{E}_{q(\theta;\lambda_\theta)\prod_{i=1}^{N} q(z_i;\lambda_i)} \Big\{ \log \frac{\Pr(\theta)}{q(\theta)} + \sum_{i=1}^{N} \log \frac{\Pr(z_i \mid \theta)}{q(z_i; \lambda_i)} + \sum_{i=1}^{N} \log \Pr(w_i \mid z_i, \theta) \Big\} \tag{2.11}$$

**2.1.1.2.2 Black-box variational inference** We follow A. B. Dieng's thesis formalism [Die21] and rewrite the ELBO's expression including all the latent variables in z. The ELBO becomes the following:

$$\text{ELBO} = \mathbb{E}_{q(z;\lambda)} [\log \Pr(x, z) - \log \text{q}(z; \lambda)] \tag{2.12}$$

Instead of the MF assumption, BBVI optimizes the ELBO using a Monte Carlo to approximate its gradients:

$$\begin{aligned}
\nabla_\lambda \text{ELBO} &= \nabla_{\boldsymbol{\lambda}} \int [q(\mathbf{z}; \boldsymbol{\lambda}) \log \Pr(\mathbf{x}, \mathbf{z}) - q(\mathbf{z}; \boldsymbol{\lambda}) \log q(\mathbf{z}; \boldsymbol{\lambda})] d\mathbf{z} \\
&= \int [\nabla_{\boldsymbol{\lambda}} q(\mathbf{z}; \boldsymbol{\lambda}) \log \Pr(\mathbf{x}, \mathbf{z}) - \nabla_{\boldsymbol{\lambda}}(q(\mathbf{z}; \boldsymbol{\lambda}) \log q(\mathbf{z}; \boldsymbol{\lambda}))] \, d\mathbf{z} \\
&= \int [\log \Pr(\mathbf{x}, \mathbf{z}) \nabla_{\boldsymbol{\lambda}} q(\mathbf{z}; \boldsymbol{\lambda}) - \log q(\mathbf{z}; \boldsymbol{\lambda}) \nabla_{\boldsymbol{\lambda}} q(\mathbf{z}; \boldsymbol{\lambda}) - q(\mathbf{z}; \boldsymbol{\lambda}) \nabla_{\boldsymbol{\lambda}} \log q(\mathbf{z}; \boldsymbol{\lambda})] \, d\mathbf{z} \\
&= \int [\log \Pr(\mathbf{x}, \mathbf{z}) - \log q(\mathbf{z}; \boldsymbol{\lambda})] \nabla_{\boldsymbol{\lambda}} q(\mathbf{z}; \boldsymbol{\lambda}) d\mathbf{z} - \int \nabla_{\boldsymbol{\lambda}} q(\mathbf{z}; \boldsymbol{\lambda}) d\mathbf{z} \\
&= \int q(\mathbf{z}; \boldsymbol{\lambda}) [\log \Pr(\mathbf{x}, \mathbf{z}) - \log q(\mathbf{z}; \boldsymbol{\lambda})] \nabla_{\boldsymbol{\lambda}} \log q(\mathbf{z}; \boldsymbol{\lambda}) d\mathbf{z} - \nabla_{\boldsymbol{\lambda}} \int q(\mathbf{z}; \boldsymbol{\lambda}) d\mathbf{z} \\
&= \mathbb{E}_{q(\mathbf{z}\boldsymbol{\lambda})} [(\log \Pr(\mathbf{x}, \mathbf{z}) - \log q(\mathbf{z}; \boldsymbol{\lambda})) \nabla_{\boldsymbol{\lambda}} \log q(\mathbf{z}; \boldsymbol{\lambda})]
\end{aligned} \tag{2.13}$$

We used the following identities: $\int \text{q}(z; \lambda) \, dz = 1$ and $\nabla_\lambda \log \text{q}(z; \lambda) = \frac{\nabla_\lambda \text{q}(z;\lambda)}{\text{q}(z;\lambda)}$. It is possible to use a Monte Carlo procedure to approximate this expectation thanks to the below formula, which is also an unbiased and consistent estimator of the actual score gradient. This estimator, however, is known to have a high variance. In this work, we use *pathwise derivatives* - also known as the *reparamaterization trick* (RT) - to approximate gradients of Monte Carlo objectives.

$$\nabla_\lambda \text{ELBO} \approx \frac{1}{S}\sum_{s=1}^{S}\left(\log \Pr\left(w, z^s\right) - \log \mathrm{q}\left(z^s; \lambda\right)\right) \nabla_\lambda \mathrm{q}\left(z^s; \lambda\right) \tag{2.14}$$

The RT's name comes from the fact that it introduces variables $\epsilon$ whose distribution $\mathrm{q}(\epsilon)$ do not depend on $\lambda$. Under this assumption, the ELBO becomes the following:

$$\text{ELBO} = \mathbb{E}_{q(\epsilon)}\left[\log \Pr\left(w, g\left(\epsilon, \lambda\right)\right) - \log \mathrm{q}\left(g\left(\epsilon, \lambda\right); \lambda\right)\right]) \tag{2.15}$$

where $z \sim \mathrm{q}(z; \lambda) \Leftrightarrow \epsilon \sim \mathrm{q}(\epsilon)$ and $z = \mathrm{g}(\epsilon; \lambda)$. Its gradient is written as follows:

$$\nabla_\lambda \text{ELBO} = \mathbb{E}_{\mathrm{q}(\epsilon)}\left[\log \Pr\left(w, \mathrm{g}\left(\epsilon; \lambda\right)\right) - \log \mathrm{q}\left(\mathrm{g}\left(\epsilon; \lambda\right); \lambda\right)\right] \tag{2.16}$$

We can use the following formula for Monte Carlo approximation:

$$\nabla_\lambda \text{ELBO} \approx \frac{1}{S}\sum_{s=1}^{S}\nabla_\lambda\left[\log \Pr\left(w, \mathrm{g}\left(\epsilon; \lambda\right)\right) - \log \mathrm{q}\left(\mathrm{g}\left(\epsilon; \lambda\right); \lambda\right)\right] \tag{2.17}$$

As per [KW14], in this work, we set $S = 1$ as it has proven enough for learning procedures.

In Deep Learning, the reparameterization trick proves useful for two reasons. On the one hand, it enables using BBVI. Conversely, it implies deterministic, differentiable, and equivalent density transformations of some specified distribution. These transformations make it possible to use backpropagation through stochastic nodes. It is the procedure used in variational autoencoders [KW14] (VAEs). The reparameterization trick, however, comes with a downside called latent variable collapse. In VAEs, latent variable collapse is when the variational posterior stops depending on the data, i.e., when the approximate posterior becomes so close to the prior that the posterior estimates of the latent variable do not represent the data's underlying structure. Formally, we can express posterior collapse as follows: $q_\phi(\mathbf{z} \mid \mathbf{x}) \approx p(\mathbf{z})$. Several authors have reported on the issue [Bow+16; Søn+17; Kin+16; Che+17; ZSE17; Yeu+17].
For instance, consider $z \sim \mathcal{N}(\mu, \sigma^2)$; a simple data point-wise reparameterization is the following :

$$z^{(i)} = \mu + \sigma \otimes \epsilon \quad \text{and} \quad \epsilon \sim \mathcal{N}(0, \mathbf{I}) \tag{2.18}$$

In the above equation, $\mu$ is the location parameter for a (variational) Gaussian, and is $\sigma$ its scale parameter. Finally, $\otimes$ stands for the Hadamard product. Without loss of generality, we will refer to this specific case where $g(.)$ is a standardization function under the name of reparameterization by standardization (RBS). RBS works with any distribution that admits location-scale parameterization. We find the Gaussian, Logistic, and Student's t distributions among these distributions. Another possible variant implies using a tractable inverse cumulative distribution function (CDF) for $g(.)$ and $\epsilon \sim \mathcal{U}(0, I)$. This variant is usable with distributions such as the Exponential, the Weibull, or the Gumbel distributions. A third possibility is to express random variables following a given distribution as a composition of random variables that follow other distributions. For instance, if $\sigma \sim \text{Gamma}(\frac{\nu}{2}, \frac{\nu}{2})$ then $z \sim \mathcal{N}(0, \sigma^2)$ is a Student t distribution; if $z_1 \sim \text{Gamma}(\alpha, 1)$ and $z_2 \sim \text{Gamma}(\beta, 1)$ then $\frac{z_1}{z_1+z_2} \sim \text{Beta}(\alpha, \beta)$; if $z_i \sim \text{Gamma}(\alpha_i, 1)$ then $(\frac{z_1}{\sum_{j=1}^{D} z_j}, \frac{z_2}{\sum_{j=1}^{D} z_j}, \ldots, \frac{z_D}{\sum_{j=1}^{D} z_j}) \sim \text{Dirichlet}(\alpha_1, \ldots, \alpha_D)$. Other suitable approximations to the inverse CDF exist, see [Dev86].
In this Section, we have presented three methods for parameter inference and deep-dived into VI due to its ability to cope with complex settings and to scale to massive data sets. Following the parameter inference step is the evaluation step. This evaluation must consider several aspects regarding a topic model's use cases to be adequate.

#### 2.1.1.3 Model criticism

Probabilistic topic models are a flexible range of techniques. These Bayesian methods can simultaneously act as feature extractors, dimensionality reduction techniques, and language models. Most of their applications in the scientific literature imply them as tools for data exploration as practitioners expect the latent variables to convey some meaning, thus yielding precious insights about the dataset at hand and easing model interpretations. This piece of research falls within this use case.

Fitting a topic model is a complex matter that needs careful handling. As probabilistic topic models are generative models, statistical goodness-of-fit is paramount. However, overlooking the semantic aspect of topic modeling would go against the foundational assumption that models capture semantically meaningful latent variables. This assumption is the principal reason for making them first-class citizens of *unsupervised* corpora exploration. The scientific literature does not show much change in the topic models' training and, perhaps more importantly, topic model evaluation since Chang & al.'s work ([Cha+09]). With the profusion of information due to the Big Data phenomenon, overall interpretability is more than ever a concerning issue that remains open to this day. In the following sections, we present two kinds of evaluations: quantitative and qualitative.

**2.1.1.3.1 Statistical assessment** Predictions are the intrinsic objective for this set of metrics. These metrics were also the first ones to appear in scientific literature. As probabilistic topic models are generative, they include some discriminative capacities. In other words, it is possible to infer a set of topics for a given document. They can also reduce a document's dimensionality, i.e., describe a document by a set of topics instead of words. When performing classification tasks, authors feed these topics to their downstream classifiers and measure their performance with classical metrics for classification. Doing so implies disposing of labeled data, which is not our case. In this piece of research, we try to uncover latent, unknown topics from nonlabeled documents coming from the Internet without any prior knowledge. As such, and despite their existence, we can not use these indicators. Instead, we focus on likelihood-based methods.
The most commonly used statistical metric is perplexity as per [Wal+09]. Perplexity measures the likelihood of an unseen document. Practitioners compute it on a held-out sample from an unseen test dataset. Its formulation is the following:

$$\text{perplexity}(\text{ test set } w) = \exp\left\{-\frac{\mathcal{L}(w)}{\text{count of tokens}}\right\} \tag{2.19}$$

In Eqn 2.19, $\mathcal{L}(w)$ is the model's log-likelihood. Authors also use the ELBO as a surrogate. Qualitative evaluation is not the only way to assess a topic model; one must also assess its quality.

**2.1.1.3.2 Quality assessment** When discussing a topic model's quality, authors can refer to two concepts: *topic coherence* and *topic interpretability.* These concepts are complex enough to justify further discussions. These discussions are beyond the scope of this work yet still need some definition to understand this research. We call topic coherence the lexically sensible co-occurrence of words. On the other hand, we call topic interpretability the possibility of unequivocally naming a topic.

In 2009, Chang & al. [Cha+09] released an article that stands out comparatively with previous works on topic model evaluation that only use perplexity. To introduce their line of thinking, they start by reconnecting the Latent Semantic Analysis (LSA) [Dee+90], i.e., one of the very first topic models, to its origins in the field of Psychology. In particular, the authors state that psychologists used the LSA to replicate human reasoning. The assumption still holds more or less implicitly. In more recent works, authors sometimes involve *humans in the loop* at different stages either by displaying a qualitative evaluation of topic models or twisting the inference process to include human insights [HBS11]. Chang & al. are the authors that went the furthest down the path of evaluating a topic model concerning a set of almost purely human criteria of their time, thus re-establishing the links between the fields of Psychology research and mainstream Machine Learning research.

Their work is insightful and helps determine how analysts understand and interpret *supposedly* semantic latent spaces. To assess a topic model's results, the authors used the probabilistic Latent Semantic Indexing (pLSI) [Hof99], the Latent Dirichlet Allocation (LDA), and the Correlated Topic Model (CTM) [LB05]. They asked a sample of people to perform two intruder detection tasks: one on topics and one on words. The experiment presented document samples and a randomized selection of topics with the highest probability according to a topic model, plus an intruder topic. The procedure also included a similar task for words. Topics and words are the analytic levels the simplest topic models summarize within their latent space. The working hypothesis is that if topics (respectively words) are coherent with a document's content, then the test subjects should non-randomly select the intruder topic, and they should answer randomly otherwise. In other words, the collective answers about the same elements experiment will allow judging about the semantic goodness-of-fit. The results confirm that a model's raw predictive power does not determine its semantical quality; on the contrary, i.e., models with lower log-likelihood than their counterparts were deemed better regarding semantics. Additionally, the more specific or fine-grained the topic gets, the fewer users can interpret them. The results suggest that relative semantic homogeneity and topic separateness are key to interpretability. To assess this aspect, Dieng & al. [DRB20] use topic diversity:

$$\text{Topic diversity} = \frac{\text{Number of unique tokens among the set of topics}}{\text{Total number of tokens in the set of topics}} \tag{2.20}$$

We retain from Chang & al.'s and previous work that achieving significant latent spaces and high statistical goodness-of-fit is as tricky as achieving a good bias-variance tradeoff in a classification task.

Despite these insights, we think the experimental protocol's design does not fully evaluate a probabilistic topic model's inner latent space, nor does it comply with reproducibility requirements for evaluation. The same applies to subsequent rating-based works that use human rating [New+10; Mim+11; AS13]. We explain our position about the protocol not being a complete evaluation of the latent space by the fact that to evaluate it, the authors have devised tasks to evaluate document-topic and topic-word levels *separately* and only inquire about intruders. The task does not determine the extent of a document's "correct topics" representativity. Also, it does not clarify what a "good" model is, nor what is a "semantically homogeneous" topic. These points are essential, as the whole mixture of elements requires an evaluation. Besides, the protocol only applies to the *simplest* probabilistic topic models; the protocol would fall short in evaluating more complicated settings.

The issues of non-reproducibility, costs, incompleteness, and discrepancy with statistical indicators of human evaluation have created the need for metrics for automated evaluation. Scientific literature includes three notable attempts at capturing coherence ([New+10; Mim+11; LNB14]). All of these metrics are mutual information-based. The most common indicator for topic coherence is the Natural Pointwise Mutual Information (NPMI) ([LNB14]). Authors usually report the mean NPMI, whose formula is the following:

$$\text{Topic coherence} = \frac{1}{K} \sum_{k=1}^{K} \frac{1}{45} \sum_{i=1}^{n} \sum_{j=i+1}^{n} f\left(w_i^{(k)}, w_j^{(k)}\right) \tag{2.21}$$

where $\left\{w_1^{(k)}, \dots, w_n^{(k)}\right\}$ is the set of the top-$n$ most likely words in topic $k \in \{1, \dots, K\}$ and $f$ is the normalized mutual information:

$$f\left(w_i, w_j\right) = \frac{\log \frac{\Pr(w_i, w_j)}{\Pr(w_i)\Pr(w_j)}}{-\log \Pr\left(w_i, w_j\right)}. \tag{2.22}$$

$\Pr\left(w_i, w_j\right)$ is the probability of words $w_i$ and $w_j$ co-occurring in a document and $\Pr\left(w_i\right)$ is the marginal probability of word $w_i$. These quantities result from the empirical count of words. This indicator oscillates between -1 and +1, with -1 indicating that the considered words never occur together, 0 indicating the absence of link, and +1 indicating that the considered words systematically appear together. The underlying assumption of (mean) NPMI is that a coherent topic's word should appear in the same context, i.e., document. This assumption corresponds to Harris' distributional hypothesis, which is the exact one that underlies Mikolov & al.'s works on word embeddings [Mik+13].To take all the aspects of quality into account, Dieng & al. [DRB20] consider it as the product between topic diversity and the topic coherence, hence:

$$\text{Topic quality} = \text{topic diversity } \times \text{ topic coherence} \tag{2.23}$$

The NPMI has been the golden standard for topic modeling evaluation for nearly a decade without being challenged. In 2021, Hoyle & al. presented a meta-analysis paper [Hoy+21] that assesses the validity of the usual evaluation when used with neural topic models (see Section 2.2.1), among others. They show that human judgment differs substantially from automated metrics and that automated metrics tend to exaggerate model differences compared with human judgment, thus undermining their utility for model selection. Moreover, the authors report that the metrics favor blurrier topics. These observations do not lead them to prefer human evaluation: it is costly and has reproducibility issues. They also show that using word familiarity as a substitute for domain expertise is not a satisfying solution. Therefore, the issue of evaluating a topic model is still an open one. This issue is particularly challenging due to the increasing interaction between probabilistic graphical modeling and deep generative learning.

### 2.1.2 Deep Learning and Probabilistic Graphical Modeling

Deep Learning has proven useful in several domains, text included. In unsupervised settings, its role is complementary to probabilistic graphical modeling. On the one hand, probabilistic graphical modeling aims to specify a mathematical structure of a dataset in the form of latent variables that follow hypothetical distributions. On the other hand, a neural network aim at capturing data structure, e.g., its correlations. Hornik & al. [HSW89] have shown that neural networks can represent any function; as such, they are flexible settings that can capture complex data dependencies and links in datasets. This flexibility is also their Achille's heel, as they are prone to overfit datasets, hence the need for specific regularization methods.

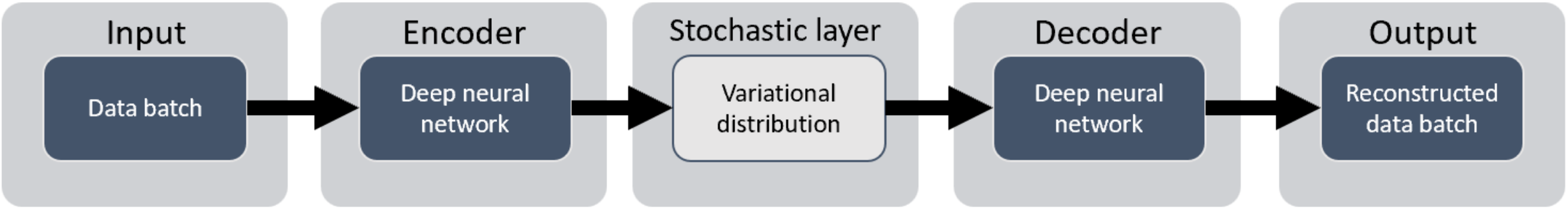

Figure 2.3 – Simplified representation of a VAE's logic

Notable contributions to deep unsupervised learning include the AutoEncoder (AE). The AE is fundamentally a dimensionality reduction framework that aims to learn a code from data, i.e., a latent, low-dimension variable called code that contains the essential characteristics of data [HS06]. The setting uses a duo of neural networks to achieve this goal; a first neural network called an encoder outputs the code. Then, a second neural network called a decoder tries reconstructing the input data thanks to the sole code. The variational autoencoder (VAE) [KW14], i.e., the probabilistic extension to the AE framework, is particularly interesting. The VAE's logic is close to AEs (Fig. 2.3). The models, however, differ because VAEs do not extract an intrinsic code from data but an underlying distribution. We provide additional practical reasons for focusing on VAEs in Appendix C. In Fig. 2.3, we display stochastic elements involving variational parameters in a white cell. This distribution serves to generate samples [2] that go through a *deterministic* standardization process. This deterministic aspect is paramount to parameter inference using backpropagation [RHW86].

The framework presents the property of matching model design with posterior inference. In other words, it uses both a data model and an approximate posterior over the latent

2. These samples are the stochastic equivalent to the code in the AE context.

variables:

$$Pr_\beta(\mathbf{w}, \mathbf{z}) = Pr\left(\mathbf{w} \mid f_\beta(\mathbf{z})\right) \cdot Pr(\mathbf{z}) \tag{2.24}$$

$$q_\phi(\mathbf{z} \mid \mathbf{w}) = q_\phi\left(\mathbf{z} \mid g_\phi(\mathbf{w})\right) \tag{2.25}$$

In Eqn. 2.24 and Eqn. 2.25, $f_\beta$ and $g_\phi$ are both neural networks. $f_\beta$ defines the likelihood, whereas $g_\phi$ parameterizes the posterior. In scientific literature, $g_\phi$ is called an encoder and $f_\beta$ is the decoder. The likelihood belongs to the exponential family.

Practitioners use stochastic gradient ascent to train both neural networks whose weights are the model's parameters. The optimization objective is still the ELBO, whose expression is as follows:

$$\text{ELBO}(\boldsymbol{\beta}, \phi) = \mathbb{E}_{q_\phi(\mathbf{z}|\mathbf{w})}\left[\log p_{\boldsymbol{\beta}}(\mathbf{w}, \mathbf{z}) - \log q_\phi(\mathbf{z} \mid \mathbf{w})\right] \tag{2.26}$$

For a given set of parameters $\phi$, maximizing the ELBO with respect to $\beta$ is equivalent to maximizing the likelihood of observations. However, maximizing the ELBO with respect to $\phi$ for a given set of parameters $\beta$ has two possible interpretations.
The first possible perspective is that of KL minimization. It is possible to write the ELBO as follows:

$$\text{ELBO}(\boldsymbol{\beta}, \phi) = \mathbb{E}_{q_\phi(\mathbf{z}|\mathbf{w})}\left[\log \Pr_\beta(\mathbf{z} \mid \mathbf{w}) + \log \Pr_\beta(\mathbf{w}) - \log q_\phi(\mathbf{z} \mid \mathbf{w})\right] \tag{2.27}$$

As $\log \Pr_\beta$ has no dependency on $\phi$, maximizing the expression with respect to $\phi$ is the same as minimizing the KLD between the variational $q_\phi(\mathbf{z} \mid \mathbf{X})$ and the true posterior $\Pr_\beta(\mathbf{z} \mid \mathbf{w})$. Under these assumptions, the objective becomes the following:

$$\text{ELBO}(\boldsymbol{\beta}, \phi) = -\text{KLD}\left(q_\phi(\mathbf{z} \mid \mathbf{w}) \| \Pr_\beta(\mathbf{z} \mid \mathbf{w})\right) + cst \tag{2.28}$$

The second perspective is that of a regularized AE. The ELBO is expressable as follows:

$$\text{ELBO}(\boldsymbol{\beta}, \phi) = \mathbb{E}_{q_\phi(\mathbf{z}|\mathbf{w})}\left[\log \Pr_{\boldsymbol{\beta}}(\mathbf{w} \mid \mathbf{z})\right] - \text{KLD}\left(q_\phi(\mathbf{z} \mid \mathbf{w}) \| \Pr(\mathbf{z})\right) \tag{2.29}$$

Without loss of generality, we consider a Gaussian with identity variance likelihood case. We assume a standard Gaussian prior and optimization through BBVI with RBS. Also, let $\mathbf{w}$ be a set of data points. Under these assumptions, the ELBO is the following:

$$\mathrm{ELBO}(\boldsymbol{\beta}, \phi) = \left\| \mathbf{w} - f_{\boldsymbol{\beta}}\left(\mathbf{z}_{\phi}(\mathbf{w})\right) \right\|_2^2 - \frac{1}{2} \left\| g_{\phi}(\mathbf{w}) \right\|_2^2 \tag{2.30}$$

The first term is the objective that corresponds to that of an AE; the second regularizes the parameters $\phi$ to bound the encoder's $L_2$ norm. The supplementary Gaussian noise from the RBS is also a form of regularization. Its purpose is to enable the trained decoder to simulate new data.

It is possible to use a conjunction of amortized variational inference (AVI) and MFVI to fit a VAE. AVI means the VAE passes data through a shared network to compute the approximate posterior. The procedure amortizes the cost of inference for models with local latent variables. On the contrary, the MFVI procedure uses the following factorization:

$$q\left(\boldsymbol{\beta}, \mathbf{z}_{1:N}; \boldsymbol{\lambda}\right) = q\left(\boldsymbol{\beta}; \boldsymbol{\lambda}_{\beta}\right) \cdot \prod_{i=1}^{N} q\left(\mathbf{z}_i; \boldsymbol{\lambda}_i\right) \tag{2.31}$$

This factorization implies the independence of all latent variables. It is possible to relax the assumption by including *conditional* independence between the local latent variables $\mathbf{z}_{1:N}$ and the global variables $\beta$. Consequently:

$$\begin{aligned} q\left(\boldsymbol{\beta}, \mathbf{z}_{1:N}; \boldsymbol{\lambda}\right) &= q\left(\boldsymbol{\beta}; \boldsymbol{\lambda}_{\beta}\right) \cdot \prod_{i=1}^{N} q\left(\mathbf{z}_i \mid \boldsymbol{\beta}; \boldsymbol{\lambda}_i\right) \\ &= q\left(\boldsymbol{\beta}; \boldsymbol{\lambda}_{\beta}\right) \cdot \prod_{i=1}^{N} q\left(\mathbf{z}_i \mid \mathbf{x}_i, \boldsymbol{\beta}; \boldsymbol{\lambda}_i\right) \end{aligned} \tag{2.32}$$

To connect MFVI with AVI, we assume the set of global latent variables $\beta$ to represent an *a posteriori* neural network with parameters $\lambda_{\beta}$ and the latent variables $\mathbf{z}_i$ to have their distribution through the data points. Doing so is equivalent to passing a data point through a neural network. Thus, we get the following expression:

$$q\left(\boldsymbol{\beta}, \mathbf{z}_{1:N}; \boldsymbol{\lambda}\right) = \prod_{i=1}^{N} q\left(\mathbf{z}_i \mid \mathbf{x}_i, \boldsymbol{\lambda}_\beta\right) \tag{2.33}$$

In this section, we have presented the Bayesian foundations for probabilistic topic modeling. We have also linked these foundations to deep generative learning. In the next section, we give an overview of state-of-the-art of topic modeling.

## 2.2 Topic modeling

This section gives the reader an overview of state-of-the-art topic modeling. We mainly focus on neural topic modeling, time dynamics, and contextualization. We then present some applications of topic modeling to media mining.

### 2.2.1 Neural topic modeling

Most neural topic extractors follow a VAE's logic [KW14]. The VAE is a generative neural framework that allows for simplified variational inference on large datasets. As it uses the same classical backpropagation from Deep Learning, one only needs to derive a formula for the lower bound on the likelihood of the model. This bound is usually referred to as Evidence Lower Bound, or ELBO. For simplicity, we will use the terms ELBO and likelihood interchangeably. Miao et al. [MYB16] have devised one of the earliest models of the kind: the Neural Variational Document Model. The Neural Variational Document Model suffers from posterior collapse despite its increased scalability and precision compared with non-neural topic models. To circumvent the issue, Srivastava & Sutton developed the ProdLDA [SS17]. Its main contribution is that it tries to get closer to a Dirichlet prior distribution thanks to its approximate relationship with the logit normal distribution. Additionally, the prior takes place in the simplex as expected for *compositional data modeling.* The ProdLDA makes topic modeling with a VAE more efficient, thus highlighting the importance of Dirichlet-like priors for topic modeling. LDA-like generative processes rely on the conjugacy relationship between the Dirichlet distribution and the categorical distribution used to indicate topics.

Other Gaussian-based developments include the TopicRNN [Die+17] and the Embedded Topic Model (ETM) [DRB20] (by chronological order). TopicRNN and the ETM both include unsupervised word information, under sequential form [Die+17] and embedding form [DRB20], respectively. These models, however, are fully parametric concerning the number of topics; consequently, it is compulsory to run them several times to find the optimal number of topics.

One of the Bayesian statistic's classical ways of dealing with compositional data whose number of mixture components [3] is not specified beforehand is the Dirichlet process (DP) family. A DP is a stochastic process that yields a probability distribution. To achieve this result, it takes on two parameters: a concentration parameter and a base distribution. In other words, a DP is a distribution built on another distribution. The choice of the base distribution is paramount and strongly depends on the use case. In topic modeling, practitioners usually choose a discrete measure. As the name indicates, the DP is closely related to the Dirichlet distribution; it is considered an infinite-dimensional Dirichlet. There are several ways to construct a DP. In variational inference, the most used one is the *stick-breaking construction* (SBC) [4]. Miao & al. [MYB16] have devised a *Gaussian* SBC to automatically determine the number of topics, thus trying to achieve the same results as a fully-fledged DP. The setting seems to perform well. However, it needs two RNNs on the encoder side (the first to learn the SBC weights and the second to bind the number of topics), and it is still *fully Gaussian*, hence not Dirichlet-related. *To our knowledge*, the first work to involve DPs in the strict sense with VAEs is Nalisnick & al.'s stick-breaking VAEs (SB-VAE) [NS17]. The work does not particularly focus on topic modeling tasks and is actually of a general extent. Due to the reparameterization trick, Nalisnick & al. have replaced the original Beta distribution with a Kumaraswamy distribution. Relying on SB-VAEs, Ning & al. [Nin+20] have devised unsupervised VAE-based topic models. Still, there is no notion of word linking within these works.

### 2.2.2 Dynamic topic modeling

One of LDA's central hypotheses is that of exchangeability between documents. In some cases, the order of the documents is essential, and thus, they are not exchangeable. Blei & al.'s Dynamic Topic Models [BL06] is perhaps the most famous work in the field. The authors build a temporal extension for LDA (Dynamic LDA or D-LDA) that includes a chain with each topic parameter embedded into a state-space model that changes with Gaussian noise. The chain's role is to ensure proper document linking. In addition to this chaining, the document-topic prior is no longer a Dirichlet distribution but a Logistic Normal. DTM is much more a framework than a model. It focuses on building a model with implicit time-

3. In topic modeling, the topics are these components.
4. We provide more details on DPs in Section 3.

dependency inclusion that captures dependencies for both document-topic and word-topic distributions. A recent application inspired by the framework is the Dynamic Embedded Topic Model (D-ETM) [DRB19]. The D-ETM extends the ETM to a dynamic context and shares a similar inference engine.

The Dynamic Mixture Model (DMM) [WSW07] is a variant of DTM. Instead of chaining priors on topics, DMM chains the topic themselves and drops word-topic dependencies. The impact of this difference is that DMM rendering is about consecutive documents instead of temporal-grouped slices of a corpus that posits the exchangeability of its members. In other words, we could consider that DTM processes streams of batches of documents and that DMM processes streams of individual documents, thus making it sound akin to an online model. The DTM or the DMM model time by passing parameters from one time slice to the following, similar to a prior. This process is an implicit, Markovian linking. Other authors have chosen explicit modeling of the time dependency, resulting in a much more orthodox Bayesian approach. For instance, in Topics Over Time (TOT) [WM06], the authors build an extension of the LDA, where a document's timestamp is associated with its tokens. Besides capturing the underlying data structure of the documents, the models also render how it changes over time in both short and long terms in a non-Markovian way. TOT assumes that topics are associated with a continuous distribution over timestamps. The mixture distribution of topics for each topic, in turn, has influences from both the observable word co-occurrences and document timestamp variables.

As our work aims at handling a massive preexistent dataset coming from several sources, it is impossible to anticipate how many topics the dataset will contain. Under our assumptions, including non-parametric aspects to topic models make sense to keep track of topic evolution. Non-parametric topic models find their premises in Teh al.'s Hierarchical Dirichlet Processes (HDP) [Teh+06]. Their stick-breaking approach has become popular and fuels the inference of many topic models, including LDA. The non-parametric version of LDA is the hierarchical LDA (HLDA) [Ble+03]. It uses an arborescent process based on the Chinese Restaurant Process (CRP) due to an analogy between its construction and Chinese restaurant customers. Wang & al. [WPB11a] have devised an online variational algorithm to make this model suitable for large datasets. Despite not having any temporal dynamic, the model still opened a way toward non-parametric, efficient models. Ahmed & Xing [AX10] have used the CRP to build their infinite dynamic topic model (iDTM), an extension of DTM.

Similarly to its parametric counterpart, iDTM posits that documents are exchangeable within the same epoch. However, they have adapted the evolution of per-document and per-word topics to distribution into a Chinese Restaurant Franchise (CRF) representation. The CRF enables the activation and deactivation of topics at any epoch. By attributing a recurrent twist to the CRF, iDTM can also capture dependencies between the topics and the popularity of each epoch. Despite these features, DTM, and subsequently, iDTM, need a discretization of time. The granularity impacts the exchangeability assumption of documents within a time slice. If the time slices are too extensive, then the temporal equivalence between documents can barely hold; on the contrary, if the time slices are too shallow, the number of variational parameters will explode.

In 2008, Ren & al. [RDC08] presented the dynamic Hierarchical Dirichlet Processes (dHDP). The model directly extends the HDP. As a Bayesian hierarchical model, dHDP posits some dependency between groups or topics. However, the model assumes exchangeability for topics corresponding to the same time slice. Wang & al. [WBH08] have developed the continuous-time dynamic topic model (cTDTM) with this criticism in mind. It is still a variant of DTM, except that it uses Brownian motion to model the latent topics through a sequential collection of documents. Thanks to this setting, cTDTM can handle arbitrary granularity, as demonstrated experimentally. Finally, the CRF applies when the temporal dimension is explicit, such as non-parametric TOT [Dub+12]. As common with classical Bayesian statistics, most models use Gibbs sampling as their inference workhorse. Despite the theoretical guarantees it offers, Gibbs sampling in general is too slow for document stream processing, thus making variational techniques preferable for this task.
In this Section, we have presented an overview of state-of-the-art of dynamic topic modeling. While time is an essential element in understanding the underlying structure of a corpus, so are contextual elements. The next Section presents an overview of the state-of-the-art of contextual topic modeling.

### 2.2.3 Contextual topic modeling

The LDA is a generative model that tries to capture topics and word co-occurrence in a corpus. However, it comes to a price, as elements in a document are conditionally independent. This exchangeability makes inference easier, but it decontextualizes a word by ignoring its vicinity. The same applies to topics, as the LDA does not capture their proximity. Context is essential to knowledge discovery for textual data [BTH21; Bia+21], as their goal is to make collections of documents interpretable to analysts without reading each of them. It is also crucial to dimensionality reduction as it enables summarizing a text to its essential features.

Several authors have tried to introduce context and proximity in a topic model. For instance, Blei & Lafferty [LB05] have tried to leverage topic correlations in their Correlated Topic Models (CTM). Du & al. [DBJ10]) have devised Sequential LDA, a topic model that makes use of the segments in a corpus (chapters, paragraphs, Etc.) to render its underlying topics. Hu & al. [HBS11] even suggested that humans can correct topic models by adding a linking constraint between words in an interactive mode. According to Zhu & al. [ZBL06], syntactic elements act as scaffolding without dropping the exchangeability assumptions. None of these approaches apply to our case; we focus on modeling a massive volume of documents from several sources. We want to extract information when we have no prior knowledge of a specific domain that often needs particular expertise or tools. Depending on additional external tools that strongly depend on external choices and data quality is risky and might hinder a model from producing quality topics. Additionally, our goal is not to build models that reproduce an expert's knowledge; it is much more to build techniques that faithfully report the datasets at hand. The way we consider it, the fidelity must depend on the fact that language and text are a sequence of words and symbols that follow a specific order. Finally, we mainly focus on leveraging fully unsupervised word semantics in topic modeling, as it is one of the most basic observable variables, the other being time.

Capturing word dependencies is achievable in several ways. The most direct way is to leverage the LDA word-topic priors by weighting terms that we want to appear together. However, similarly to Hu & al.'s interactive solution [HBS11], there is no guarantee that the user's modifications correspond to what is inside the dataset, even if the weighting relied on a thorough expert diagnosis. It is merely a prior. Yan & al. introduced the Biterm topic model (BTM) [Yan+13] in a much more unsupervised way. It treats words in pairs to introduce context. The finality is to better model short documents, one of LDA's identified pitfalls. The experiments on the Tweets2011 Collection [5] and the 20 Newsgroup [6] datasets show that BTM outperforms LDA on short and "normal" text settings. However, the approach is highly impractical in our industrial context due to the need to form bi-terms, thus considerably increasing the volume of data. In the same vein, Balikas & al. [Bal+16] have designed the senLDA. In this work, the authors assume that the terms occurring within the same sentence come from the same topic. The method is generalizable to longer or shorter texts, thus making the LDA a particular case of senLDA. LDA performs better on perplexity evaluation, whereas senLDA is better for classification tasks and converges faster. This kind of linking still assumes that elements are somewhat exchangeable, even if the considered text spans are variable in length.

Despite being tools primarily aiming at knowledge discovery, topic models are also perceivable as language models. Some language models take word order into account to reproduce language effectively. To our knowledge, the first Markovian topic model for language modeling is Griffiths & al.'s HMM-LDA [Gri+04]. This model has the particularity of considering both "short-range syntactic dependencies and long-range semantic dependencies" instead of solely focusing on syntax and semantics. In other words, the model integrates word meanings and order to extract faithful topics. According to the authors, only a subset of words – the content words – will produce long-range semantic dependencies, yet they depend on local dependencies – syntax – as all words. Syntax also carries contextual elements. The model yields good results on several tasks, such as topic extraction and POS

5. https://trec.nist.gov/data/tweets/
6. http://qwone.com/~jason/20Newsgroups/

tagging. It behaves similarly to LDA on document classification on the Brown corpus [7]. HMM-LDA has a neural alternative named TopicRNN [Die+17]. The underlying ideas of this model are the same and combine a topic modal and a recurrent neural network (RNN) for topic extraction and sentiment analysis. TopicRNN, however, does not model the topics explicitly. The RNN encodes the topic extraction information in its hidden states instead.

Another approach relies on word embeddings. For recall, word embeddings are dense vectorial representations of words obtained from a bag of words representation of documents. These vectors not only encode words; they also encode their context. One of the most famous approaches is Word2Vec [Mik+13]; other authors have released other representation techniques since Word2Vec, yet they remain trendy and used in various works and industrial applications. Dieng & al. have successfully adapted word embeddings to LDA-like topic modeling in their Embedded Topic Model [DRB20]. The model not only learns topic representations (or contexts); they are also capable of learning word representations (or embeddings). As with other embedding techniques, it is possible to use pre-trained embeddings and to learn topics only. The model considers the word vicinity only: the model is still non-Markovian. However, the model explicitly captures both the topics and the elements. Despite the model's performance on topic extraction tasks, it is not usable for other tasks that require word sequence reporting, such as POS tagging. Other methods use pre-trained embeddings [DZD15; Xun+17; HBS11]. The process even works in nonparametric settings, as reported in the *spherical Hierarchical Dirichlet Process* [Bat+16]. We present a few applications in the following Section.

7. `http://korpus.uib.no/icame/brown/bcm.html`

## 2.3 Applications to media mining

Topic models uses have been reviewed in both the academic [BHM17; Jel+19] and the industrial [VJ21; Xio+18] contexts, including marketing [GS21] and social media research. Our work is not the first to involve web-collected data. Most of the previous works we noticed use a single data source [Qia+22] - Twitter or Reddit, for instance -, thus allowing for specific data cleaning processes and standardization steps [8], even when involving web-scraping[Arm+21; Mau+21]. Some works also explicitly involved collection through APIs, and as such, no data parsing nor markup languages removal was necessary[El +21; HP21]. *As far as we know*, applications of neural-based topic models almost exclusively concern health issues (mental health and COVID-19). We believe this interest is, first and foremost, due to the closeness between the moment they were released and the various health issues caused by the pandemia. Contrarily to our use case, the applications we listed do not particularly focus on specific named entities such as products. In one of them, [HP21] topic modeling only serves to reduce text dimension for classification without any other concern for the extracted topics. The work we find closest to ours is Bennett & al.'s [BMT21]. The authors compared state-of-the-art neural and non-neural topic models and applied these same models to COVID-19 Twitter data. Contrasting with our approach, they emphasized parametric neural topic models, while our work focuses on applying non-parametric ones. Like all the other works, they did not try to use the model's properties to cope with data-inherent noise. Last but not least, they introduced a novel regularization term, which is something beyond the scope of this work.

## 2.4 Conclusion

In this Chapter, we have presented the Bayesian foundations of probabilistic graphical modeling. We have shown its links with probabilistic topic modeling and deep generative learning. We have also presented state-of-the-art topic modeling and some applications for media mining. In the next Chapter, we present the theoretical framework we derivate from this knowledge.

8. E.g., setting words to lowercase or punctuation removal.

# Chapter 3

# The Embedded Dirichlet Processes

In this Chapter, we present two contributions to solving the problem of large-scale, unsupervised topic extraction in social media: the Embedded Dirichlet Process and the Embedded Hierarchical Dirichlet Process. These models can automatically detect the number of topics in a corpus and compute word embeddings and topic embeddings for better text exploration. These approaches do not need prior learning in the sense of transfer learning to reach these goals. We start by laying the theoretical foundations for building our two models, i.e., the Embedded Topic Model and the Dirichlet Processes. The Embedded Topic Model is a neural-based topic model that presents the particularity of capturing semantic links in long-tail vocabularies. This property makes the model particularly interesting for Web-extracted, heterogeneous, and noisy corpora, where the writers often use Internet slang or misspell words. It also captures links between topics, thus considering that specific topics can exhibit some degree of relatedness. Dirichlet Processes, in turn, are particular probability distributions that determine latent variables' number of components or topics. We focus on the stick-breaking variant of the Dirichlet Processes and show how they relate to Deep Learning. We then test our models on two standardized datasets before applying them to an industrial dataset proceeding from several social media sources. We show that they perform better than their concurrent counterparts in empirical comparative studies.

## 3.1 Related work

The following sections present the theoretical foundations for our Embedded Dirichlet Processes. These foundations are two-fold, as we mainly rely on the Embedded Topic Model and the Dirichlet Processes. Last, we present an alternative to the RBS that enables sampling from Gamma and Gamma-derivated distributions instead of using surrogates.

### 3.1.1 The Embedded Topic Model

The Embedded Topic Model (ETM) [DRB20] extends the Latent Dirichlet Allocation [BNJ03] to include Continuous Bag-Of-Words-like (CBOW) embeddings [Mik+13] in its generative process. The algorithm learns word embeddings from data using entire document contexts (or topics) instead of surrounding words. The contexts themselves are topic embeddings that are visualizable in the same space as the word embeddings. It is also possible - yet optional - to initialize the embedding layers with pre-fitted and more complex word representations, including words that do not appear in the dataset. In this setting, the model will still fit representations for these words according to their lexical vicinity. The ETM shares LDA's mixture assumptions, except that words and topics can show similarities in their embedding space in contrast with the latter. It can also capture the distribution of rare words and the long tail of language data. This property proves useful in heterogeneous datasets. ETM uses a VAE setting [KW14] for parameter inference. The authors use a Logistic Normal prior as a surrogate distribution for a Dirichlet (Fig. 3.4) to work with the reparameterization trick.

**Algorithm 2** Generative process for the ETM

```
1: Choose N_d ~ Poisson(λ)
2: Draw topic proportions θ_d ~ LN(0, I)
3: for all word w_n in document d do
4:     Draw topic assignment z_dn ~ Categorical(θ_d)
5:     Draw the word w_dn ~ softmax(ρ^⊤ α_{z_dn})
6: end for
```

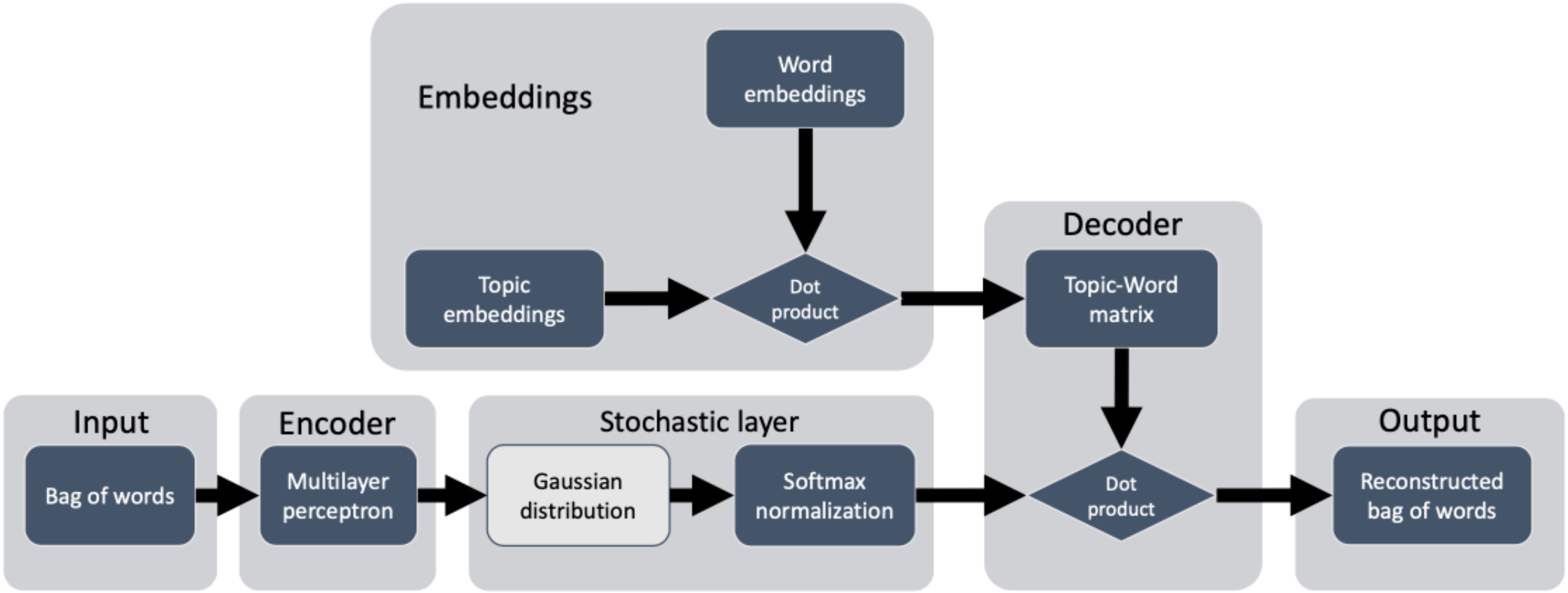

Figure 3.1 – Simplified representation of the ETM

In Alg. 2, $\theta_d = \text{softmax}(\delta_d)$ where $\delta_d \sim \mathcal{N}(0, I)$, $\rho$ is a $L \times V$ tensor, and $\alpha$ is a $L \times K$ tensor. $L$ is a fixed embedding dimension, $V$ is the vocabulary length, and $K$ is a hyperparameter for the number of topics. Its ELBO is the following:

$$\text{ELBO}(\nu) = \mathbb{E}_q [\log \Pr(\mathbf{w} \mid \delta, \rho, \alpha)] - \text{KLD}(q(\delta; \mathbf{w} \mid \nu) \parallel \Pr(\delta)) \tag{3.1}$$

In Eqn. 3.1 [1], $\mathbf{w}$ is the document set, $\nu$ represents the weight coefficients of a MultiLayer Perceptron (MLP) that acts as an encoder - or inference network - for a Gaussian variational distribution parameterized with $\mu$ and $\Sigma$. Applying the softmax function normalizes the samples from this Gaussian and embeds them on the simplex. Finally, the optimization process uses the Adam optimizer (Alg. 3).

1. We disclose the formula details in Appendix A.

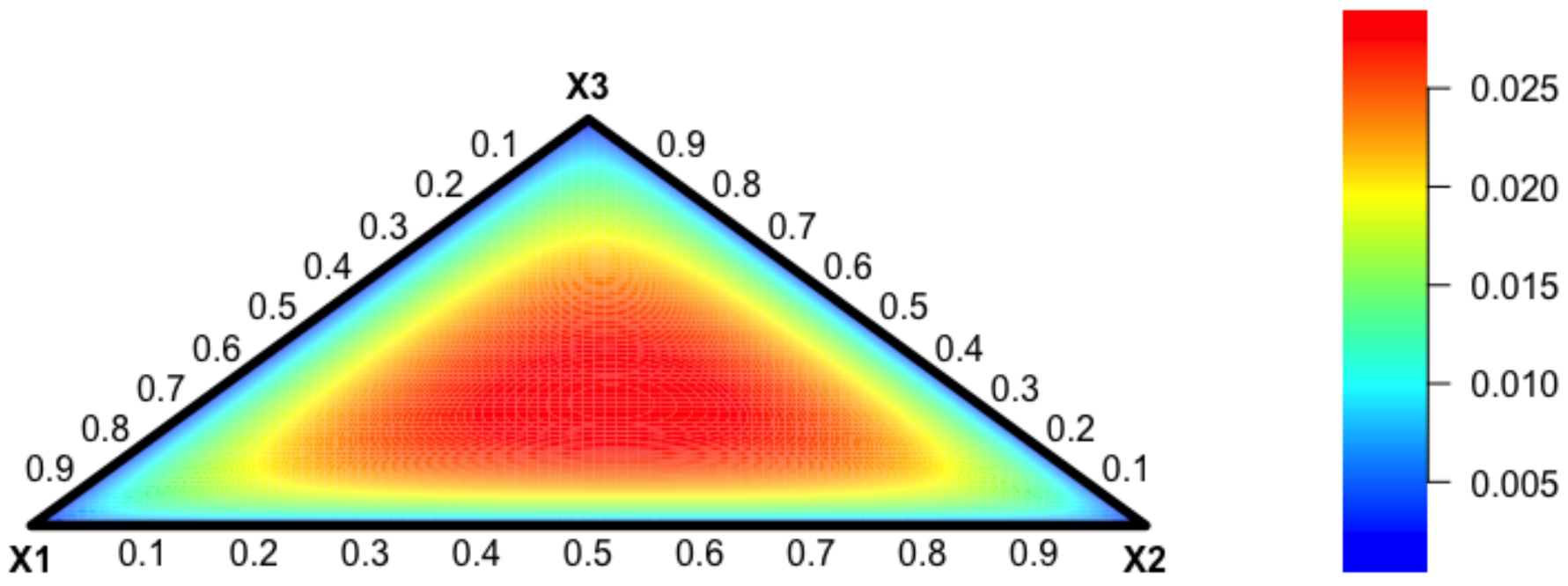

Figure 3.2 – Contour plot of a Dirichlet distribution

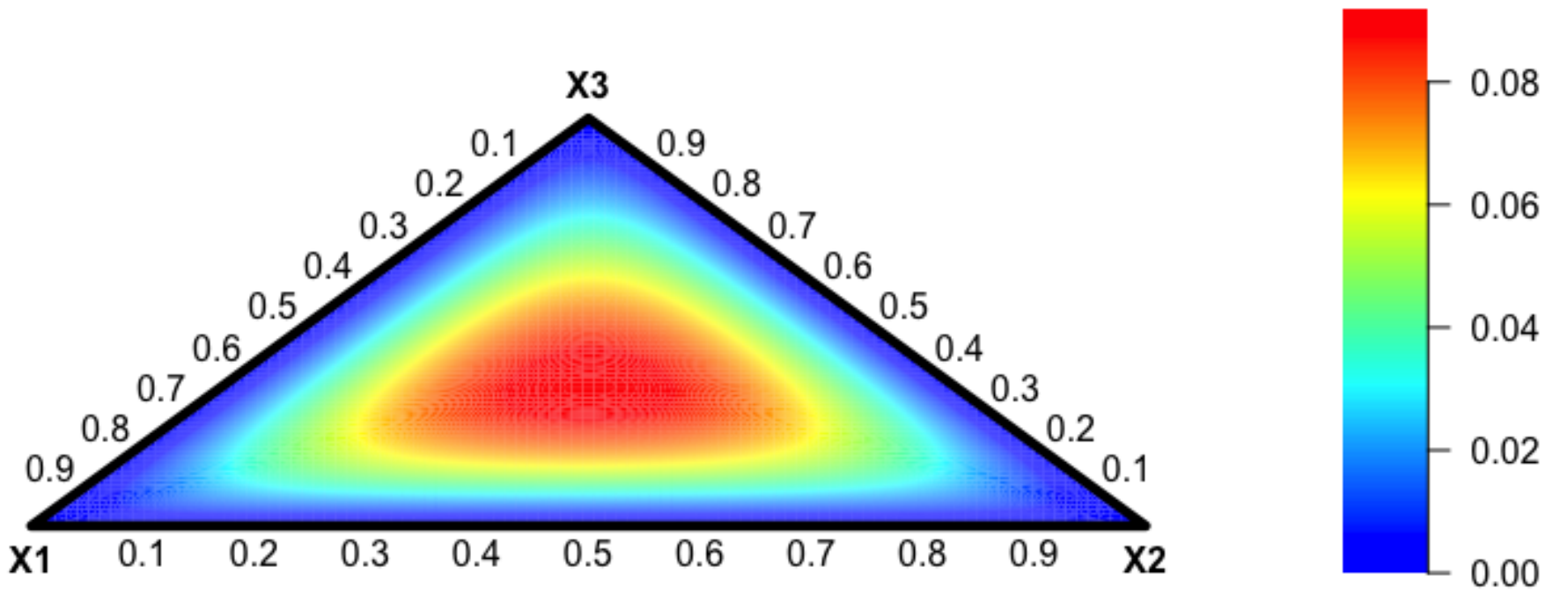

Figure 3.3 – Contour plot of a logit-normal distribution

Figure 3.4 – Densities of the Dirichlet and the logit-normal distributions

**Algorithm 3** Inference process for the ETM

Initialize the model and its variational parameters
**for** $i \leftarrow 1$ to maximum number of iterations **do**
  Compute $\beta_k = \text{softmax}(\rho^\top \alpha_k)$ for each topic $k$
  Choose a minibatch $\mathcal{B}$ of documents
  **for** each document $d$ in $\mathcal{B}$ **do**
    Get a bag of words representation $\mathbf{x}_d$
    Compute $\mu_d = \text{MLP}\left(\mathbf{x}_d; \nu_\mu\right)$
    Compute $\Sigma_d = \text{MLP}\left(\mathbf{x}_d; \nu_\Sigma\right)$
    Sample $\theta_d \sim \mathcal{LN}\left(\mu_d, \Sigma_d\right)$
    **for all** word $w$ in document $d$ **do**
      Compute $\Pr\left(w_{dn} \mid \theta_d\right) = \theta_d^\top \beta_{\cdot, w_{dn}}$
    **end for**
  **end for**
**end for**
Estimate the ELBO and its gradient through backpropagation
Update model parameters $\alpha_{1:K}$
Update variational parameters $(\nu_\mu, \nu_\Sigma)$

### 3.1.2 Dirichlet processes and neural variational inference

In this section, we present the Dirichlet processes. The Dirichlet processes are a classical Bayesian tool for determining the number of underlying components in a dataset [Teh+06; Ble+03]. The Dirichlet processes typically replace a Dirichlet prior in topic models that infer the number of topics from data. We then move on to presenting how to use Dirichlet processes with neural variational inference.

#### 3.1.2.1 Dirichlet processes and stick-breaking construction

Let $(\Theta, \mathcal{B})$ be a measurable space, with $G_0$ a probability measure defined on that space. Let $\alpha_0$ be a positive real number. A Dirichlet process $\text{DP}(\alpha_0, G_0)$ is a probability distribution of a random probability measure G over $(\Theta, \mathcal{B})$ such that, for any finite measurable partition $(A_1, A_2, \cdots, A_r)$ of $\Theta$, the random vector $(\text{G}(A_1), \text{G}(A_2), \cdots, \text{G}(A_r))$ is distributed as a finite-dimensional Dirichlet distribution with parameters $(\alpha_0 \text{G}_0(A_1), \alpha_0 \text{G}_0(A_2), \cdots, \alpha_0 \text{G}_0(A_r))$. We write $G \sim \text{DP}(\alpha_0, G_0)$ is G is a random probability measure with distribution given by the Dirichlet process. Three different ways of constructing a Dirichlet process exist, each corresponding to a metaphor: the Pólya urn model, the Chinese restaurant process, and the stick-breaking construction. Variational inference involving Dirichlet processes often uses the latter.

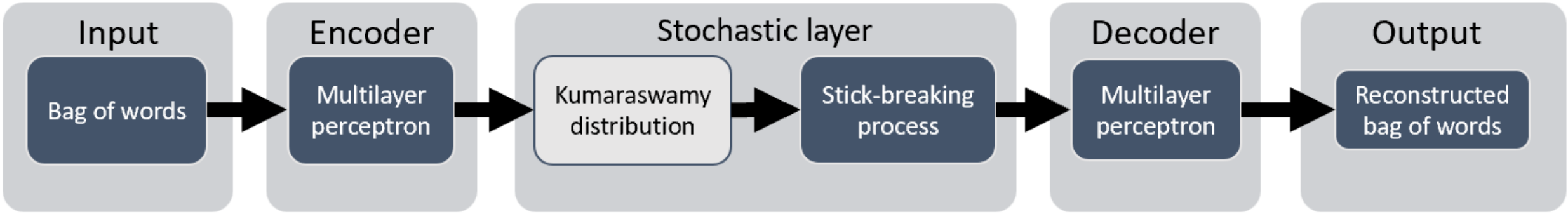

Figure 3.5 – Simplified representation of the SB-VAE

Sethuraman introduced the stick-breaking construction for Dirichlet processes in 1994 [Set94]. It uses independent sequences of i.i.d. random variables $(\pi'_k)_{k=1}^{\infty}$ and $(\phi_k)_{k=1}^{\infty}$:

$$\pi'_k \,|\alpha_0, G_0 \sim \text{Beta}\,(1, \alpha_0) \quad \phi_k|\,\alpha_0, G_0 \sim G_0 \tag{3.2}$$

We then define a random measure $G$ as follows:

$$\pi_k = \pi'_k \prod_{l=1}^{k-1} (1 - \pi'_l) \quad G = \sum_{k=1}^{\infty} \pi_k \delta_{\phi_k} \tag{3.3}$$

In Eqn. 3.3, $\delta_\phi$ is a probability measure concentrated at $\phi$, e.g., a Dirac. Additionally, $G$ is a random probability measure distributed according to $\text{DP}(\alpha_0, G_0)$. The sequence $\pi = (\pi_k)_{k=1}^{\infty}$ satisfies $\sum_{k=1}^{\infty} \pi_k = 1$. As $\text{Beta}(1, \alpha_0)$ is equivalent to $\text{GEM}(\alpha_0)$, we can write $\boldsymbol{\pi'} \sim \text{GEM}\,(\alpha_0)$ and $\boldsymbol{\pi'} \sim \text{Beta}\,(1, \alpha_0)$ interchangeably.

#### 3.1.2.2 Stick-breaking variational autoencoder

The stick-breaking VAE (SB-VAE) is an adaptation of the SGVB framework for Bayesian nonparametric processes [NS17]. More particularly, it aims to learn the stick-breaking process' weights through posterior inference. Despite being of a general extent, the SB-VAE algorithm is also applicable to topic modeling [Nin+20].
Let $\mathbf{w} = \{\mathbf{w}_1, \ldots, \mathbf{w}_d\}$ be a corpus of $D$ documents, where $\mathbf{w}_d$ is a collection of $N_d$ words.

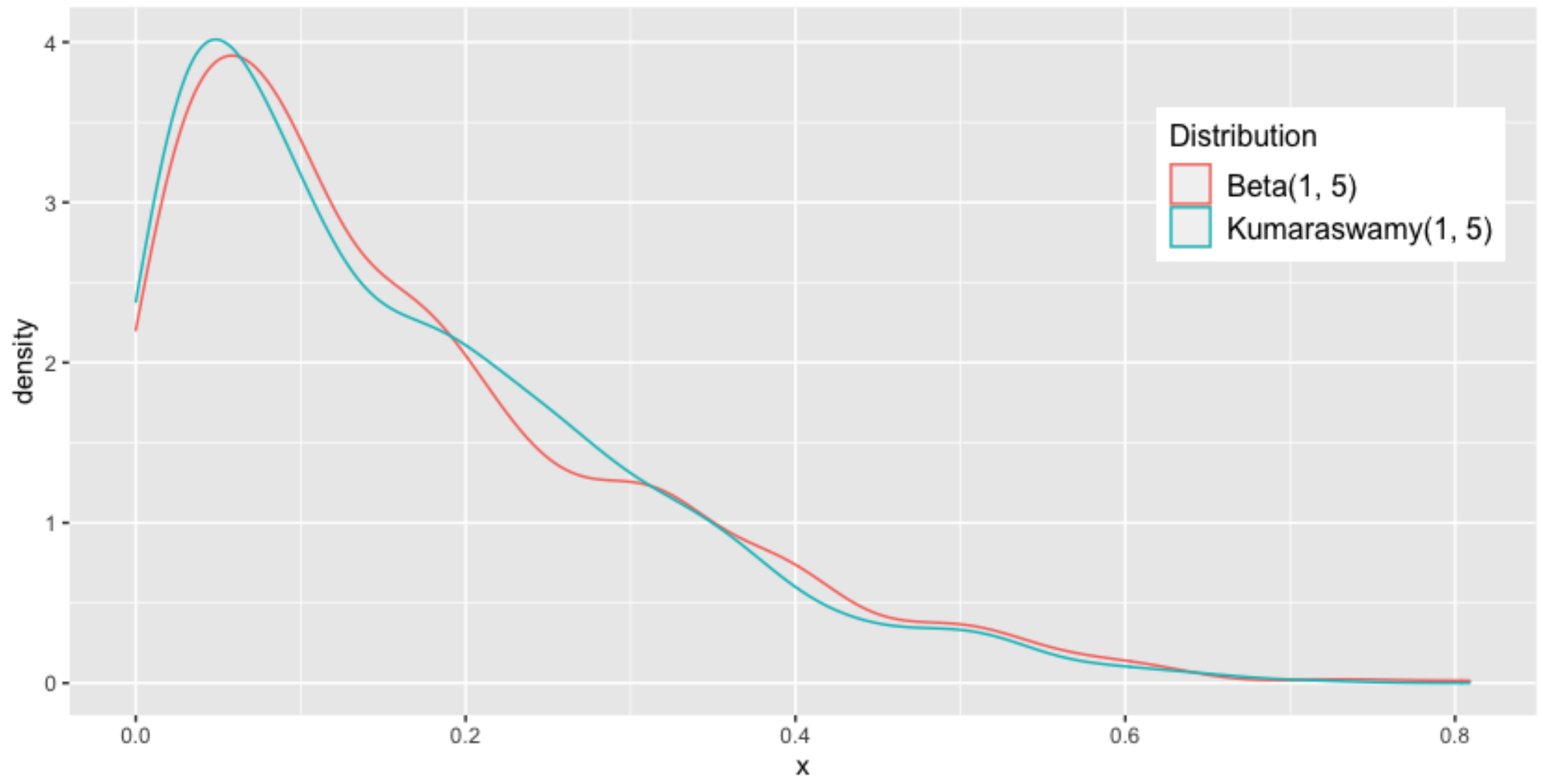

Figure 3.6 – Samples from the densities of the Beta and the Kumaraswamy distributions

Each document representation is a bag of words $\mathbf{w}_d$.

**Algorithm 4** Generative process for the SB-VAE

1: Choose $N_d \sim \text{Poisson}(\lambda)$
2: Get a document-specific $G^{(d)}\left(\theta; \pi^{(d)}, \Theta\right) = \sum_{k=1}^{\infty} \pi_k^{(d)} \delta_{\theta_k}(\boldsymbol{\theta})$, with $\pi^{(d)} \sim \text{GEM}(\alpha_0)$
3: **for all** word $w_{dn}$ in the document **do**
4: Draw a topic $\hat{\theta}_{dn} \sim G^{(d)}\left(\theta; \pi^{(d)}, \Theta\right)$
5: Draw a word $w_{dn} \sim \text{Categorical}\left(\hat{\theta}_{dn}\right)$
6: **end for**

In Alg. 4, $q_\psi(\cdot)$ denotes the family of variational distributions, $\psi$ denotes the neural network parameters, and $a$ and $b$ are the variational parameters learned by a MultiLayer Perceptron (MLP), and $v$ denotes the weights for the stick-breaking step. Contrarily to Ishwaran & James [IJ01], it is not possible to use a variational Beta distribution for stick-breaking with the RBS from Kingma & Welling's work [KW14]. The RBS requires a differentiable non-centered parameterization that the Beta distribution cannot provide. Consequently, Nalisnick & al. replaced the Beta variational with a Kumaraswamy distribution [Kum80].

Although with higher entropy, this distribution is proper as a surrogate for a Beta distribution and has a DNCP (Fig. 3.6). It is a two-parameter continuous distribution on a unit interval whose density function is the following:

$$\text{Kumaraswamy}(x; a, b) = abx^{a-1}\left(1 - x^a\right)^{b-1} \tag{3.4}$$

where $x \in (0, 1)$ and $a, b > 0$. It is possible to obtain the required samples from the Kumaraswamy distribution as follows:

$$x \sim \left(1 - u^{\frac{1}{b}}\right)^{\frac{1}{a}} \text{ where } u \sim \text{Uniform}(0, 1) \tag{3.5}$$

The KLD from the optimization objective is, thus, between a Kumaraswamy distribution and a Beta distribution. Its formulation is the following:

$$\text{ELBO}(\nu) = \mathbb{E}_q\left[\log \Pr\left(\mathbf{w} \mid \pi\right)\right] - \text{KLD}\left(q\left(\pi \mid \mathbf{w}, \nu\right) \parallel \Pr\left(\pi\right)\right) \tag{3.6}$$

In Eqn. 3.6 [2], $\mathbf{w}$ is the document set. We use *amortized variational inference* and Adam to fit the model concerning all the parameters (Alg. 5).

2. We disclose the formula details in Appendix A.

**Algorithm 5** Inference process for the SB-VAE

Initialize the model and its variational parameters
**for** $i \leftarrow 1$ to maximum number of iterations **do**
  Compute $\beta_k = \text{softmax}(\rho^\top \phi_k)$ for each topic $k$
  Choose a minibatch $\mathcal{B}$ of documents
  **for** each document $d$ in $\mathcal{B}$ **do**
    Get a bag of words representation $\mathbf{x}_d$
    Compute $a = \text{MLP}\left(\mathbf{x}_d; \psi_a\right)$
    Compute $b = \text{MLP}\left(\mathbf{x}_d; \psi_b\right)$
    Sample $v \sim \text{Kumaraswamy}\left(a, b\right)$
    Compute $\pi = \begin{cases} v_1 \text{ if } k = 1 \\ v_k \Pi_{j<k}\left(1 - v_j\right) \text{ for } k > 1 \end{cases}$
    **for all** word $w_d$ in the document **do**
      Compute $\Pr\left(w_{dn} \mid \pi\right) = \text{softmax}\left(\text{MLP}(\pi)\right)$
    **end for**
  **end for**
**end for**
Estimate the ELBO and its gradient through backprogation
Update the parameters

### 3.1.3 Implicit reparameterization gradients

Machine Learning and Statistics have made extensive use so far of pathwise gradient estimators. In the context of Machine Learning, pathwise gradient estimators are usually the engine under the hood of the widely known reparameterization trick - or reparameterization by standardization (RBS) in this research. On the one hand, RBS makes backpropagation through stochastic nodes possible in a variety of contexts, including that of VAEs [KW14]. On the other hand, it only works with distributions with location-scale parameterization, tractable inverse cumulative distribution functions (CDF), or distributions that are expressible through deterministic transformations. These conditions exclude distributions, including the Gamma, the Beta, and the Dirichlet. As seen in [DRB20], [SS17], or in [NS17], it is possible to use surrogate distributions. These distributions, however, only sometimes exhibit all the good properties needed to model data. In particular, surrogate distributions can struggle to capture sparsity [RTB16], which is paramount to efficient topic modeling. Figurnov & al. introduced an alternative to RBS in [FMM18] they call the *implicit reparameterization gradients*. The implicit reparameterization gradients (IRG) allegedly provide unbiased estimators for continuous distributions whose CDF is numerically tractable, faster, and more accurate and enable using the Gamma, Beta, and Dirichlet distributions, among others. They also present applications of the IRG in the context of VAEs, thus making them fully compatible with our desiderata.

In the context of RBS, let $\mathbb{E}_{q_\phi(\boldsymbol{z})}[f(\boldsymbol{z})]$ be an expectation of some continuously differentiable function $f(\boldsymbol{z})$ with respect to a set of distribution parameters $\phi$. Suppose we want to optimize this expectation. We assume the existence of a standardization function $\mathcal{S}_\phi(\boldsymbol{z})$ that removes the dependence on the set of distribution parameters when applied to a sample $q_\phi(\boldsymbol{z})$. This standardization function presents two essential characteristics: it is invertible and continuously differentiable with respect to both its arguments and the set of parameters:

$$\mathcal{S}_\phi(\boldsymbol{z}) = \boldsymbol{\varepsilon} \sim q(\boldsymbol{\varepsilon}) \quad \boldsymbol{z} = \mathcal{S}_\phi^{-1}(\varepsilon) \tag{3.7}$$

We can express the objective as an expectation with respect to $\boldsymbol{\varepsilon}$, thus transferring the dependence on $\phi$ into $f$:

$$\mathbb{E}_{q_\phi(\boldsymbol{z})}[f(\boldsymbol{z})] = \mathbb{E}_{q(\boldsymbol{\varepsilon})}\left[f\left(\mathcal{S}_\phi^{-1}(\boldsymbol{\varepsilon})\right)\right] \tag{3.8}$$

This transfer allows us to compute the gradient of the expectation as the expectation of the gradients:

$$\nabla_\phi \mathbb{E}_{q_\phi(\boldsymbol{z})}[f(\boldsymbol{z})] = \mathbb{E}_{q(\boldsymbol{\varepsilon})}\left[\nabla_\phi f\left(\mathcal{S}_\phi^{-1}(\boldsymbol{\varepsilon})\right)\right] = \mathbb{E}_{q(\boldsymbol{\varepsilon})}\left[\nabla_{\boldsymbol{z}} f\left(\mathcal{S}_\phi^{-1}(\boldsymbol{\varepsilon})\right)\nabla_\phi \mathcal{S}_\phi^{-1}(\boldsymbol{\varepsilon})\right] \tag{3.9}$$

To achieve the IRG, the authors start from Eqn. 3.9. The first idea is to avoid the inversion of the standardization function. They perform a *change of variable* such as $z = \mathcal{S}_\phi^{-1}(\varepsilon)$, thus modifying the expression:

$$\nabla_{\boldsymbol{\phi}} \mathbb{E}_{q_\phi(\boldsymbol{z})}[f(\boldsymbol{z})] = \mathbb{E}_{q_\phi(\boldsymbol{z})}\left[\nabla_{\boldsymbol{z}} f(\boldsymbol{z}) \nabla_{\boldsymbol{\phi}} \boldsymbol{z}\right]; \quad \nabla_\phi \boldsymbol{z} = \nabla_\phi \mathcal{S}_\phi^{-1}(\boldsymbol{\varepsilon})\big|_{\boldsymbol{\varepsilon}=\mathcal{S}_\phi(\boldsymbol{z})} \tag{3.10}$$

The second idea is to leverage *implicit differentiation* to compute $\nabla_\phi \mathbf{z}$. They reportedly applied the total gradient $\nabla_{\boldsymbol{\phi}}^{\mathrm{TD}}$ to $\mathcal{S}_\phi(\boldsymbol{z}) = \boldsymbol{\varepsilon}$. Thanks to the chain rule, they expand the total gradient in terms of the partial gradient, then make the standardization function depend on the set of parameters $\phi$ and its argument $\mathbf{z}$. By definition, the noise $\varepsilon$ is independent of $\phi$. Consequently, and considering $\nabla_\phi z$, the authors had to solve the equation $\nabla_{\boldsymbol{z}} \mathcal{S}_\phi(\boldsymbol{z}) \nabla_\phi \boldsymbol{z} +$

$\nabla_\phi \mathcal{S}_\phi(\boldsymbol{z}) = \mathbf{0}$. The result is the following:

$$\nabla_\phi \boldsymbol{z} = -\left(\nabla_{\boldsymbol{z}} \mathcal{S}_\phi(\boldsymbol{z})\right)^{-1} \nabla_\phi \mathcal{S}_\phi(\boldsymbol{z}) \tag{3.11}$$

In Eqn. 3.11, the standardization function no longer needs inversion but differentiation. The IRG applies to the Gamma distribution. The Gamma distribution takes two parameters in its shape-rate expression: a shape parameter $\alpha > 0$ and a rate parameter $\beta > 0$. If $z \sim \text{Gamma}(\alpha, 1)$, then $z/\beta \sim \text{Gamma}(\alpha, \beta)$. It is possible to build the Beta and Dirichlet distributions with Gamma samples. We first consider the Beta distribution. If $z_1 \sim \text{Gamma}(\alpha, 1)$ and $z_2 \sim \text{Gamma}(\beta, 1)$, then $\frac{z_1}{z_1+z_2} \sim \text{Beta}(\alpha, \beta)$. On the other hand, if $z_i \sim \text{Gamma}(\alpha_i, 1)$, then $(\frac{z_1}{\sum_{j=1}^{D} z_j}), \ldots, \frac{z_D}{\sum_{j=1}^{D} z_j} \sim \text{Dirichlet}(\alpha_1, \ldots, \alpha_D)$.

## 3.2 The models

This section presents two novel models: the Embedded Dirichlet Process and the Embedded Hierarchical Dirichlet Process.

### 3.2.1 The Embedded Dirichlet Process

The Embedded Dirichlet Process (EDP) is a VAE-based model. Unlike common VAEs, its core distributions are not Gaussian; it uses a Dirichlet Process instead. It leverages the implicit reparameterization trick to sample from a Beta variational distribution instead of a Kumaraswamy.

Let $\{\mathbf{w}_1, \ldots, \mathbf{w}_D\}$ be a corpus of $D$ documents, where $\mathbf{w}_D$ is a collection of $N_D$ words. Each document representation is a bag of words $\mathbf{w}_D$.

**Algorithm 6** Generative process for the EDP

1: Choose $N_d \sim \text{Poisson}(\lambda)$
2: Get a document-specific $G^{(d)}\left(\theta; \pi^{(d)}, \Theta\right) = \sum_{k=1}^{\infty} \pi_k^{(d)} \delta_{\theta_k}(\boldsymbol{\theta})$, with $\pi^{(d)} \sim \text{GEM}(\beta)$
3: **for all** word $w_{dn}$ in document $d$ **do**
4: Draw a topic $\hat{\theta}_d \sim G^{(d)}\left(\theta; \pi^{(d)}, \Theta\right)$
5: Draw a word $w_{dn} \sim \text{softmax}\left(\rho^T \phi\right)$
6: **end for**

The EDP decomposes the word level in a dot product between the (transposed) word embeddings $\rho$ and the context embeddings $\phi$ (line 5 from Alg. 6). As this decomposition forms a log-linear model, the word embeddings and the topic embeddings evolve in the same space, thus making it possible to compare words' and topics' positions. These properties make the EDP a tool for deeper textual exploration than a classical Dirichlet Process-based topic model. The model's joint distribution is the following:

$$\Pr\left(\mathbf{w}, \pi, \hat{\theta} \mid \beta, \Theta, \xi\right) = \Pr(\pi \mid \beta) \times \Pi_{d=1}^{D} \Pr\left(\mathbf{w}_d \mid \hat{\theta}_d, \xi\right) \Pr\left(\hat{\theta}_d \mid \pi, \Theta\right) \tag{3.12}$$

where $\Pr(\pi \mid \beta) = \text{GEM}(\beta)$, $\Pr\left(\hat{\theta} \mid \pi, \Theta\right) = G(\theta; \pi, \Theta)$, $\Pr\left(\mathbf{w} \mid \hat{\theta}, \xi\right) = \text{softmax}\left(\hat{\theta}\xi\right)$, and $\xi = \text{softmax}\left(\rho^T \phi\right)$. We use a family of variational distributions to bound the log-marginal likelihood, as described in [KW14]. We aim to maximize the following expression [3]:

$$\text{ELBO}(\nu) = \mathbb{E}_q \left[\log \Pr(\mathbf{w} \mid \pi, \xi)\right] - \text{KLD}\left(q(\pi \mid \mathbf{w}, \nu) \parallel \Pr(\pi)\right) \tag{3.13}$$

where $q(\cdot)$ denotes the family of variational distributions, $\nu$ denotes the neural network parameters, and $a$ and $b$ are the Beta parameters learned by a MultiLayer Perceptron (MLP). The KLD is between two Beta distributions. We use *amortized variational inference* and Adam to fit the model concerning all the parameters.

**Algorithm 7** Inference process for the EDP

Initialize the model and its variational parameters
**for** $i \leftarrow 1$ to maximum number of iterations **do**
  Compute $\beta_k = \text{softmax}(\rho^\top \phi_k)$ for each topic $k$
  Choose a minibatch $\mathcal{B}$ of documents
  **for** each document $d$ in $\mathcal{B}$ **do**
    Get a bag of words representation $\mathbf{x}_d$
    Compute $a = \text{MLP}(\mathbf{x}; \nu_a)$
    Compute $b = \text{MLP}(\mathbf{x}; \nu_b)$
    Sample $v \sim \text{Beta}(a, b)$
    Compute $\pi = \begin{cases} v_1 \text{ if } k = 1 \\ v_k \Pi_{j<k}(1 - v_j) \text{ for } k > 1 \end{cases}$
    **for all** word $w_d$ in the document **do**
      Compute $\Pr(w_{dn} \mid \pi) = \text{softmax}(\pi \xi_{\cdot, w_{dn}})$
    **end for**
  **end for**
**end for**
Estimate the ELBO and its gradient through backprogation
Update the parameters

3. We disclose the formula details in Appendix A.

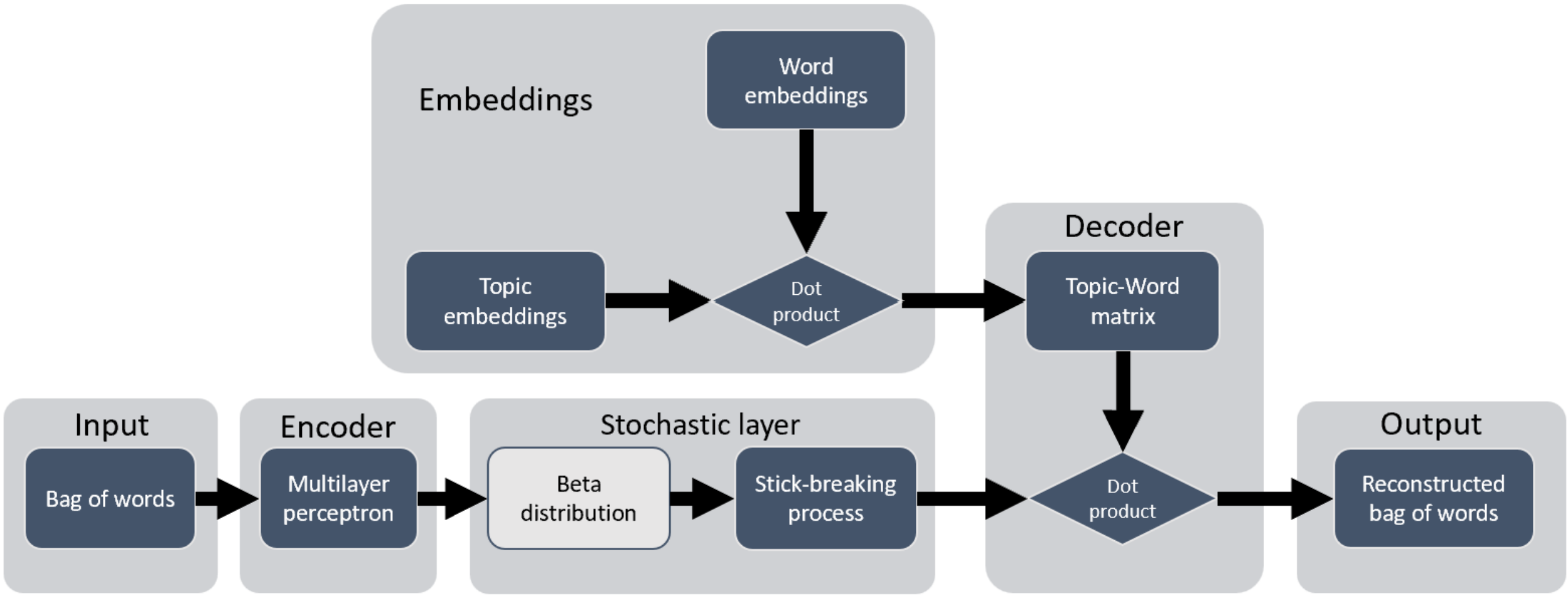

Figure 3.7 – Simplified representation of the EDP

### 3.2.2 The Embedded Hierarchical Dirichlet Process

The EDP's GEM prior parameter $\beta$ is equivalent to a Dirichlet's concentration parameter; the greater the parameter, the more stick breaks and, by extension, the more topics we get (Fig. 3.8). Learning this parameter from data enables controlling topic number growth. It is possible to do so using a Gamma hyperprior [EW95; MBJ06; Nin+20] as it is conjugate to the GEM distribution.

**Algorithm 8** Generative process for the EHDP

1: Choose $N_d \sim \text{Poisson}(\lambda)$
2: Draw $\beta \sim \text{Gamma}(\gamma_1, \gamma_2)$
3: Get a document-specific $G^{(d)}\left(\theta; \pi^{(d)}, \Theta\right) = \sum_{k=1}^{\infty} \pi_k^{(d)} \delta_{\theta_k}(\boldsymbol{\theta})$, with $\pi^{(d)} \sim \text{GEM}(\beta)$
4: **for all** word $w_{dn}$ in document $d$ **do**
5:     Draw a topic $\hat{\theta}_d \sim G^{(d)}\left(\theta; \pi^{(d)}, \Theta\right)$
6:     Draw a word $w_{dn} \sim \text{softmax}\left(\rho^T \phi\right)$
7: **end for**

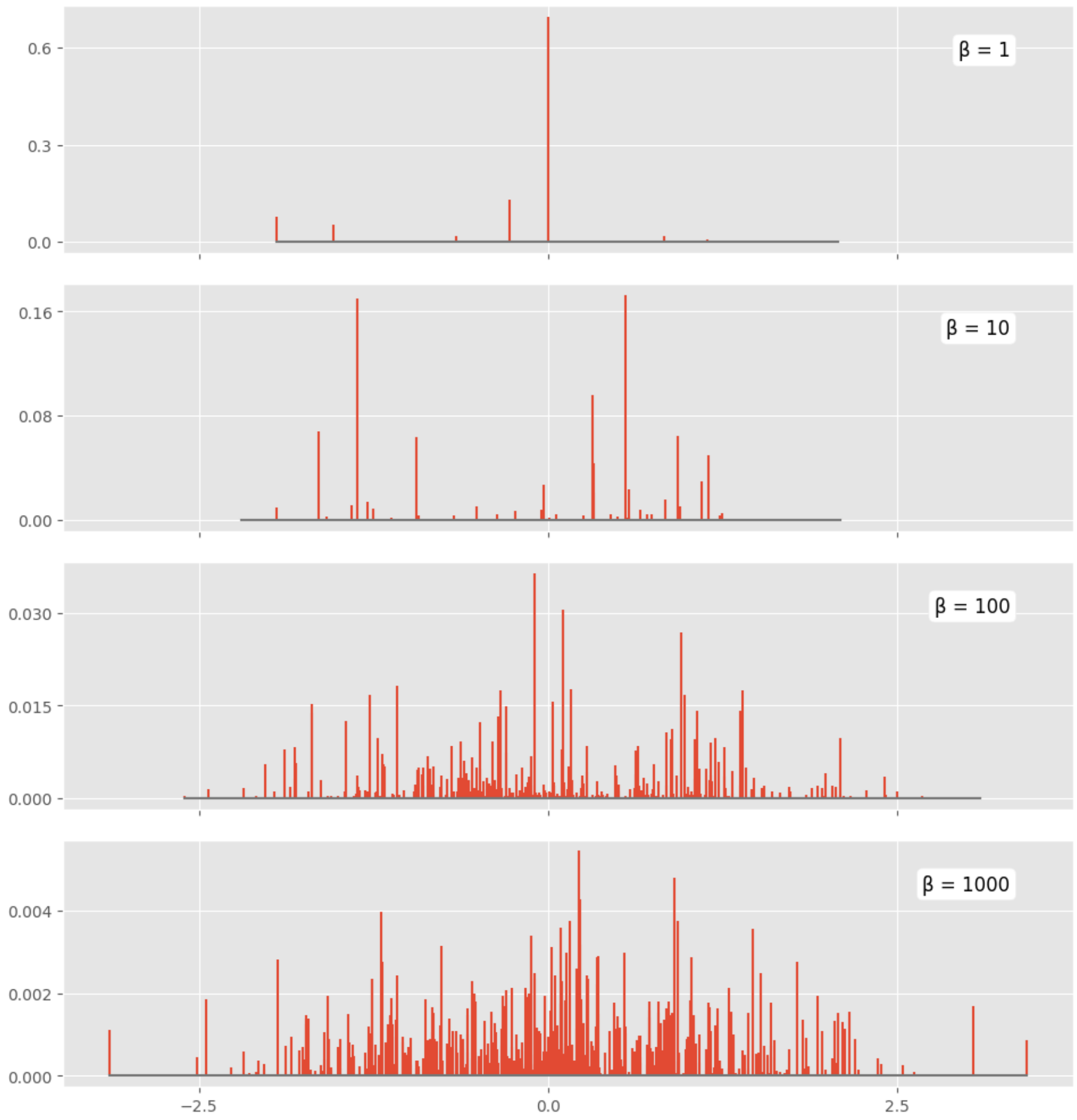

Figure 3.8 – Draws from a Dirichlet process with a standard Gaussian base function

The optimization objective is the following[4]:

$$\begin{aligned}\mathrm{ELBO}(\nu) = {} & \mathbb{E}_q\left[\log \Pr\left(\mathbf{w} \mid \pi, \xi\right)\right] \\ & + \mathbb{E}_q\left[\log \Pr\left(\nu \mid \beta\right)\right] \\ & - \mathbb{E}_q\left[\log q\left(\nu \mid \mathbf{w}\right)\right] \\ & - \mathrm{KLD}\left(q\left(\beta \mid g_1, g_2, \mathbf{w}\right) \parallel \Pr\left(\beta \mid \gamma_1, \gamma_2\right)\right)\end{aligned} \tag{3.14}$$

where $g_1$ and $g_2$ are the variational's distribution parameters, and $\gamma_1$ and $\gamma_2$ are the hyper-

4. We disclose the formula details in Appendix A.

prior's parameters. The inference process (Alg. 9) differs slightly from the EDP's (see line 12).

**Algorithm 9** Inference process for the EHDP

1: Initialize the model and its variational parameters
2: **for** $i \leftarrow 1$ to maximum number of iterations **do**
3: Compute $\beta_k = \mathrm{softmax}(\rho^\top \phi_k)$ for each topic $k$
4: Choose a minibatch $\mathcal{B}$ of documents
5: **for** each document $d$ in $\mathcal{B}$ **do**
6: Get a bag of words representation $\mathbf{x}_d$
7: Compute $a = \mathrm{MLP}\left(\mathbf{x}_d; \nu_a\right)$
8: Compute $b = \mathrm{MLP}\left(\mathbf{x}_d; \nu_b\right)$
9: Sample $v \sim \mathrm{Beta}\left(a, b\right)$
10: Compute $\pi = \begin{cases} v_k \text{ if } k = 1 \\ v_k \Pi_{j<k}\left(1 - v_j\right) \text{ for } k > 1 \end{cases}$
11: **for all** word $w_d$ in the document **do**
12: Compute $\Pr\left(w_{dn} \mid \pi, g_1, g_2\right) = \mathrm{softmax}\left(\pi \xi_{\cdot, w_{dn}}\right)$
13: **end for**
14: **end for**
15: **end for**
16: Estimate the ELBO and its gradient through backprogation
17: Update the parameters

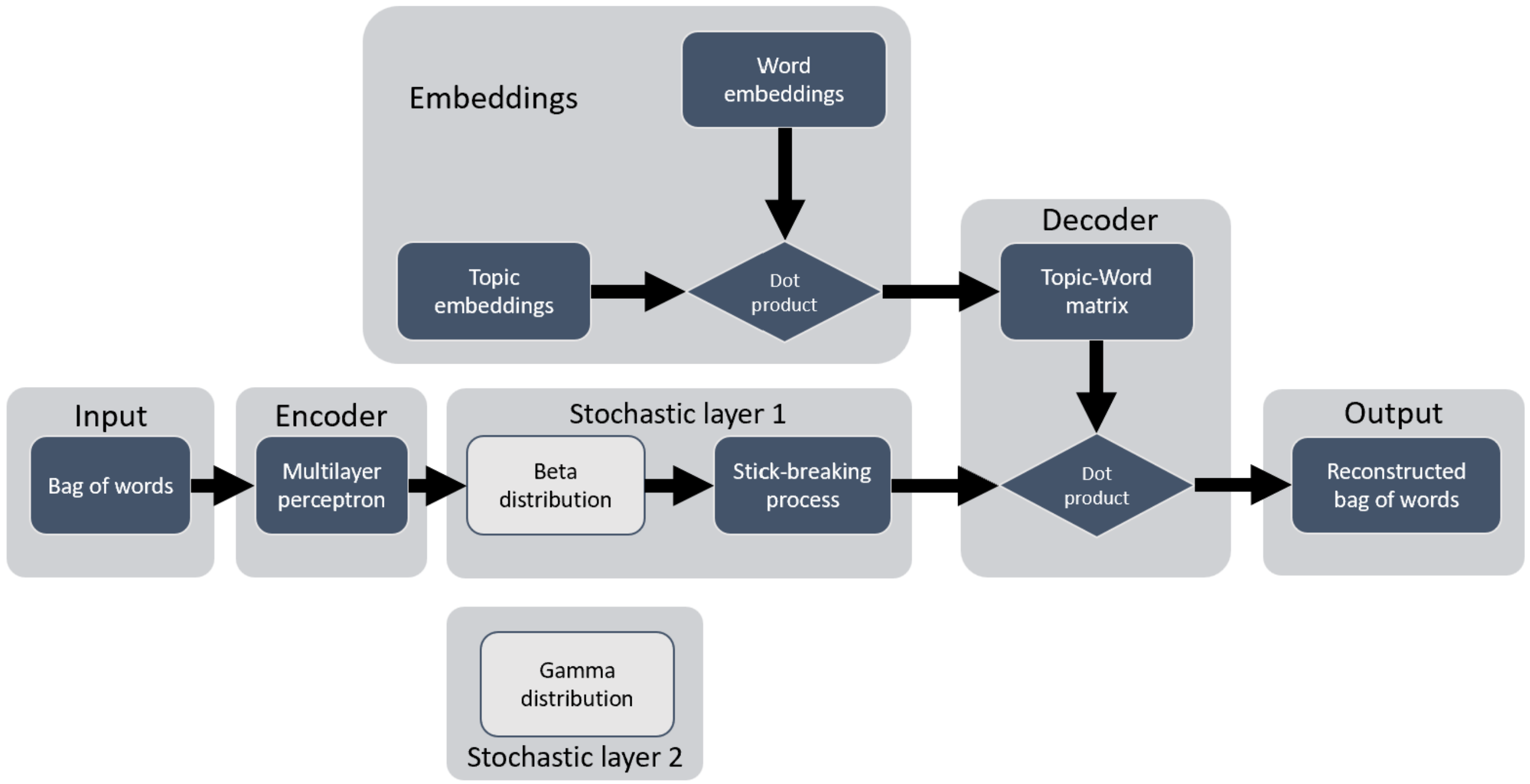

Figure 3.9 – Simplified representation of the EHDP

## 3.3 Empirical studies

This section tests and compares these models on two benchmark datasets and one industrial dataset. We start by giving precisions about our test metrics before turning to the actual experimentations.

### 3.3.1 Metrics

Our experimentations include a statistical assessment and a quality assessment. We use perplexity for the statistical goodness-of-fit. For quality (TQ), we use the product of diversity (TD) and mean NPMI (TC). We include an additional human evaluation by a domain expert for the industrial dataset and text annotation for the benchmark datasets to evaluate coherence. Please refer to Section 2.1.1.3 for more details.

### 3.3.2 Benchmark datasets

#### 3.3.2.1 Description and data preparation

The experiments feature the 20 Newsgroups (20NG) [5] and the Humanitarian Assistance and Disaster Relief articles (HADR) [Hor17] annotated datasets. HADR comes with a lexicon we will use for the qualitative estimation of results. Both datasets consist of collections of articles about several topics: 20 in the case of 20 Newsgroups and 25 for HADR. The 20 Newsgroups contain 18846 articles, while HADR contains approx. 504000 ones in different languages.

Due to technical limitations for this work, we retained a random subset of 20000 HADR articles for our experimentations. In each case, we used 85% of the entire dataset for the training sets, 10% for the validation sets, and 5% for the test sets. We filtered out words that do not appear in at least four documents and removed stopwords to accommodate our computational capabilities, thus yielding $V$-vocabularies of 28307 words from 20 Newsgroups and 32794 words from HADR.

5. `http://qwone.com/~jason/20Newsgroups/`

#### 3.3.2.2 Training settings

We compare our results with Ning and al.'s iTM-VAE-Prod, iTM-VAE-G [Nin+20], and the ETM [DRB20]. The iTM-VAE-Prod is a nonparametric topic model that places a GEM prior to a Kumaraswamy distribution, but the model does not include any word similarity mechanism. For further study, we also adapted iTM-VAE-Prod to include the implicit reparameterization trick. To avoid posterior collapse and stabilize VAE training, we used batch normalization with a batch size of 1000 documents and chose Adam with a learning rate of 0.002. We optimized the ELBOs for both the model and the variational parameters simultaneously for each model. We performed exponential decay on both first (0.95) and second moment (0.99) estimates. We also used weight decay ($1.2 \times 10^{-6}$). Last but not least, and following [DRB20], we normalized bag-of-words representations of documents by dividing them by the number of words for document length accommodation. We chose all the parameters and hyperparameters with cross-validation, including distributions and encoder sizes. The cross-validation, however, included both quantitative and qualitative metrics. For each model, we used multilayer perceptrons with two hidden layers of 100 neurons. We set the prior parameters to $\alpha = 1$ and $\beta = 5$ for both iTM-VAE-HP and EDP, $\delta_1 = 1$ and $\delta_2 = 20$ for both iTM-VAE-G and EHDP, and a standard Gaussian for ETM. We kept the same settings for both datasets. We give parametric model capacities for 50 and 200 topics and nonparametric models capacities for up to 200.
In practice, analysts require a topic model to provide both good insights about the topics and good predictability of unseen documents. Most topic models, however, are only trained and selected from a statistical point of view, with topic coherence computed periodically due to its expense. In this configuration, coherence is an additional indicator almost set apart from the training process. Our work focuses on maintaining a fair trade-off between goodness of fit and interpretability. During the validation step, we select our models based on a topic quality - perplexity ratio.

#### 3.3.2.3 Results and discussion

Nonparametric topic models that use the explicit reparameterization trick all suffered posterior collapse during the experiments. They started producing $NaN$s as soon as the second epoch; consequently, we excluded them from our analysis. The phenomenon, how-ever, confirms the value of using the original distribution instead of a surrogate as it enables building more robust probabilistic settings. As for the other algorithms, found that they all have similar predictive power 3.4. Thus, according to our selection criteria, topic quality is predominant in determining the best models. Tab. 3.5 displays topic quality for every model that did not experience posterior collapse. Our Embedded Hierarchical Dirichlet Process significantly outperforms the other techniques in terms of topic quality, even ETM and EDP with implicit reparameterization, despite these algorithms sharing the same decoder as EHDP. In addition, note that the NPMI only considers the co-occurrence of words in a single document when word embeddings are cross-corpora. Imagine two documents (1 and 2) and three words A, B, and C. Let A and B appear together in the first document, and B and C appear together in the second document.
Similarly to a transitive relation, word embeddings will find that if A and B are close and B and C are close, then A and C also exhibit some similarity. The NPMI will not take this aspect into account. Consequently, and for these reasons, the NPMI underestimates coherence in document models with word embeddings. Besides, statistical goodness-of-fit is the sole driver for model optimization, excluding semantical coherence. It is an additional indicator for model selection, i.e., we use coherence to choose amongst models with similar goodness of fit. Involving model quality in optimization may improve the results. This inclusion could be some form of regularization. As for topic coverage (Tab. 3.1), EHDP falls second to iTM-VAE-G but does not collapse to a single topic as its nonparametric pairs. These results clearly show that EHDP is a robust technique with a solid ability to adapt to datasets with large vocabularies, even when augmenting the number of words by nearly 16% when switching from 20 Newsgroups to HADR.

| Datasets | 20NG (%) | HADR (%) |
|---|---|---|
| *iTM-VAE-G* | **10 (50%)** | 1 (4%) |
| *EHDP* | 9 (45%) | **8 (32%)** |
| *iTM-VAE-HP* | 1 (5%) | 1 (4%) |
| *EDP* | 1 (5%) | 1 (4%) |

Table 3.1 – Topic coverage with respect to human judgement

The last topic, i.e., the French stopwords, is pure serendipity. In [DRB20], the authors show that ETM can handle stopwords and separate them in a specific topic, but the authors only tested the ETM in monolingual settings. HADR, however, is a multilingual dataset. Some documents in French remained after we filtered the dataset. EDHP still distinguished stopwords among the vocabulary while classifying them by language. We explain this result because our embeddings work with contexts and French words are much more likely to appear within French documents. Despite its accidental origin, the result is interesting, as multilingual topic modeling and text mining is still open issue [Vul+15; Yan+19]. To confirm our intuition about multilingual topic modeling, we benefit from our model's ability to generate embeddings. In particular, we extracted the 50 nearest neighbors of our French stopwords. Tab. 3.3 shows an example. As expected, most neighbors are also French words. However, as we get lower in the ranking, non-French words appear (in bold). We hypothesize that some words appear in several languages, especially event-related nouns, in a multilingual corpus. We think these words can act as pivots to link words from other languages, thus potentially enabling both supervised and unsupervised cross-lingual topic modeling with no additional adaptations.

| Topic | Word list |
|---|---|
| *India & Bangladesh* | tongi, manu, gorai, rly, storey, serjganj, kanaighat |
| *United Nations* | gva, dhagva, metzner, masayo, pbp, spaak, pos |
| *Weather* | tpc, nws, knhc, outward, forecaster, accumulations, ast |
| *Floods & landslides* | floods, landslides, padang, flooding, rain, mudslides, sichuan |
| *Africa* | drc, lusaka, monuc, burundi, darfur, congolese, amis |
| *Economic development* | development, financing, macroeconomic, management, reduction, usaid, sustainable |
| *Politics & diplomacy* | paragraph, decides, resolution, pursuant, vii, welcomes, stresses |
| *French stopwords* | les, qui, de, que, à, une, des |

Table 3.2 – Complete list of topics extracted from HADR by the EHDP

| |
|---|
| les, par, à, que, », rapport, pas, «, lui, ses, nt, entre, fin, qui, cet, aux, ou, gouvernement, fui, bien, han, ces, haïtien, deux, manger, **gomez**, unies, rdc, ya, sont, ne, une, notre, ont, locales, avril, sur, première, dans, cours, santé, **una**, croix, **zona**, **santander**, vie, **sailing**, mais, milliers, **casos** |

Table 3.3 – Nearest neighbors of the "les" french stopword in decreasing order

#### 3.3.2.4 Conclusion

We found that EHDP outperforms other state-of-the-art algorithms in most configurations and shows increased robustness to adapt to the dataset. Besides, we found that the EHDP can handle stopwords and make regroupings in a multilingual environment. As for its summarization capabilities, we noticed that the algorithm tends to combine several human-annotated topics into a single one.

| | **Perplexity** | | | | **TQ** | | | |
|---|---|---|---|---|---|---|---|---|
| | **20NG** | | **HADR** | | **20NG** | | **HADR** | |
| | *50* | *200* | *50* | *200* | *50* | *200* | *50* | *200* |
| *ETM* | 686.94 | 695.85 | 689.34 | 688.65 | 0.07 | -0.02 | 0.07 | 0.02 |
| *iTM-VAE-Prod* | 680.16 | | | | 0.09 | | 0.16 | |
| *EHDP* | **653.58** | | | | **0.20** | | **0.32** | |
| *iTM-VAE-HP implicit* | 662.66 | | | | 0.04 | | 0.12 | |
| *EDP implicit* | 664.54 | | | | 0.04 | | 0.08 | |

Table 3.4 – Dataset-wise and topic number-wise perplexity and topic quality

| Models \ No. topics | **TD** | | | | **TC** | | | | **TQ** | | | |
|---|---|---|---|---|---|---|---|---|---|---|---|---|
| | **20NG** | | **HADR** | | **20NG** | | **HADR** | | **20NG** | | **HADR** | |
| | *50* | *200* | *50* | *200* | *50* | *200* | *50* | *200* | *50* | *200* | *50* | *200* |
| *ETM* | 0.47 | 0.32 | 0.45 | 0.28 | 0.15 | -0.06 | 0.15 | 0.07 | 0.07 | -0.02 | 0.07 | 0.02 |
| *iTM-VAE-Prod* | 0.91 | | **1.0** | | 0.10 | | 0.16 | | 0.09 | | 0.16 | |
| *EHDP* | 0.52 | | **1.0** | | **0.38** | | **0.32** | | **0.20** | | **0.32** | |
| *iTM-VAE-HP implicit* | **1.0** | | **1.0** | | 0.04 | | 0.12 | | 0.04 | | 0.12 | |
| *EDP implicit* | **1.0** | | **1.0** | | 0.04 | | 0.08 | | 0.04 | | 0.08 | |

Table 3.5 – Dataset-wise and topic number-wise topic quality

| |
|---|
| [QUOTE="DBMandrake, post: 2885630, member: 15953"] Wish I'd known that before fitting them... [/QUOTE] Well, you knew GY Efficient Grip were the quietest you could buy because I told you. I have also advised that they are not too bad on the white slippy stuff either |

Table 3.6 – A document sample from the industrial database

### 3.3.3 Industrial dataset

#### 3.3.3.1 Description and data preparation

As we evolve in an industrial context related to the automotive industry with a particular emphasis on tires, our dataset is a subset of a database that contains documents scraped from 1073 websites, 442 of which are in English. As we do not take multilingual nor cross-lingual settings into account in our models, we discarded non-English speaking websites. We filter our data by source and employ language detection thanks to an off-the-shelf solution. Consequently, there is no guarantee that non-English documents are among these, as generic tools best work on standardized (i.e., very clean and homogeneous) text corpora such as journal articles. Our current setting considers non-English elements as noise that our models should isolate. Our data contains partial annotations according to *tire experts' insights* with specific tokens that replace original tokens or sets of tokens with in-house codification, thus forming a custom dictionary and a custom ontology. However, these annotations are not the focus of our interest for the in-house codification and ontology related to product characteristics, thus inducing biases. Experts tend to *focus on products*, while customers tend to *focus on their experience of the product.* We recall that we focus on customer insight, not expert insight.

Consequently, we rely on *totally unsupervised insight extraction on customers' reviews.* The experts' token replacement is hard-coded in our *ad-hoc* industrial process in charge of data cleaning. The pipeline performs DOM parsing, removes as many markup languages as possible, detects languages, then replaces tokens accordingly with in-house conventions, but does not fix typos. Despite these cleaning steps, there is no warranty whatsoever that the noise will not remain. To illustrate our point, we show a document from the database in Tab. 3.6. Noise appears in red, while interesting information appears in green. We could not retrieve the annotation-free text without a full database cleaning replay. We treated the in-house tokens on the same level as other tokens. We did not remove classical stopwords as they act as structuring elements for word embeddings. *We could not remove the punctuation as tire references (e.g., 205/50ZR15) and dimensions (e.g., 7.2/32") include some, and as their variability of denomination makes regexes almost useless for detecting them.* Last but not least, we did not set tokens to lowercase for the same reason as for punctuation. In the same way, as latex (the material) and $\LaTeX$ are not the same, continental (the adjective or the climate) is not the same as Continental (the tire brand).

The data sources are diverse, and so are the document lengths. Our documents include customer opinions from e-commerce websites, review articles, blog posts, forum threads, and even tweets. A document's length is closely related to the diversity of written expression. The issue is essential as topic models generally are bag-of-words-based. We posit three reasons:

1. there is a direct link between document length and document sparsity
2. a long document should have lesser sparsity than a short one yet should contain a high recurrence of a small vocabulary subset, as shown by Zipf's law
3. we want the topics to represent all the documents; said otherwise, we want to avoid topics that represent short or long documents only

We studied a database sample of 96910 documents to hint at how long a "standard" document is. Our statistical unit for this segmentation is the token, i.e., a string separated from others with blank space, even for expressions that could qualify as collocations (e.g., "state of the art" yields four tokens: "state," "of," "the," "art"). Our first task was to test if document length relates to linguistic diversity, i.e., the ratio between the number of distinct tokens and total tokens. The Spearman rank correlation test shows such correlation ($-0.86$ with $p < .05$), thus confirming that Zipf's law applies to our case. As for stopwords, their weight importance should remain the same in the results. As they are everywhere, they are more likely to act as noisy factors that the models should cope with than as discrimination factors.
We chose an adaptation of the Kolmogorov-Smirnov test to discrete distributions [Con72] to assess a fit for the document-wise number of tokens. We can model document length thanks to a Poisson distribution of parameter $\lambda = 107.9$ ($p < .05$). The distribution is a relatively good fit for 98.75% of our global sample (approx. 95699 documents). Consequently, we removed all documents whose size exceeds 450 tokens as the length distribution is different starting from this point. Due to limited computational capabilities, we needed to form a subsample. To preserve representativeness, we decided to stratify our sample using document length as classes. In particular, we intended to maintain a trade-off between having a good bin width and having a reasonable number of classes. The Poisson distribution hypothesis and Doane's formula allow us to segment the data with 22 bins histogram. We then determined the proportions for each class in the dataset. We achieved a stratified subsampling of 40000 documents according to these proportions. We dedicated approximately 80% of the subsample for training, 10% for validating, and 10% for testing while always respecting our segmentation. We retained a total $V$-vocabulary of 26731 words for our experimentations.

| Min. | Q1 | Median | Mean | Q3 | Max. |
|---|---|---|---|---|---|
| 2 | 45 | 80 | 107.9 | 135 | 4231 |

Table 3.7 – Document length description

#### 3.3.3.2 Training settings and evaluation

We test our models using two comparison grids: neural versus non-neural topic models and parametric versus non-parametric. Our work, first and foremost, attempts to solve large-scale topic modeling. We use online stochastic variational inference for all models with a batch size of 1000 documents. On the nonparametric side, we compare our models to the neural iTM-VAE-Prod and iTM-VAE-G models [Nin+20] and non-neural HDP[WPB11b]. We also devised a variant of the SB-VAE [NS17] that uses the IRT [FMM18]; we denote this variant *SB-VAE implicit*. Note that the *SB-VAE implicit* is the same as the iTM-VAE-Prod, except for the reparameterization. On the parametric side, we compare with the neural ProdLDA [SS17], ETM [DRB20], and the non-neural LDA [Hof+13]. We selected all models' parameters either by learning them from data whenever possible [6], or with cross-validation. For parametric topic models, we report the number of topics that yielded the

6. We did it for the EHDP, iTM-VAE-G, the LDA, and the HDP. We used the implementations provided by Gensim for the LDA and the HDP.

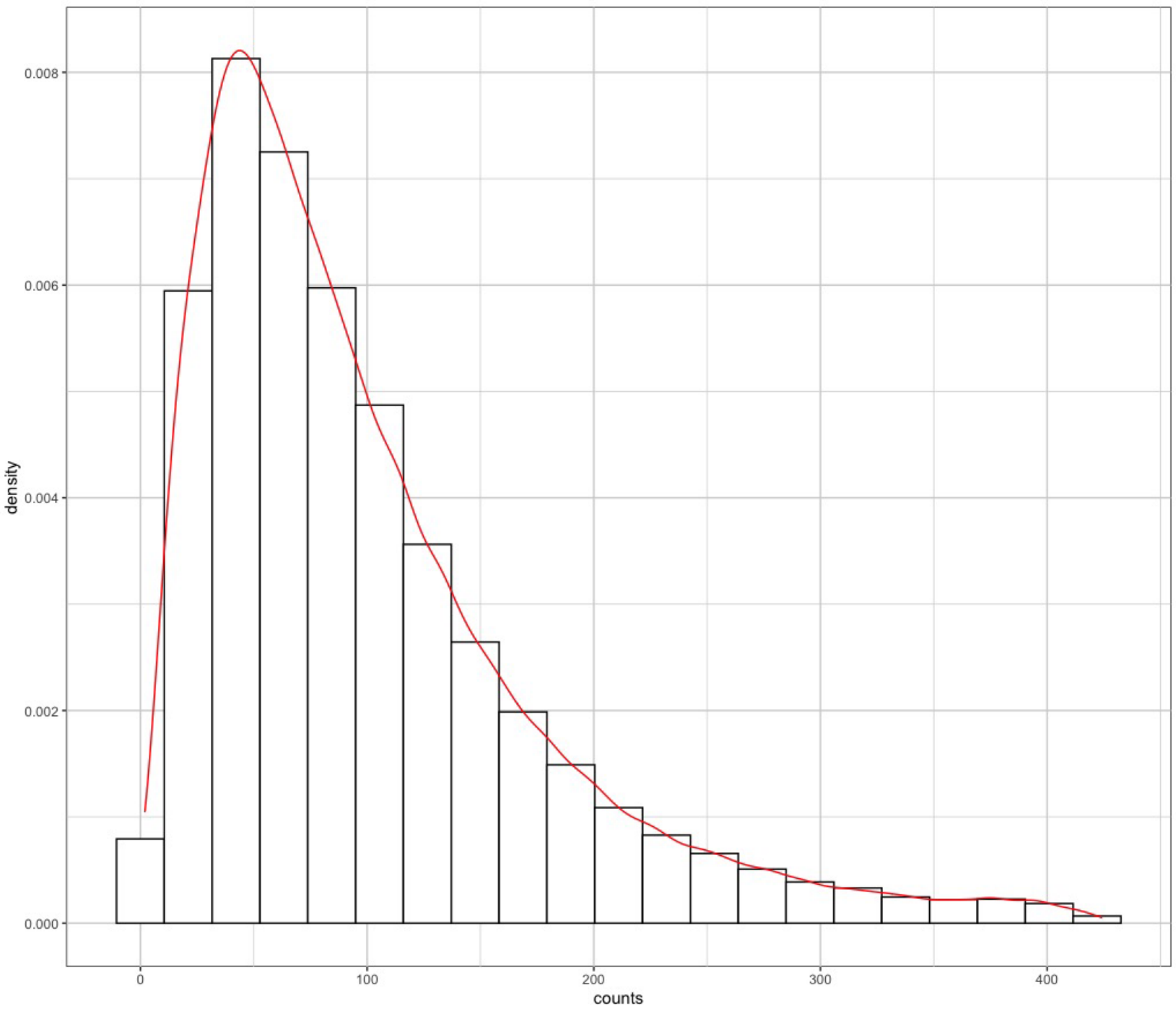

Figure 3.10 – Document length distribution segmentation and actual density

best log-likelihood. For VAE-topic models, we set the encoders to be multilayer perceptrons with two layers of 100 neurons each and used Adam with a learning rate of 0.002 and L2-regularization of $1.2e-6$. For the ETM and our models, we set the embedding sizes to 300. Thanks to transfer learning, we added two trials respecting the word embedding capabilities of these models. For the first one, we used the Glorot (also known as Xavier) normal initialization, and for the second one, we initialized the word embedding component with Skip-grams [Mik+13]. We call the models trained with each of these modalities raw (-R suffix) and transfer (-T suffix), respectively. We trained the Skip-grams embeddings with a window size of 4, an amount of 10 negative samples, and a dimension of 300. Regardless of the experiments, we let every model run until convergence within a limit of 150 iterations.

#### 3.3.3.3 Results and discussion

Our results (Tab. 3.8) clearly show that the neural nonparametric topic models achieve better results than any other kind of topic model. They achieve the lowest perplexities and have the best overall quality. The iTM-VAE is the exception. The ProdLDA and the ETM did not suffer from the phenomenon, as they could describe the data in the training step but could not generalize to unseen datasets. As the SB-VAE implicit is virtually the same as the iTM-VAE-G, the implicit reparameterization trick's effects are neatly visible, thus confirming the importance of the prior to properly extracting latent variables in the VAE context. The iTM-VAE-G exhibits approximately the same perplexity level as its nonparametric neural counterparts but has poorer topic diversity and is non-interpretable. The LDA achieves a similar quality but has a much higher perplexity than the other models. The transfer learning variants yielded close results as their "raw" counterparts regarding the goodness-of-fit and topic quality. The models with word embeddings resemble each other, but the SB-VAE implicit outperforms them in terms of coherence. For all that, we recall that the mean NPMI is an indicator whose values oscillate between $-1$ and 1. The indicator value shows independence between words for the best models (SB-VAE implicit included) at most, but the word embeddings' similarities show otherwise (Fig. 3.11). As the SB-VAE does not include word embeddings, it is more uncertain whether the terms it reports are coherent regarding the dataset.

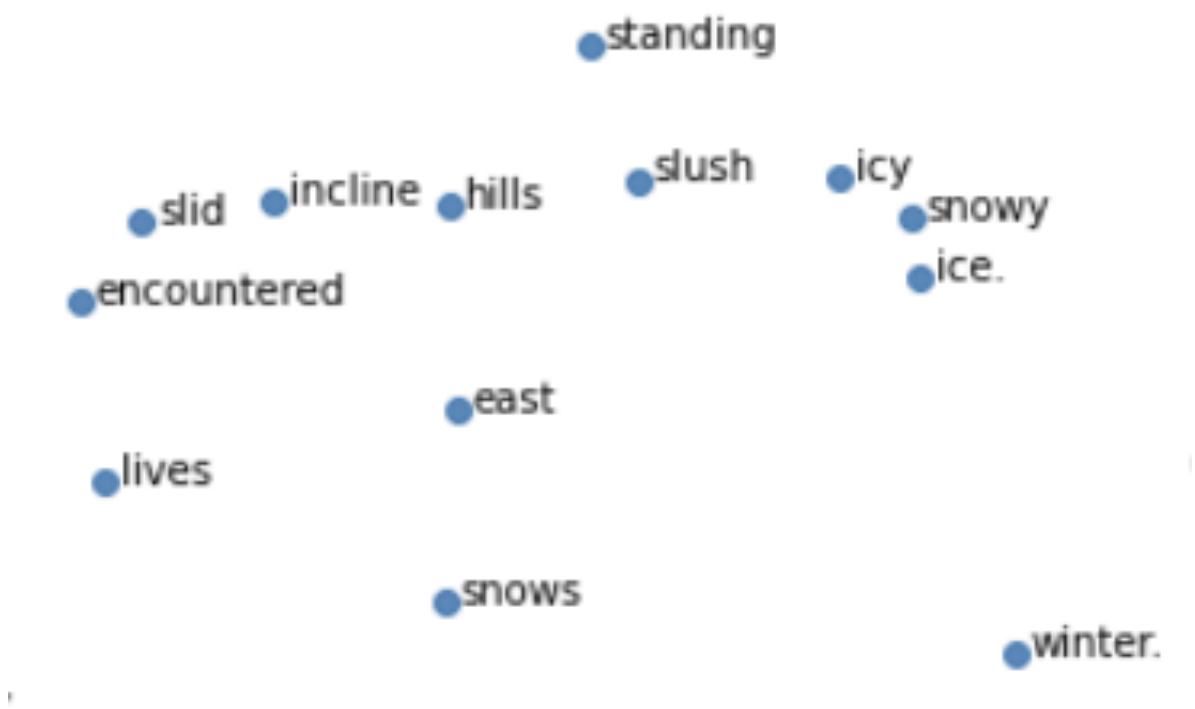

Figure 3.11 – Zoomed-in t-SNE representation of the tire adhesiveness topic of EHDP-T

| Model type | Model | No. topics | Perplexity | TC | TD | TQ |
|---|---|---|---|---|---|---|
| **Neural nonparametric** | *EHDP-R* | 9 | 664 | -0.05 | 0.97 | -0.05 |
| | *EHDP-T* | 10 | 655 | -0.05 | **1.0** | -0.05 |
| | *EDP-R* | 9 | 625 | 0.05 | **1.0** | 0.05 |
| | *EDP-T* | 9 | **622** | -0.02 | **1.0** | -0.02 |
| | *iTM-VAE* | NaN | NaN | NaN | NaN | NaN |
| | *SB-VAE implicit* | 12 | 631 | **0.23** | 0.73 | **0.17** |
| | *iTM-VAE-G* | 10 | 680 | 0.17 | 0.11 | 0.02 |
| **Nonneural nonparametric** | *HDP* | 50 | $2.7{\times}10^6$ | 0.21 | 0.02 | 0 |
| **Neural parametric** | *ProdLDA* | 10 | NaN | -0.65 | 0.8 | -0.52 |
| | *ETM-R* | 10 | NaN | -0.63 | **1.0** | -0.63 |
| | *ETM-T* | 10 | NaN | -0.65 | 0.93 | -0.60 |
| **Non-neural parametric** | *LDA* | 10 | $2.06 \times 10^5$ | 0.06 | 0.48 | 0.03 |

Table 3.8 – Results on the industrial dataset

Overall, we are much more mitigated about the qualitative results of the models, except for the EHDP-R and -T (Tables 3.9 and 3.10, respectively). The only models other than the EHDP that could extract interpretable topics were the EDP-R, the EDP-T, and the SB-VAE implicit. The SB-VAE implicit extracted 2 to 3 additional topics compared with the two versions of the EHDP and the EDP, all of which were poorly interpretable and with repeating words, hence the lower score on topic diversity (Table 3.8). The other topics it extracted were qualitatively similar to the EHDP-R's, but with different rankings for the words, more noise, and a blurrier separation among topics (within the same model's results), i.e., that some words appeared in several topics. For the shared topics, there is little difference between the EHDP-R and the EHDP-T; they took approximately the same amount of time to fit (around 55 epochs each), and the words that appear are mostly the same, although with different rankings. To verify this fact, we selected words from the results of each topic model and queried both versions of the EDP and the EHDP's word embeddings to see what their neighbor embeddings were under both configurations. As the results were similar and nearly equal, we conclude that the EDP and the EHDP models are capable of fitting word embeddings whose quality is at least equivalent to properly fitted Skip-grams. The EHDP-T, however, extracted one additional topic, looking more precise. The transfer learning step, it seems, has somewhat yet marginally helped topic extraction by providing pre-fitted word representations to rely on.

Considering the topics' labels, we annotated them according to their contents, i.e., we did not use any specific technique apart from domain knowledge. The models do not assume elements outside the dataset itself. We checked the word embeddings to deal with uncertainty about a word and queried the dataset thanks to the topics and the appearing word. For instance, let's consider the "tire adhesiveness" topic from Table 3.10, and the embeddings from Figure 3.11. The word "east" appeared in the topic and is close to the embedding corresponding to "hills." We deduced from our verifications from the dataset that it was about hills located in the East. The same goes for "Blackcircles" in the "Stopwords & online provider" topic. Blackcircles is a retailer, so it makes sense that price tags surround the word. The rest of the topic, however (e.g., "expand...", "[/QUOTE]"), is pure noise, so we will need to refine the ETL process around this data source, if not all retailer-related sources. In comparison with the EHDP-R's result (Table 3.9), we could only see that this topic was not only about stopwords, thanks to the benefits of the transfer learning step. Despite our fairly clear results, topic model evaluation is still an open issue under active development in the topic modeling field [Hoy+21].

#### 3.3.3.4 Conclusion

In this section, we successfully applied the EDP and the EHDP models to a customer insight extraction task from web-scraped data related to the tire industry. We found that according to a domain expert, the EDP and the EHDP outperform other state-of-the-art algorithms in this precise task. The fully-fledged Dirichlet Process priors enable better capture of corpora hidden properties. The models also allowed for word disambiguation and showed capabilities to work in a very noisy context. We showed that the EDP and the EHDP are useful for refining cleaning processes as they regrouped a retailer with its inner noise. Last but not least, we found discrepancies between the metrics and the intrinsic quality of the models, thus making developing efficient indices an open issue.

| | |
|---|---|
| **Performance & maneuverability** | "performance", "dry", "conditions.", "season", "conditions", "cornering", "handling", "confidence", "warm", "performance." |
| **Evaluation criteria** | "4", "5", "3", "1", "great", "off_road_note:", "ride_comfort_note:", "treadwear_note:", "recommended:", "durability_note:" |
| **Tire references** | "assy", "zz5's", "ZZ5's", "f1s", "f1", "rainsports", "Falkan", "FK45x", "eagle", "F1's" |
| **Handling in winter conditions** | "treads", "east", "handles", "stuck.", "rides", "ice", "snow.", "threw", "shipped", "son" |
| **Satisfaction about the durability** | "lasted", "far.", "had.", "pleased", "replaced.", "5k", "30k", "far,", "impressed", "definitely" |
| **Tire adaptatibility** | "fitment:", "road_types:", "road_conditions:", "Ta11", "Mixed", "4x2", "Suv", "ice_traction_note:", "N5000", "ort_note:" |
| **Confidence in the product** | "confidence", "daily", "driving.", "comfortable", "driven", "corners", "Continental", "rain", "winter.", "review" |
| **Climatic conditions** | "winter", "summer", "weather", "excellent", "wet", "winter.", "weather.", "most", "poor", "season" |
| **Stopwords** | "tyre", "{[}/QUOTE{]}", "fitted...", "tyres.", "member:", "expand...", "£110", "ago,", "post:", "tyres," |

Table 3.9 – Complete list of topics extracted by EHDP-R

| | |
|---|---|
| **Performance & maneuverability** | "performance", "conditions,", "all-season", "cornering", "dedicated", "performance.", "braking", "superior", "standing", "dry" |
| **Evaluation criteria** | "5", "4", "Rain", "winter_traction_note:", "3", "G2", "great", "1", "handling_note:", "off_road_note:" |
| **Tire references** | "zz5's", "FK45x", "assy", "ZZ5's", "Falkan", "f1s", "5:08", "influence.", "f1", "Vorti." |
| **Tire adhesiveness** | "incline", "encountered", "icy", "snowy", "hills", "lives", "slid", "slush", "ice.", "east" |
| **Durability and satisfaction** | "lasted", "far.", "had.", "Haven't", "replaced.", "far,", "25k", "complaints", "20k", "5k" |
| **Tire adaptability** | "fitment:", "road_types:", "road_conditions:", "Mixed", "4x2", "ice_traction_note:", "N5000", "ort_note:", "Suv", "Nt850" |
| **Confidence in the brand** | "Continental", "spirited", "review", "Mazda", "snows", "average", "daily", "driven", "corners", "driving." |
| **Climatic conditions** | "winter", "summer", "winter.", "season", "weather", "weather.", "seasons", "wet", "excellent", "conditions" |
| **Stopwords & online provider** | "expand...", "member:", "post:", "{[}/QUOTE{]}", "quid", "£110", "£50", "ZE914", "Blackcircles,", "£100" |
| **Driving experience** | "wore", "these.", "Civic", "Ecsta", "SL", "Definitely", "terrible", "wear,", "noisy.", "these," |

Table 3.10 – Complete list of topics extracted by EHDP-T

## 3.4 General conclusion

This section presents two novel models: the Embedded Dirichlet Process and the Embedded Hierarchical Dirichlet Process. These two nonparametric, VAE-based topic models can capture the number of topics and their contents, as well as topic embeddings and word embeddings that are viewable in the same space. We used them in benchmark and industrial settings and found their results better than their state-of-the-art counterparts. These approaches also prove helpful in distinguishing noise from data sources. Despite these promising results, we also found that current topic modeling metrics do not align with practitioners' expectations on topic models' evaluation.

# Chapter 4

# The Dynamic Embedded Dirichlet Process

In this Chapter, we extend our Embedded Dirichlet Process to capture time dynamics. We call this extension Dynamic Embedded Dirichlet Process. Similarly to its static counterparts, this extension can automatically detect the number of topics in a corpus and compute word embeddings and topic embeddings for better text exploration. The topic embeddings, however, take time into account. We start by laying the theoretical foundations for building our temporal extension. We then test our model on two standardized datasets before applying them to two industrial datasets from several social media sources. We show that it performs equally well to better than its counterparts in empirical, comparative studies.

## 4.1 Related work

The following sections present the theoretical foundations for our Dynamic Embedded Dirichlet Process. We first present Blei & al.'s dynamic topic modeling framework [BL06], then show how to adapt stick-breaking processes to capturing time dynamics.

### 4.1.1 Dynamic topic modeling

This section presents the Dynamic Topic Model (DTM) framework and an example with the corresponding inference algorithm.

#### 4.1.1.1 Theoretical framework

The DTM framework [BL06] extends the LDA-based topic models. To achieve this extension, it replaces the exchangeability assumption on a document collection by enforcing time dependencies. It is also possible to consider the DTM as a framework for time series that focuses on categorical data instead of continuous data. The authors introduce their framework in their article by showing how they extend the LDA to time-dependent settings. We call this extension Dynamic LDA (D-LDA).
Suppose that a data set is divisible in $T$ time slices and that each time slice $t \in \{1, \ldots, T\}$ evolves from time slice $t-1$, i.e., each time slice $t$ depends on time slice $t-1$. Let $\beta_{1:K}$ be a set of $K$ topics, each representing a distribution over a fixed $V$-vocabulary. Also, let $\beta_{t,k}$ denote the $V$-vector of natural parameters for topic $k \in \{1, \ldots, K\}$ in time slice $t$. The LDA uses a word-level multinomial distribution; so does the D-LDA. The usual representation for a multinomial is its mean reparameterization we denote with $\pi$. The $i^{\text{th}}$ component of the natural parameter for the multinomial is the following mapping:

$$\beta_i = \log\left(\pi_i / \pi_V\right) \tag{4.1}$$

As the Dirichlet distribution is not amenable to sequential modeling, the authors switch to a topic-wise chain of Gaussians to model uncertainty about the distributions over words:

$$\beta_{t,k} \mid \beta_{t-1,k} \sim \mathcal{N}\left(\beta_{t-1,k}, \sigma^2 I\right) \tag{4.2}$$

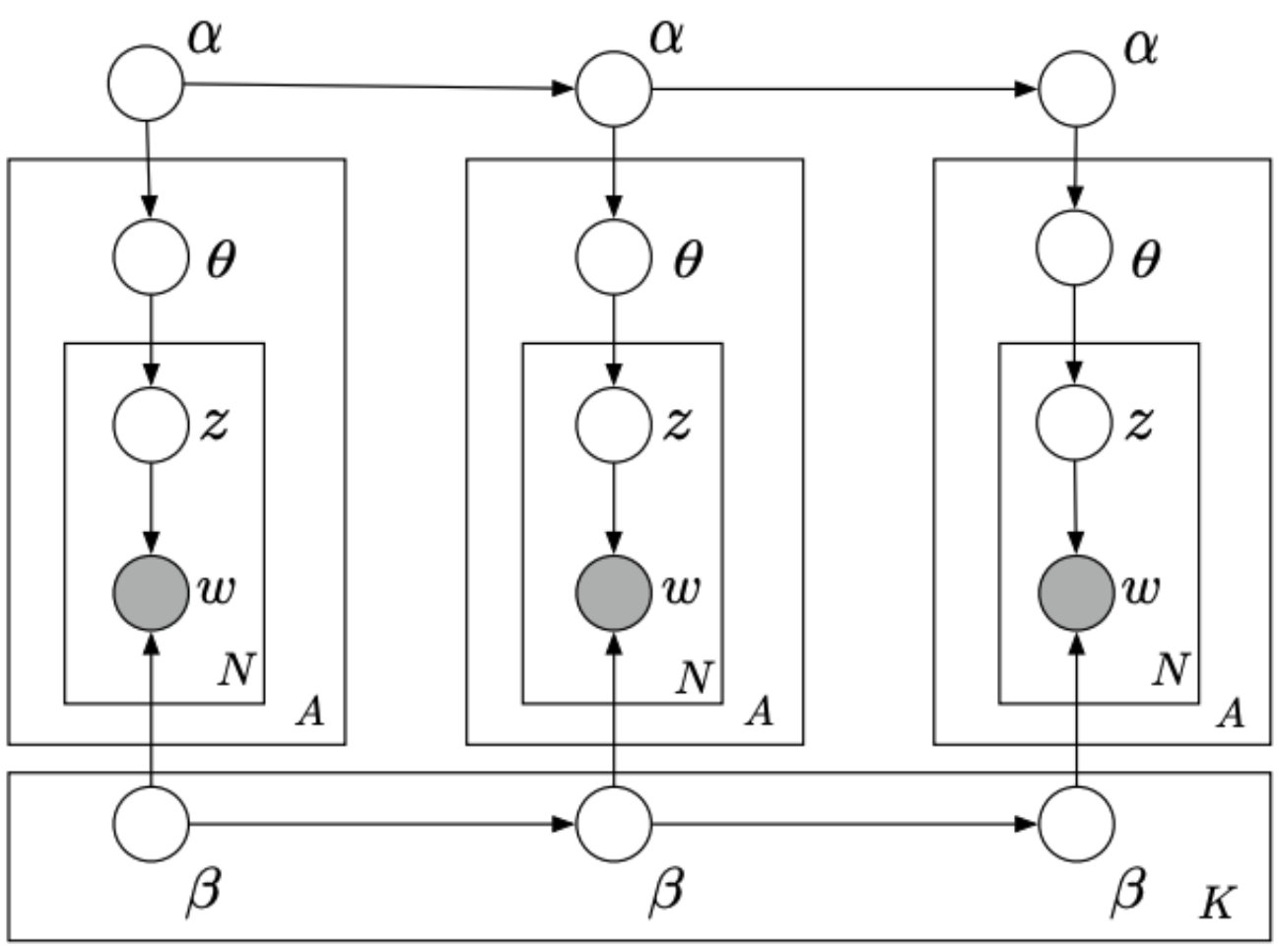

Figure 4.1 – D-LDA's graphical model

The logistic-normal distribution then maps the emitted values to the simplex, thus extending the logistic-normal distribution to time-series simplex data.

The D-LDA uses a topic-level logistic normal distribution with mean $\alpha$ for the same reasons. A dynamic model also captures the time dependencies:

$$\alpha_t \mid \alpha_{t-1} \sim \mathcal{N}\left(\alpha_{t-1}, \delta^2 I\right) \tag{4.3}$$

The idea is different from the CTM despite using a logistic-normal distribution. The D-LDA uses a diagonal covariance matrix, thus not modeling topic correlation dynamics.

**Algorithm 10** Generative process for the D-LDA

Draw topics $\beta_t \mid \beta_{t-1} \sim \mathcal{N}\left(\beta_{t-1}, \sigma^2 I\right)$
Draw $\alpha_t \mid \alpha_{t-1} \sim \mathcal{N}\left(\alpha_{t-1}, \delta^2 I\right)$
**for all** document $d$ **do**
    Draw $\eta \sim \mathcal{N}\left(\alpha_t, a^2 I\right)$
    **for all** word $w_n$ in the document **do**
        Draw $z \sim \text{Multinomial}(\pi(\eta))$
        Draw $W_{t,d,n} \sim \text{Multinomial}\left(\pi\left(\beta_{t,z}\right)\right)$
    **end for**
**end for**

The $\pi$ function maps the multinomial's natural parameters to its mean parameters:

$$\pi\left(\beta_{k,t}\right)_w = \frac{\exp\left(\beta_{k,t,w}\right)}{\sum_w \exp\left(\beta_{k,t,w}\right)} \tag{4.4}$$

For parameter inference, Blei & Laferty [BL06] have two variational methods for a mean-field approximation. We refer the reader to the original publication for more details about these methods. The non-conjugacy between the Gaussian and the multinomial complexifies the inference step. The optimization objective aims at fitting the following latent variables: the topic parameters $\beta_{t,k}$, the mixture proportions $\theta_{t,d}$, and the topic indicators $z_{t,d,n}$. The objective is the following:

$$\prod_{k=1}^{K} q\left(\beta_{k,1},\ldots,\beta_{k,T} \mid \hat{\beta}_{k,1},\ldots,\hat{\beta}_{k,T}\right) \times \prod_{t=1}^{T}\left(\prod_{d=1}^{D_t} q\left(\theta_{t,d} \mid \gamma_{t,d}\right) \prod_{n=1}^{N_{t,d}} q\left(z_{t,d,n} \mid \phi_{t,d,n}\right)\right) \tag{4.5}$$

The mean-field approximation considers latent variables independently of the others. On the left-hand side, the topic parameters are fit to minimize the KLD between a Gaussian resulting posterior and a non-Gaussian true posterior. The document-level latent variables on the right-hand side exhibit the same form as the LDA [BNJ03]. Each proportion parameter $\theta_{t,d}$ is endowed with a free Dirichlet parameter $\gamma_{t,d}$ and each topic indicator $z_{t,d,n}$ is endowed with a free multinomial parameter $\phi_{t,d,n}$.
Contrasting with the D-LDA, the AEVB framework performs amortized variational inference, thus considering parameters conjointly. The AEVB framework is also scalable. On the other hand, the coordinate ascent variational inference approach can fit this non-conjugate approach but is not scalable as it needs observation-wise re-evaluation. The following section presents a neural, AEVB-based application of the DTM framework.

#### 4.1.1.2 An implementation: the Dynamic Embedded Topic Model

The Dynamic Embedded Topic Model (D-ETM) [DRB19] is an extension to both the D-LDA and the ETM, thus exhibiting properties from both. It is also an implementation of the DTM framework. Like the D-LDA and the ETM, the algorithm learns word embeddings from data using entire document contexts (or topics) instead of surrounding words. The contexts themselves are topic embeddings that are visualizable in the same space as the word embeddings. Contrasting with the ETM, the D-ETM's topic embeddings include a time dimension to model the topics' evolution. This time dimension implies the inclusion of an additional variational objective to model the embedding's time dynamics. The word embeddings, however, do not include any time dimension: the model focuses on topic dynamics rather than on semantics dynamics.

Let $v$ be a term from a $V$-vocabulary. For each term $v$, we consider an $L$-dimensional embedding representation $\rho_v$. The D-ETM posits a topic embedding $\alpha_k^{(t)} \in \mathbb{R}^L$ for a given $K$-set of topics at a given timestamp $t = 1, \ldots, T$. Under the D-ETM, the probability of a word under a given topic is the following:

$$\Pr\left(w_{dn} = v \mid z_{dn} = k, \alpha_k^{(t_d)}\right) \propto \exp\left\{\rho_v^\top \alpha_k^{(t_d)}\right\} \tag{4.6}$$

where $\rho_v^\top \alpha_k^{(t_d)}$ is a normalized dot-product. According to the authors, the probability for a given term is higher when the word embedding for the term and the topic's embedding are in agreement. Consequently, semantically similar words are assigned to similar topics.

The D-ETM uses a Markov Chain over the topic embeddings $\alpha_k^{(t)}$ such that the topic representations evolve with Gaussian noise with variance $\gamma^2$:

$$\Pr\left(\alpha_k^{(t)} \mid \alpha_k^{(t-1)}\right) = \mathcal{N}\left(\alpha_k^{(t-1)}, \gamma^2 I\right) \tag{4.7}$$

Last but not least, and similarly to the D-LDA, the D-ETM captures how the general topic usage evolves over time. The prior over $\theta_d$ has dependencies on a latent variable $\eta_{td}$ where $t_d$ is the timestamp for document $d$.

$$\Pr\left(\theta_d \mid \eta_{t_d}\right) = \mathcal{LN}\left(\eta_{t_d}, a^2 I\right) \text{ where } \Pr\left(\eta_t \mid \eta_{t-1}\right) = \mathcal{N}\left(\eta_{t-1}, \delta^2 I\right) \tag{4.8}$$

Contrasting with the D-LDA, the D-ETM uses data subsampling [Hof+13] and amortization [GG14] to scale to large datasets and reduce the number of variational parameters. Likewise, the variational objective does not include non-conjugate distributions. The ELBO is the following [1]:

$$\text{ELBO}(\nu) = \mathbb{E}_q\left[\log \Pr(\mathbf{w}, \theta, \eta, \alpha) - \log q_\nu(\theta, \eta, \alpha)\right] \tag{4.9}$$

1. We disclose the formula details in Appendix A.

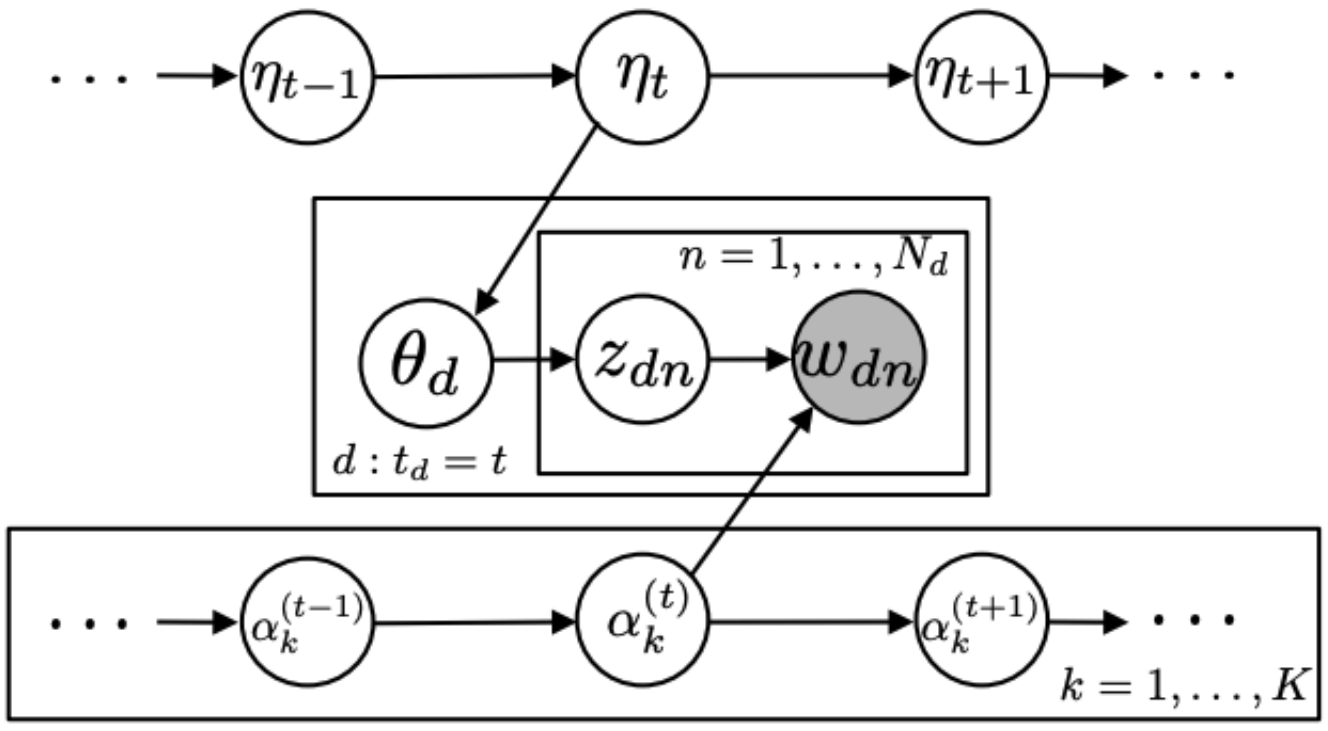

Figure 4.2 – D-ETM's graphical model

In the ELBO, $\Pr(\mathbf{w}, \theta, \eta, \alpha)$ is a function of the data, whereas $q_\nu(\theta, \eta, \alpha)$ is a structured variational family. The authors introduce the following approximation:

$$q(\theta, \eta, \alpha) = \prod_d q\left(\theta_d \mid \eta_{t_d}, \mathbf{w}_d\right) \times \prod_t q\left(\eta_t \mid \eta_{1:t-1}, \widetilde{\mathbf{w}}_t\right) \times \prod_k \prod_t q\left(\alpha_k^{(t)}\right) \tag{4.10}$$

$q\left(\theta_d \mid \eta_{t_d}, \mathbf{w}_d\right)$ represent the distribution over the topic proportions. These distributions are logistic-normal distributions whose mean, and covariance parameters are functions of the latent mean $\eta_{t_d}$ and the bag-of-words representation of the $d$-th document $\mathbf{w}_d$. Also, feed-forward neural networks whose inputs are both $\eta_{t_d}$ and $\mathbf{w}_d$ parameterize them. The distributions over the latent means $q\left(\eta_t \mid \eta_{1:t-1}, \widetilde{\mathbf{w}}_t\right)$ depend on all previous latent means $\eta_{1:t-1}$ and the normalized representation of documents with the $t^{\text{th}}$ timestamp, $\widetilde{\mathbf{w}}_t$. A generative LSTM parameterizes these Gaussian distributions. Last but not least, and contrasting with the terms before that use structured variational inference, the authors use the mean-field family for the topic embeddings $q\left(\alpha_k^{(t)}\right)$.

**Algorithm 11** Generative process for the D-ETM

Choose $N_d \sim \text{Poisson}(\lambda)$
Draw initial topic embedding $\alpha_k^{(0)} \sim \mathcal{N}(0, I)$
Draw initial topic proportion mean $\eta_0 \sim \mathcal{N}(0, I)$
**for all** For time step $t = 1, \dots, T$ **do**
  Draw topic embeddings $\alpha_k^{(t)} \sim \mathcal{N}\left(\alpha_k^{(t-1)}, \gamma^2 I\right)$ for $k = 1, \dots, K$
  Draw topic proportion means $\eta_t \sim \mathcal{N}\left(\eta_{t-1}, \delta^2 I\right)$
**end for**
**for all** document $d$ **do**
  Draw topic proportions $\theta_d \sim \mathcal{LN}\left(\eta_{t_d}, a^2 I\right)$
  **for all** word $w_n$ in the document **do**
    Draw topic assignment $z_{dn} \sim \text{Cat}(\theta_d)$
    Draw word $w_{dn} \sim \text{Categorical}\left(\text{softmax}\left(\rho^\top \alpha_{z_{dn}}^{(t_d)}\right)\right)$
  **end for**
**end for**

**Algorithm 12** Inference process for the D-ETM

Initialize the model and its variational parameters
**for** $i \leftarrow 1$ to maximum number of iterations **do**
  Sample the latent means and the topic embeddings, $\eta \sim q(\eta \mid \widetilde{\mathbf{w}})$ and $\alpha \sim q(\alpha)$
  Compute the topics $\beta_k^{(t)} = \text{softmax}(\rho^\top \alpha_k^{(t)})$ for $k = 1, \dots, K$ and $t = 1, \dots, T$
  Choose a minibatch $\mathcal{B}$ of documents
  **for** each document $d$ in $\mathcal{B}$ **do**
    Sample the topic proportions $\theta_d \sim q(\theta_d \mid \eta_{td}, \mathbf{w}_d)$
    **for all** word $w_{dn}$ in the document **do**
      Compute $\Pr(w_{dn} \mid \theta_d) = \sum_k \theta_{dk} \beta_{k, w_{dn}}^{(t_d)}$
    **end for**
  **end for**
**end for**
Estimate the ELBO and its gradient through backpropagation
Update the model and variational parameters using Adam

### 4.1.2 Dirichlet processes and time dynamics

Dirichlet Processes are reportedly not amenable to modeling spatial and temporal dependencies [RD11; Ren+11] in data. In their paper, Nalisinick & al. (2017), the Probit Stick-Breaking Process (PSBP) [RD11] is a designated alternative to weight sampling in the Dirichlet Processes. More precisely, it is possible to obtain Gaussian samples and then use a squashing function to map them on $(0, 1)$:

$$v_k = g\left(\mu_k + \sigma_k \otimes \epsilon\right) \tag{4.11}$$

In the above formula, $\mu$ and $\sigma$ are, respectively, the location and scale parameters for a Gaussian and $\epsilon \sim \mathrm{N}(0, 1)$ is an additional, element-wise Gaussian noise. Also, $g(\cdot)$ is the Gaussian cumulative distribution function (CDF).

In their paper, [RD11] propose the PSBP with spatial and temporal dependency modeling in mind. In the PSBP, the distributions share the same atoms from $G_0$, and the $\mu$ and $\sigma$ parameters control the variance of the sampled distributions around the mean $G_0$. Rodriguez & Dunson emphasize their approach as being close to the continuation ratio logit, and continuation ratio probit models used in discrete-time survival analysis [Agr19; AC01] and flexible enough to create various nonparametric models. To our knowledge, only a few works exist about the combination of topic modeling, and survival analysis [Li+20]. Additionally, the PSBP maintains the theoretical guarantee about the discreteness of a sample from a base distribution and that a truncated model is an excellent approximation to the infinite process, thus preserving computational simplicity. These properties make this reparameterization choice as enjoyable as the original Beta distribution used in the original Dirichlet Processes. The Gaussian CDF, however, does not have a closed form. Consequently, [NS17] replace the Gaussian CDF with a logistic function: $g(x) = 1/\left(1 + e^{-x}\right)$. We design this alternative as Logit Stick Breaking Process (LSBP) for simplicity. Despite the name, we emphasize that this is very close, yet not the same process as in [Ren+11]. We believe, nonetheless, that our model is extensible to capturing spatial configurations in data using an appropriate kernel distance in the stick-breaking process. However, this desirable feature is beyond our industrial project's scope and is left for future research.

## 4.2 The model

The Dynamic Embedded Dirichlet Process (D-EDP) extends the D-ETM to nonparametric settings. Consequently, it is also an application of the DTM framework. To automatically determine the number of topics, it uses the Gauss-Logit parameterization suggested in [NS17].

Let $v$ be a term from a $V$-vocabulary. For each term $v$, we consider an $L$-dimensional embedding representation $\rho_v$. The D-EHDP posits a topic embedding $\alpha_k^{(t)} \in \mathbb{R}^L$ for $k \in \{1, \ldots, \infty\}$ at a given timestamp $t = 1, \ldots, T$. Under the D-EHDP, the probability of a word under a given topic is the following:

$$\Pr\left(w_{dn} = v \mid z_{dn} = k, \alpha_k^{(t_d)}\right) \propto \exp\left\{\rho_v^\top \alpha_k^{(t_d)}\right\} \tag{4.12}$$

where $\rho_v^\top \alpha_k^{(t_d)}$ is a normalized dot-product. The D-EHDP uses a Markov Chain over the topic embeddings $\alpha_k^{(t)}$ such that the topic representations evolve with Gaussian noise with variance $\gamma^2$:

$$\Pr\left(\alpha_k^{(t)} \mid \alpha_k^{(t-1)}\right) = \mathcal{N}\left(\alpha_k^{(t-1)}, \gamma^2 I\right) \tag{4.13}$$

Last but not least, the prior over $\theta_d$ has dependencies on a latent variable $\eta_{td}$ where $t_d$ is the timestamp for document $d$.

$$\Pr\left(\theta_d \mid \eta_{t_d}\right) = \mathcal{N}\left(\eta_{t_d}, a^2 I\right) \text{ where } \Pr\left(\eta_t \mid \eta_{t-1}\right) = \mathcal{N}\left(\eta_{t-1}, \delta^2 I\right) \tag{4.14}$$

**Algorithm 13** Generative process for the D-EDP

Choose $N_d \sim \text{Poisson}\,(\lambda)$
Draw initial topic embedding $\alpha_k^{(0)} \sim \mathcal{N}(0, I)$
Draw initial topic proportion mean $\eta_0 \sim \mathcal{N}(0, I)$
**for all** For time step $t = 1, \ldots, T$ **do**
    Draw topic embeddings $\alpha_k^{(t)} \sim \mathcal{N}\left(\alpha_k^{(t-1)}, \gamma^2 I\right)$ for $k = 1, \ldots, K$
    Draw topic proportion means $\eta_t \sim \mathcal{N}\,(\eta_{t-1}, \delta^2 I)$
**end for**
**for all** document $d$ **do**
    Draw topic proportions $\theta_d \sim \mathcal{N}\,(\eta_{t_d}, a^2 I)$
    **for all** word $w_n$ in the document **do**
        Draw topic assignment $z_{dn} \sim \text{Cat}\,(\theta_d)$
        Draw word $w_{dn} \sim \text{Cat}\left(\text{softmax}\left(\rho^\top \alpha_{z_{dn}}^{(t_d)}\right)\right)$
    **end for**
**end for**

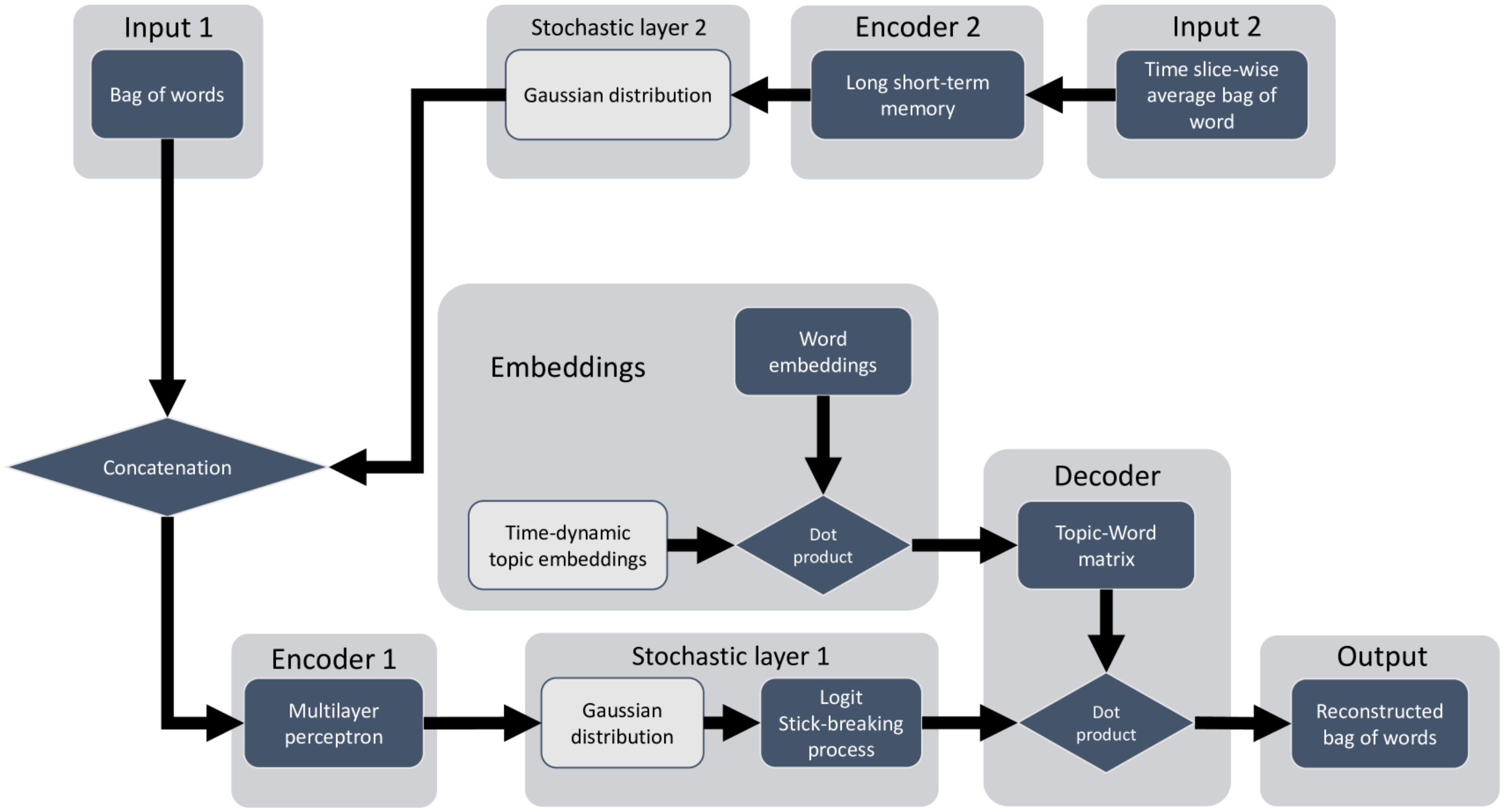

Figure 4.3 – Simplified representation of the D-EDP

**Algorithm 14** Inference process for the D-EDP

```
Initialize the model and its variational parameters
for i ← 1 to maximum number of iterations do
    Sample the latent means and the topic embeddings, η ∼ q(η | w̃) and α ∼ q(α)
    Compute the topics β_k^(t) = softmax(ρ^⊤ α_k^(t)) for k = 1, ..., ∞ and t = 1, ..., T
    Choose a minibatch 𝓑 of documents
    for each document d in 𝓑 do
        Sample θ_d ∼ q(θ_d | η_td, w_d)
        Compute v = logistic(θ_d)
        Compute π_d = { v_1 if k = 1 ; v_k Π_{j<k} (1 − v_j) for k > 1 }
        for all word w_dn in the document do
            Compute Pr(w_dn | π_d) = Σ_k π_dk β_{k,w_dn}^(t_d)
        end for
    end for
end for
Estimate the ELBO and its gradient through backpropagation
Update the model and variational parameters using Adam
```

The D-EDP uses data subsampling [Hof+13] and amortization [GG14] to scale to large datasets and reduce the number of variational parameters. Likewise, the variational objective does not include non-conjugate distributions. The ELBO is the following [2]:

$$\mathrm{ELBO}(\nu) = \mathbb{E}_q \left[\log \Pr(\mathbf{w}, \theta, \eta, \alpha) - \log q_\nu(\theta, \eta, \alpha)\right] \tag{4.15}$$

In the ELBO, $\Pr(\mathbf{w}, \theta, \eta, \alpha)$ is a function of the data, whereas $q_\nu(\theta, \eta, \alpha)$ is a structured variational family. Following the D-ETM's authors, we introduce the following approximation:

$$q(\theta, \eta, \alpha) = \prod_d q\left(\theta_d \mid \eta_{t_d}, \mathbf{w}_d\right) \times \prod_t q\left(\eta_t \mid \eta_{1:t-1}, \widetilde{\mathbf{w}}_t\right) \times \prod_k \prod_t q\left(\alpha_k^{(t)}\right) \tag{4.16}$$

$q\left(\theta_d \mid \eta_{t_d}, \mathbf{w}_d\right)$ represent the distribution over the topic proportions. This distribution is a Gaussian distribution whose mean and covariance parameters are functions of the latent mean $\eta_{t_d}$ and the bag-of-words representation of the $d$-th document $\mathbf{w}_d$. Also, feed-forward neural networks whose inputs are both $\eta_{t_d}$ and $\mathbf{w}_d$ parameterize them. The distributions

2. We disclose the formula details in Appendix A.

over the latent means $q\left(\eta_t \mid \eta_{1:t-1}, \widetilde{\mathbf{w}}_t\right)$ depend on all previous latent means $\eta_{1:t-1}$ and the normalized representation of documents with the $t$-th timestamp, $\widetilde{\mathbf{w}}_t$. A generative LSTM parameterizes these Gaussian distributions. Last but not least, and contrasting with the terms before that use structured variational inference, we use the mean-field family for the topic embeddings $q\left(\alpha_k^{(t)}\right)$.

## 4.3 Empirical studies

This section tests and compares these models on two benchmark and two industrial datasets. We start by giving precisions about our test metrics before turning to the actual experimentations. These industrial datasets respectively contain documents in English and French.

### 4.3.1 Metrics

Our experimentations include a statistical assessment and a quality assessment. We use perplexity for the statistical goodness-of-fit. For quality (TQ), we use the product of diversity (TD) and mean NPMI (TC). We include an additional human evaluation by a domain expert for the industrial dataset and text annotation for the benchmark datasets to evaluate coherence. Please refer to Section 2.1.1.3 for more details.

### 4.3.2 Benchmark datasets

#### 4.3.2.1 Description and data preparation

We use two datasets for model benchmarking: the UN debates corpus [BDM17] and the ACL Anthology corpus [Bir+08]. The UN debates corpus contains forty-six years of speeches by leaders presenting their government's perspective on major global issues. The corpus contains the transcription of these statements for each represented country at the UN General Assembly. On the other hand, the ACL Anthology corpus is a collection of articles discussing issues in computational linguistics and natural language processing from 1973 to 2006. We use the preprocessed versions of these corpora delivered in the GitHub

| Dataset | Number of documents | Train (85%) | Validation (5%) | Test (10%) | Number of time slices | Vocabulary size |
|---|---|---|---|---|---|---|
| *UN* | 230950 | 196290 | 11563 | 23097 | 46 | 12466 |
| *ACL* | 10514 | 8936 | 527 | 1051 | 31 | 35108 |

Table 4.1 – Benchmark datasets' characteristics

repository indicated in [DRB19] [3]. Unfortunately, the Science dataset is not among them.

For the recall, [DRB19] reportedly apply standard preprocessing techniques such as tokenization and number and punctuation marks removal. They considered words in more than 70% of the documents as stopwords and other words present in a provided list. They also removed low-frequency words. For the UN debates corpus, these low-frequency words are the words that appear in less than 30% of the corpus and 10% of the corpus for the ACL Anthology corpus. They retained 85% randomly chosen documents for training, 10% for testing, and 5% for validation. Last but not least, the authors excluded one-word documents from the testing and validation subsamples of the datasets.

#### 4.3.2.2 Training settings

We compare our D-EDP against two variants of D-LDA and the D-ETM. The results proceed from [BL06] for the D-LDA and from [DRB19] for the D-LDA-REP and the D-ETM. The D-LDA-REP model is the same as the D-LDA, except for the inference algorithm. Dieng & al. devised a different D-LDA inference algorithm to separate the performance gains caused by the actual modeling and that caused by the inference algorithm. We follow the same settings as [BL06] to parameterize our D-EDP for model comparison. We reproduce these settings below.
We set the variances of the priors to $\delta^2 = \sigma^2 = \gamma^2 = 0.005$ and $a^2 = 1$. We used a batch size of 200 documents for the UN dataset and 100 documents for the ACL dataset. Contrarily to [DRB19], we did not use pre-fitted word embeddings for the D-EDP on the UN dataset and obtained word representations thanks to the model. We use a multilayer perceptron (MLP) with ReLU activations and two layers of 800 hidden units each for the topic proportions $\theta_d$. Linear maps of the MLP's output parameterize the Gaussian for the topic proportions.

3. `https://github.com/adjidieng/DETM`

We regularize the MLP before outputting values for the mean and (log)-variance through dropout with a rate of 0.1. Regarding the time dynamics-related latent means $\eta_{1:T}$, we map each (normalized) bag-of-words representation $\tilde{\mathbf{w}}_t$ to a low-dimensional space of dimensionality 400. We then feed the output to a 4-layer LSTM with 400 hidden units each. We concatenated the LSTM's output with the previous latent mean $\eta_{t-1}$ and mapped the results to a $K$-dimensional space to get the mean and log-variance for $\eta_t$. For parametric topic models, $K$ refers to the number of topics. In [DRB19], the authors train the model for $K = 50$ components. On the other hand, this number of topics is theoretically infinite for our nonparametric topic model. Consequently, we truncate the topic to $K$ components. This truncation is equivalent to giving the model a maximum capacity for $K$ components, not determining the number of topics. The Dirichlet Processes "neutralize" the additional topics by attributing them a value of 0 for a given corpus. We determined the number of $K = 10$ components with cross-validation in a held-out perplexity (or log-likelihood) task.

We applied Alg. 14 for model training for a maximum of 400 epochs on the UN dataset and a maximum of 1000 epochs for ACL, with learning rates of 0.001 and 0.0008, respectively. We used KL-annealing to get adaptive learning rates during the training step. We further regularized the model with weight decay on all network parameters. This applied weight decay is of $1.2 \times 10^{-6}$. We applied a gradient clipping of 2.0 on the ELBO to stabilize training. Last, the stopping criterion depends on a held-out perplexity task on the validation set.

Concerning the D-LDA, Dieng & al. [DRB19] used the implementation provided with [BL06]. Dieng & al. report that the D-LDA has scalability issues: it took almost two days on each dataset versus less than 6 hours for the D-ETM. Circumventing the scalability issue, Dieng & al. [DRB19] used an alternative algorithm [4]. The authors used a coordinate-ascent algorithm that involves a Kalman filter. They used a reparameterization-based stochastic optimization algorithm with batches of 1000 documents each. They initialized both variants of the D-LDA with LDA, emphasizing that they ran five epochs of LDA followed by 120 epochs of D-LDA. They used the RMSProp algorithm to set the step size, thus setting the learning rate to 0.05 for the mean and 0.005 for the variance parameters.

4. The authors do not provide this variant in the GitHub repository

#### 4.3.2.3 Results and discussion

Unfortunately, the authors from [DRB19] need to provide their qualitative results on the ACL dataset and the full extent of their results on the UN dataset. Consequently, we voluntarily restrain our comments to the results contained in [DRB19] and to our own. In this section, we only display the D-EDP's results on the UN dataset (Fig. 4.4 and Tab. 4.4) to avoid visual clutter. However, the rest of the results, including those of [DRB19], is available in Appendix B.

**4.3.2.3.1 Quantitative evaluation** Quantitative results show that the D-EDP outperforms all the results reported by Dieng & al. [DRB19] regarding topic quality (TQ) on the UN dataset (Tab. 4.2). The overall topic coherence (TC) explains the results on this particular dataset; it falls second, however, to the D-ETM with 50 topics in terms of both perplexity and topic diversity (TD). The D-EDP has detected eight topics on the UN dataset, considering a maximum capacity of ten topics. To set both the D-ETM and the D-EDP in an equal setting, we ran a new experiment on the D-ETM with a maximum capacity of ten topics. The results show that the D-ETM with ten topics outperforms all models in terms of perplexity, TD, and TQ, but not on TC, where the D-EDP still ranks first. The TQ of the D-ETM with ten topics is only marginally superior to that of the D-EDP. We reproduced this closeness between both settings on the ACL dataset (Tab. 4.3).
The D-ETM with 15 topics exhibits the best perplexity compared with all the models, including its counterpart with 50 topics. However, it falls second to the D-EDP in terms of topic quality. The model that achieves the best topic quality is the D-ETM with 50 topics. The performance, however, seemingly came at a cost. The D-ETM with 50 topics has better quality but exhibits a perplexity of slightly less than double that of the D-ETM and the D-EDP. The gains in terms of quality look marginal in comparison. The topic diversity metric is the factor that most explains the difference in quality. For the recall, this metric serves to identify topic uniqueness, i.e., a fundamental criterion for topic interpretability for a human being [Cha+09] along with topic coherence. Considering topic coherence and within the limits of the mean NPMI pertinence, the D-LDA, the D-ETM settings, and the D-EDP match concepts equally well on the ACL dataset. The D-LDA, however, did not discriminate topics compared with the D-ETM with 50 topics, when the capacity of 50 topics should allow for more room for topic distinction than a capacity of 15 topics, hence a higher value for TD. In other words, the granularity is finer. However, whether this additional granularity makes sense is a question we cannot answer as [DRB19] did not provide all the results on the ACL dataset.

| Model | No. of topics (capacity) | Perplexity | TC | TD | TQ |
|---|---|---|---|---|---|
| *D-LDA [BL06]* | 50 (50) | 2393.5 | 0.1317 | 0.6065 | 0.0799 |
| *D-LDA-REP [DRB19]* | 50 (50) | 2931.3 | 0.1180 | 0.2691 | 0.0318 |
| *D-ETM [DRB19]* | 50 (50) | 1970.7 | 0.1206 | 0.6703 | 0.0809 |
| *D-ETM* | 10 (10) | **1720.9** | 0.1814 | **0.6650** | **0.1206** |
| *D-EDP* | 8 (10) | 2069.8 | **0.2033** | 0.5677 | 0.1154 |

Table 4.2 – Results on the UN dataset

| Model | No. of topics (capacity) | Perplexity | TC | TD | TQ |
|---|---|---|---|---|---|
| *D-LDA [BL06]* | 50 (50) | 4324.2 | 0.1429 | 0.5904 | 0.0844 |
| *D-LDA-REP [DRB19]* | 50 (50) | 5836.7 | 0.1011 | 0.2589 | 0.0262 |
| *D-ETM [DRB19]* | 50 (50) | 4120.6 | **0.1630** | **0.8286** | **0.1351** |
| *D-ETM* | 15 (15) | **2360.8** | 0.1401 | 0.6347 | 0.0891 |
| *D-EDP* | 15 (15) | 2492.7 | 0.1504 | 0.6321 | 0.0951 |

Table 4.3 – Results on the ACL dataset

In the case of parametric [5] topic models, contrary to their nonparametric counterparts, there is no implicit assumption for topic ranking. They can exhibit more significant proportions than others, but there is an implicit assumption of equivalence in terms of importance. The adequacy of such arbitration depends on practitioners' criteria according to their use case. This arbitration goes well beyond the scope of this work.

Consequently, it is not easy to posit that the D-ETM with 50 topics achieves more valuable results than its counterpart with 15 topics or the D-EDP. On the contrary, we can affirm that it discriminates better than the D-LDA on both the UN and the ACL dataset.

**4.3.2.3.2 Qualitative evaluation** The D-EDP extracted eight topics from the UN dataset. Several topics show similarity with those extracted by the D-ETM as reported

5. Concerning the number of topics.

| **Reference word** | **Nearest neighbors** |
|---|---|
| *economic* | development, countries, social, order, international, level, political, cooperation, world, institutions, economy, based, natural, part, developing, system, community, national, promote |
| *assembly* | general, session, united, delegation, work, nations, role, member, mr, secretarygeneral, success, express, members, behalf, organization, deliberations, confident, important, election |
| *security* | peace, council, states, united, continue, part, nations, international, view, make, efforts, community, process, members, hope, time, support, made, regard |
| *management* | reform, efficient, financial, reforms, priority, resources, efficiency, improve, enhance, primary, report, structural, programmes, performance, sustainable, budget, contributions, effective, effectively |
| *debt* | servicing, debts, fund, indebtedness, developing, growth, cancellation, earnings, creditors, developed, creditor, exports, economies, income, product, rescheduling, debtservicing, commodity, macroeconomic |
| *rights* | human, fundamental, rule, dignity, respect, discrimination, freedom, justice, equality, freedoms, democracy, society, principles, violations, constitution, protection, based, law, citizens |
| *africa* | african, south, southern, israel, apartheid, continent, middle, situation, east, continues, assistance, policies, palestinian, support, continue, implementation, africans, arab, national |

Table 4.4 – Word embeddings extracted by the D-EDP on the UN dataset

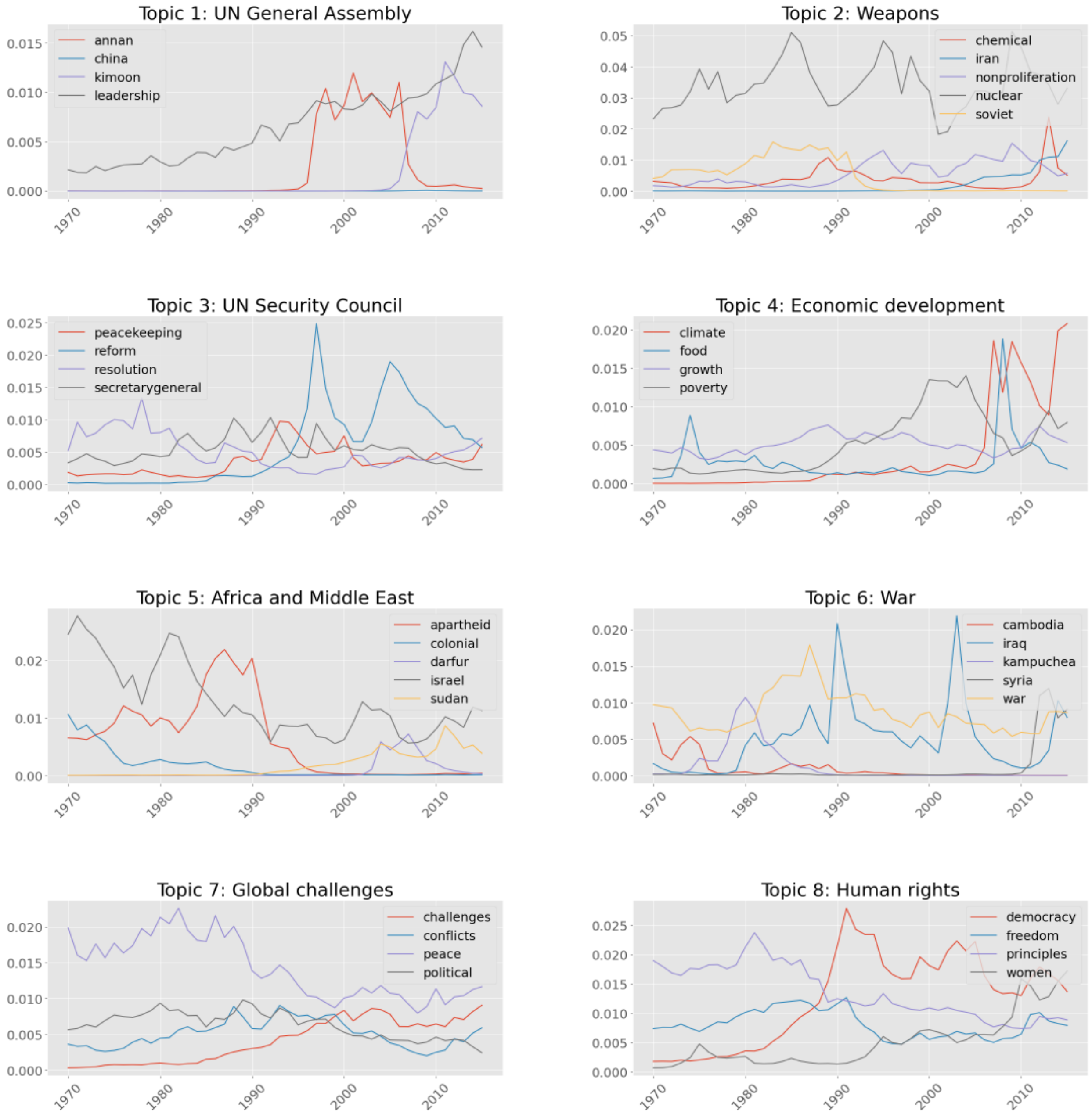

Figure 4.4 – Evolution of word probability for all the topics extracted by the D-EDP on the UN dataset

in [DRB19] (Fig. B.1 in Appendix B) [6]. For instance, our topics about weapons (topic 2),

6. These results are only an excerpt from those obtained by the authors, as they reportedly trained their model for 50 topics.

economic development (topic 4), Africa and Middle East (topic 5), war (topic 6), and human rights (topic 8) look like topics "Nuclear Weapons," "Poverty & Development," "Africa," "War," and "Human Rights." The D-EDP, however, seems to fusion several of the shown topics in a single. For instance, the topic about economic development (topic 4) looks like both the "Poverty & Development" and "Climate Change" topics from the D-ETM at the same time.
Overall, the D-EDP finds that the topics' evolution over time corresponds to historical events[7]. For instance, in topic 1, we can see that the probability for the word "annan" rises and stays relatively stable during Kofi Annan's mandate at the UN Assembly, then decays when the probability for the word "kimoon," corresponding to Ban-Ki-moon, rises. In topic 2, we can see that the probability for the word "soviet" fell during the 1990s and that chemical weapons became more of a concern starting in 2010. Likewise, the apartheid disappears during the early 1990s (topic 5), and the probabilities for the words "cambodia" and "kampuchea" (topic 6) evolve in opposite ways at the same time. Last but not least, the word "freedom" (topic 8) stays approximately at the same level across all time slices.

#### 4.3.2.4 Conclusion

This section presents the theoretical foundations for the D-EDP, a new time-aware topic model, and its statistical specifications. We also ran a set of experiments on benchmark datasets and found that our model yields meaningful results in terms of both topics and word embeddings. When not higher, the D-EDP achieves results close to the best models regarding perplexity and topic quality. The difference between models is globally marginal on the UN and ACL datasets, yet, at the very least, the D-EDP selects the number of topics from the data. Concluding on this situation, we first and foremost express concerns about the adequacy of the current mainstream indicators used to evaluate topic models. In the next section, we test the D-ETM and the D-EDP on two industrial datasets regarding the automotive industry. We decided to exclude the D-LDA due to the required running time. According to [DRB19], it took around two days to train on benchmark datasets, thus making it inappropriate for industrial settings. On the other hand, the D-ETM and the

7. Note that we do not specifically display the words with the highest probability in each topic. Because of our static support, we have chosen words that allow us to capture both the meaning of the topics and the temporal dynamics.

| Dataset | Number of documents | Number of time slices | Document length | | | | | |
|---|---|---|---|---|---|---|---|---|
| | | | *Min.* | *Q1* | *Median* | *Mean* | *Q3* | *Max.* |
| *English* | 97714 | 61 | 1 | 23 | 40 | 53.05 | 66 | 2997 |
| *French* | 63590 | 112 | 1 | 17 | 33 | 57.9 | 65 | 3690 |

Table 4.5 – Industrial datasets' characteristics prior to preprocessing

D-EDP took approximately the same running time, i.e., less than six hours each. For these reasons, we believe the two models are the best candidates for in-house deployment.

### 4.3.3 Industrial datasets

#### 4.3.3.1 Description and data preparation

We use two industrial datasets from an in-house use case regarding the automotive industry, with a particular emphasis on tires. These datasets come from a web scraping commercial activity performed on several websites. The difference between these datasets is their reference markets and the writing language: English and French. Otherwise, they include all the company's available material for these languages. For simplicity, we will refer to these datasets as EN and FR, respectively. We applied the same preprocessing steps whenever possible. However, we could not guarantee that there were no documents in languages other than the main one in each of these datasets are our filtering is source-based. The descriptions below apply to all the datasets unless otherwise specified.
Web scraping implies that no API is involved in the data collection process. Consequently, the corpora originally came along with source-specific document structures. Contrasting with our previous work on synchronic topic modeling, we have set up an ad-hoc pipeline for industrial topic modeling based on the previous one. This pipeline also performs DOM parsing and proceeds to remove as many markup languages and URIs as possible. We then perform POS-tagging using spaCy [Hon+19]. We use the pre-trained transformers (one per language) provided by Explosion AI [8] with the highest number of parameters for the POS-tagging step. spaCy also allows us to get lemmatized versions of words. We retain all the lemmas except the ones corresponding to punctuation, numerals, auxiliaries, and determiners.

8. `https://spacy.io/models`

| Dataset | Number of documents | Train (85%) | Validation (5%) | Test (10%) |
|---|---|---|---|---|
| *English* | 97714 | 83056 | 4887 | 9771 |
| *French* | 63590 | 54051 | 6359 | 3180 |

Table 4.6 – Industrial datasets' splitting

Before splitting datasets into three subsets for training, validating, and testing, we study them in terms of time slice-wise document counts and document length. Similarly to our synchronic topic models, our basic unit for the document length study is the token. The reasons for doing so are the same. The retained granularity for the time slices is the month. Our datasets comprise all the available documents for each language, i.e., 97714 documents for English and 63590 for French. Table 4.5 shows that the document lengths are very close. They follow a Poisson distribution with parameter $\lambda = 53.03$ and $\lambda = 57.9$, respectively. The distribution is a good fit for more than 85% of the documents in each case. We discard all the documents whose length does not fall within an interval of 20 to 200 words, bounds included. However, while we have more documents for the English dataset, we have fewer time slices than for the French dataset. We split our datasets into 85% randomly chosen documents for training, 10% for testing, and 5% for validation. Our $V$-vocabulary sizes are 13364 English tokens and 13421 French tokens.

#### 4.3.3.2 Training settings

We compare our D-EDP against the D-ETM with the same set of common parameters on both datasets. As previously, we set the variances of the priors to $\delta^2 = \sigma^2 = \gamma^2 = 0.005$ and $a^2 = 1$. We used a batch size of 200 documents. Contrarily to [DRB19], we did not use pre-fitted word embeddings to init the D-EDP nor the D-ETM models and trained them with 300-dimensional topic and word embedding layers. We use a multilayer perceptron (MLP) with ReLU activations and two layers of 800 hidden units each for the topic proportions $\theta_d$. Linear maps of the MLP's output parameterize the Gaussian for the topic proportions. We regularize the MLP before outputting values for the mean and (log)-variance through dropout with a rate of 0.5. Regarding the time dynamics-related latent means $\eta_{1:T}$, we map each (normalized) bag-of-words representation $\tilde{\mathbf{w}}_t$ to a low-dimensional space of dimensionality 200. We then feed the output to a 3-layer LSTM with 300 hidden units each. We concatenated the LSTM's output with the previous latent mean $\eta_{t-1}$ and mapped the results to a $K$-dimensional space to get the mean and log-variance for $\eta_t$. We set the number of topics, $K$, to 50 and 10 for the D-ETM and 10 for the D-EDP. Again, this truncation is equivalent to giving the model a maximum capacity for $K$ components, not determining the number of topics. We determined the truncation level and all the other common parameters with cross-validation in a held-out perplexity (or log-likelihood) task. We applied Alg. 14 for model training for a maximum of 50 epochs on both datasets, with a learning rate of 0.005. We used KL-annealing to get adaptive learning rates during the training step. We further regularized the model with weight decay on all network parameters. This applied weight decay is of $1.2 \times 10^{-6}$. Last, the stopping criterion depends on a held-out perplexity task on the validation set.

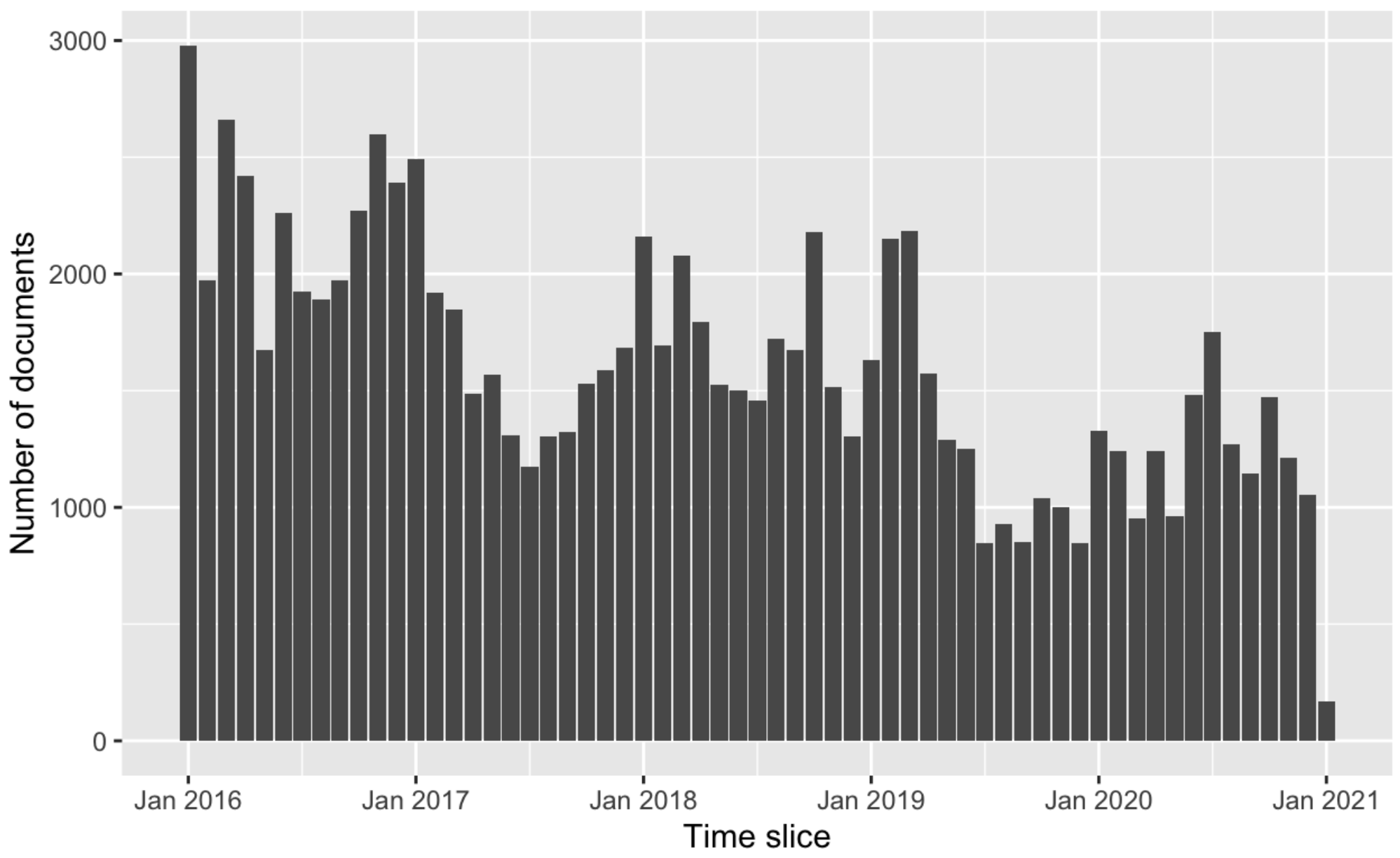

Figure 4.5 – Number of English documents per-time slice

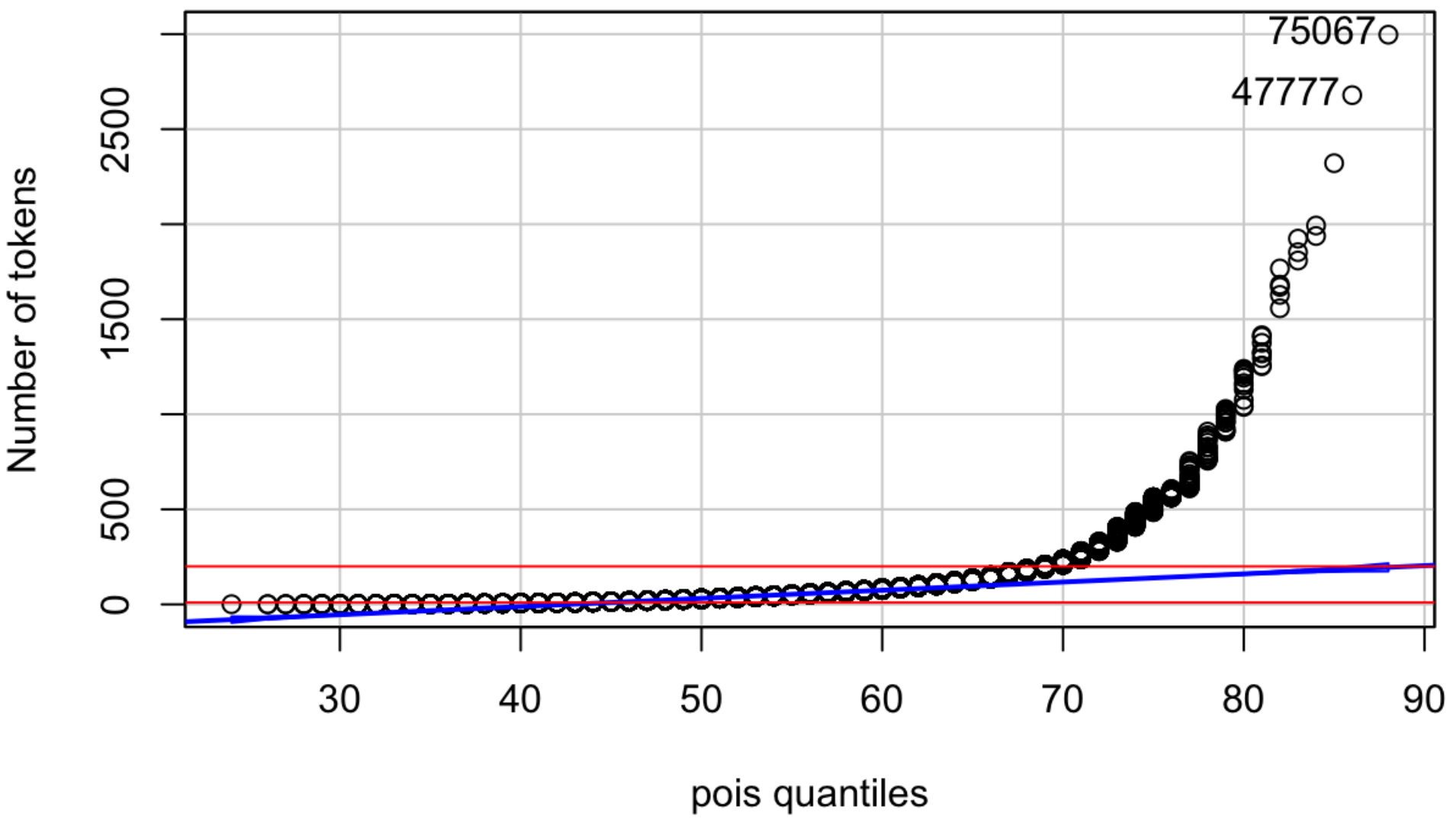

Figure 4.6 – Q-Q plot for the English documents' length against a Poisson($\lambda = 53.03$)

Figure 4.7 – EN dataset's characteristics

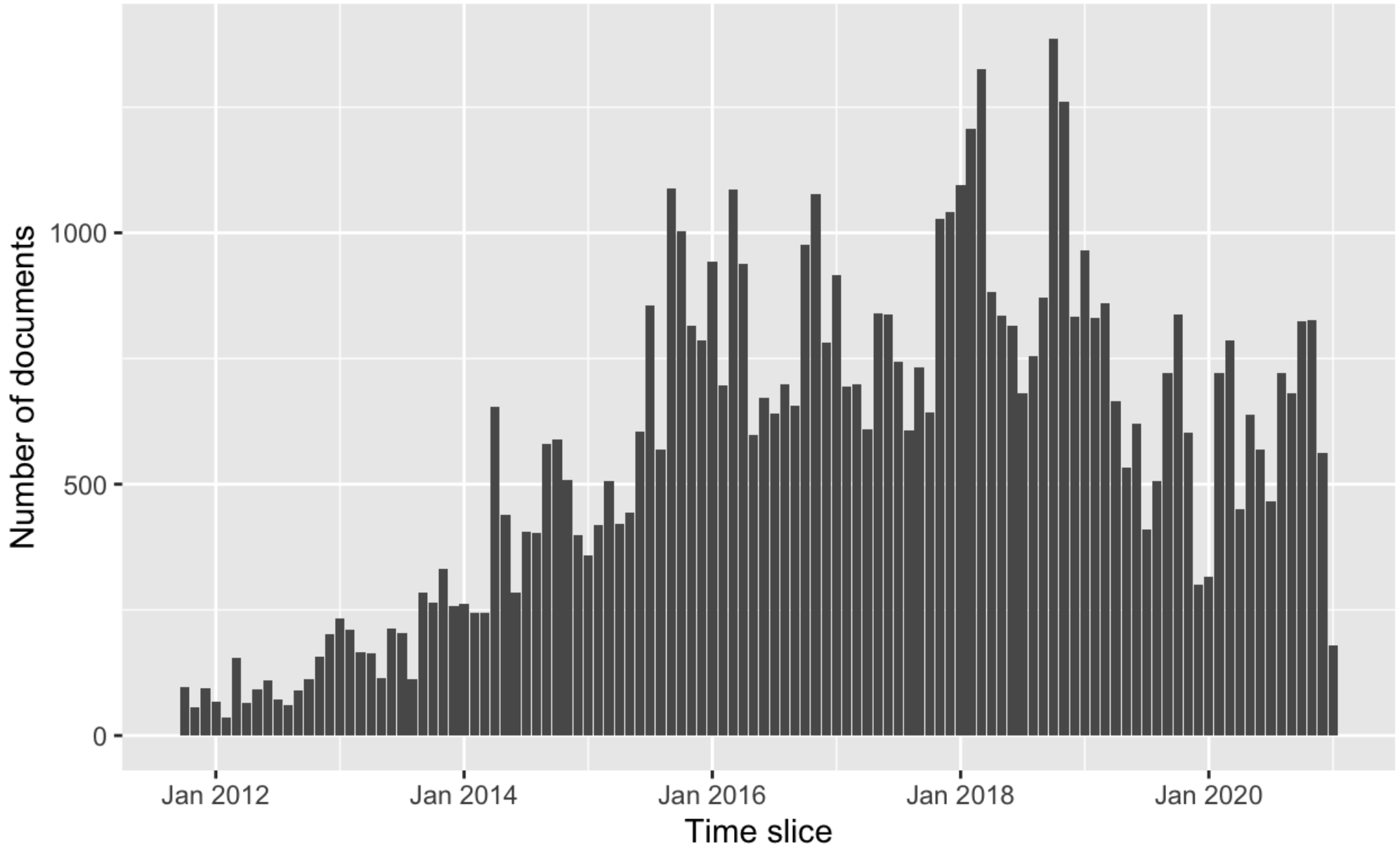

Figure 4.8 – Number of French documents per-time slice

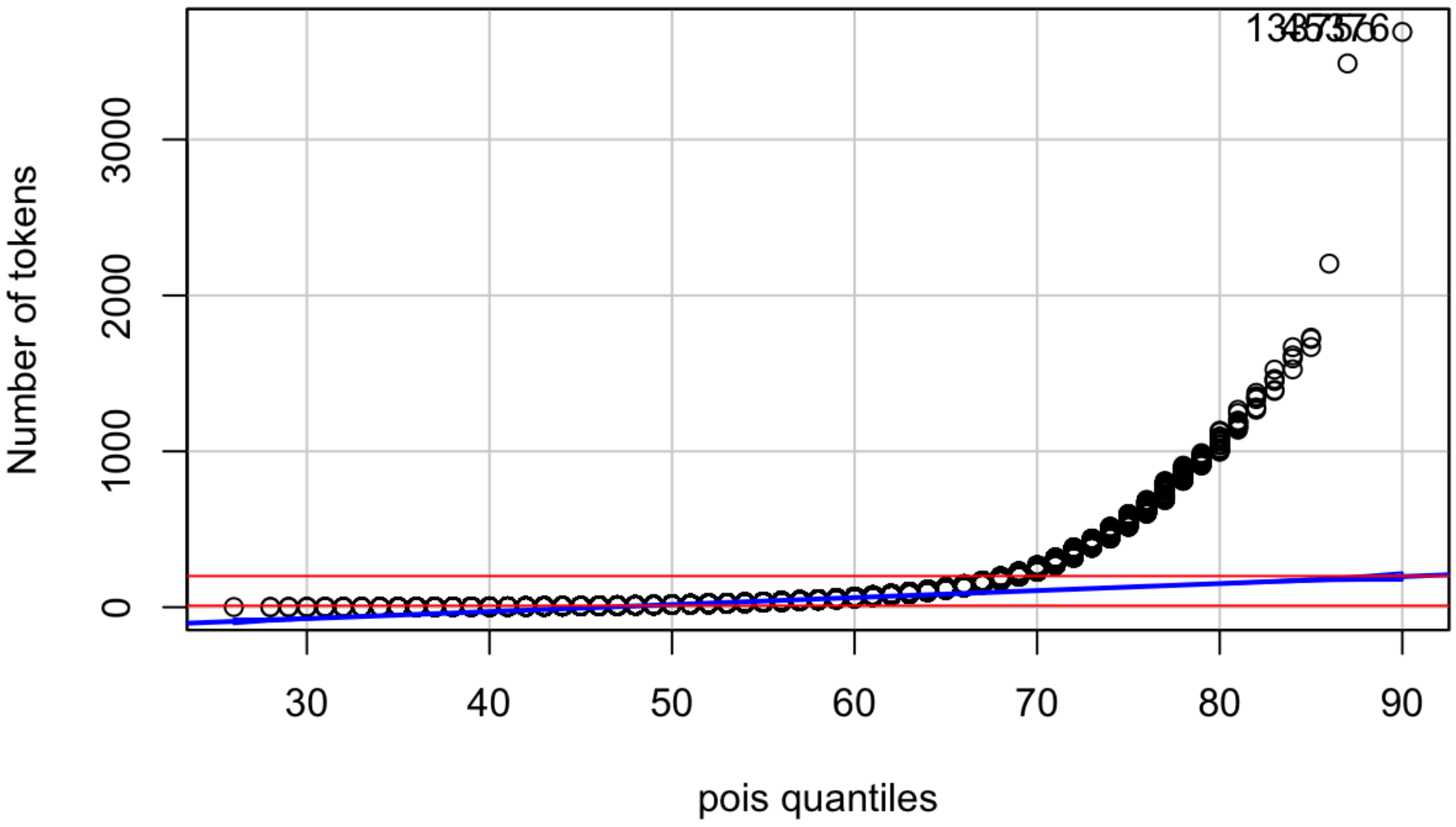

Figure 4.9 – Q-Q plot for the French documents' length against a Poisson($\lambda = 57.9$)

Figure 4.10 – FR dataset's characteristics

#### 4.3.3.3 Results and discussion

**4.3.3.3.1 Quantitative evaluation** The D-ETM with 50 topics and the D-EDP exhibit similar perplexities but differ significantly in terms of quality on the EN dataset (Tab. 4.7). Despite displaying close topic coherence, the D-ETM struggled to discriminate between different topics. However, the D-ETM with ten topics is much closer to the D-EDP yet less performant in terms of quality. It is, nonetheless, the best model in terms of perplexity. The D-ETM with ten topics also shows a lower perplexity metric than its counterparts and is still better than the D-ETM with 50 topics in terms of quality on the FR dataset (Tab. 4.9). However, its topic diversity performance is only marginally better than that of the D-EDP; similarly, the D-EDP is only marginally better than the D-ETM with ten topics in terms of coherence.

**4.3.3.3.2 Qualitative evaluation** Except for specific topics and concepts, the models extract relatively stable elements [9]. The most variable topics are driving conditions-related topics (topics 1 and 1 from 4.11 and 4.12, respectively) and R&D-related topics (topics 6 and 5 from Fig. 4.11 and Fig. 4.12, respectively). For instance, the words for snow and winter show seasonal probability patterns during winter periods. Likewise, prizes awarding (topic

9. Note that we do not specifically display the words with the highest probability in each topic. Because of our static support, we have chosen words that allow us to capture both the meaning of the topics and the temporal dynamics.

| Model | No. of topics (capacity) | Perplexity | TC | TD | TQ |
|---|---|---|---|---|---|
| *D-ETM* | 50 (50) | 1295.3 | 0.2271 | 0.0889 | 0.0202 |
| *D-ETM* | 10 (10) | **1070.9** | 0.2029 | 0.5790 | 0.1175 |
| *D-EDP* | 6 (10) | 1251.9 | **0.2425** | **0.6848** | **0.1660** |

Table 4.7 – Results on the EN dataset

six from Fig. 4.11) show one-off peaks. Last but not least, we can observe an increasing interest in alternative energy sources for car engines. In particular, the topic of electric engines has been gaining momentum during the last months (topic six from Fig. 4.11 and topic five from 4.12).

#### 4.3.3.4 Conclusion

This Section presents the second set of experiments involving the D-ETM and the D-EDP. The D-ETM with ten topics achieves the best perplexity on both the EN and FR datasets. It also marginally outperforms the D-EDP in terms of topic quality on the FR dataset but exhibits slightly lower quality than its nonparametric counterpart on the EN dataset. Regardless of the models, we show that these approaches can work in at least two languages and still capture topic and language dynamics.

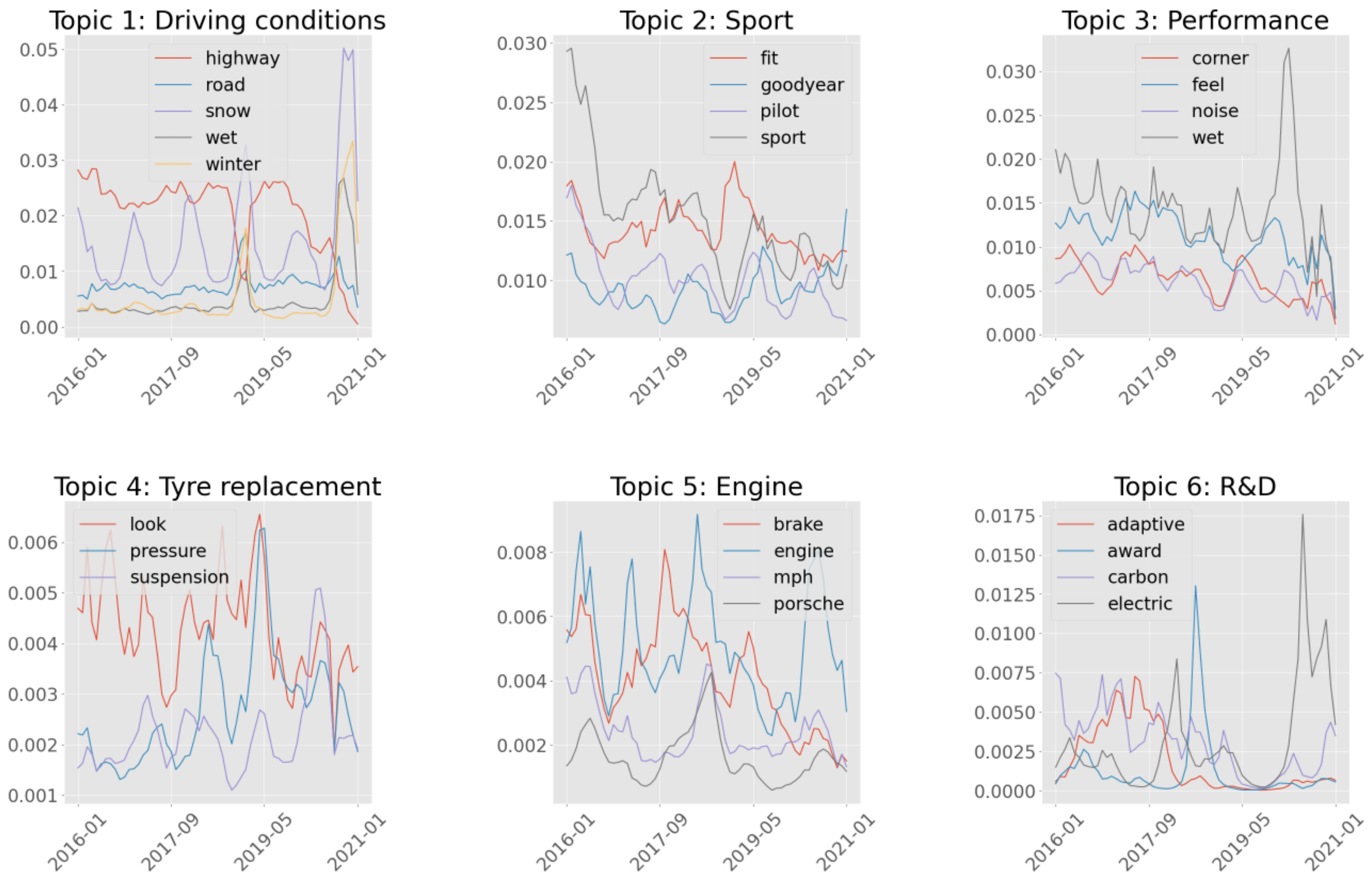

Figure 4.11 – Evolution of word probability for the topics extracted by the D-EDP on the EN dataset

| Reference word | Nearest neighbors |
|---|---|
| *tire* | driving, product, rate, style, vehicle, highway, location, review, city, combine, condition, mile, average, spirited, purchase, traction, drive, rain, truck |
| *handling* | handle, dry, wet, noise, road, snow, drive, comfort, traction, feel, overall, ride, high, performance, short, continental, low, test, good |
| *performance* | road, high, feel, like, noise, low, well, good, long, wet, speed, time, corner, compare, grip, come, day, ride, year |
| *noise* | wet, road, dry, ride, good, wear, drive, buy, well, long, tread, bad, performance, great, like, handle, corner, little, grip |
| *braking* | resistance, distance, test, brake, handling, rolling, read, dry, wet, st, negative, short, aquaplane, overall, positive, straight, comfort, result, th |
| *carbon* | fibre, adjustable, litre, unit, button, leather, exhaust, lightweight, interior, paint, wing, body, seat, chassis, mod, production, splitter, panel, damper |
| *electric* | battery, motor, charge, petrol, engineer, emission, develop, support, plug, seat, renault, gearbox, interior, body, engine, chassis, paint, torque, litre |

Table 4.8 – Word embeddings extracted by the D-EDP on the EN dataset

| Model | No. of topics (capacity) | Perplexity | TC | TD | TQ |
|---|---|---|---|---|---|
| *D-ETM* | 50 (50) | 1984.3 | 0.0860 | 0.3381 | 0.0291 |
| *D-ETM* | 10 (10) | **1534.5** | 0.1544 | **0.5325** | **0.0822** |
| *D-EDP* | 7 (10) | 2110.6 | **0.1595** | 0.4845 | 0.0773 |

Table 4.9 – Results on the FR dataset

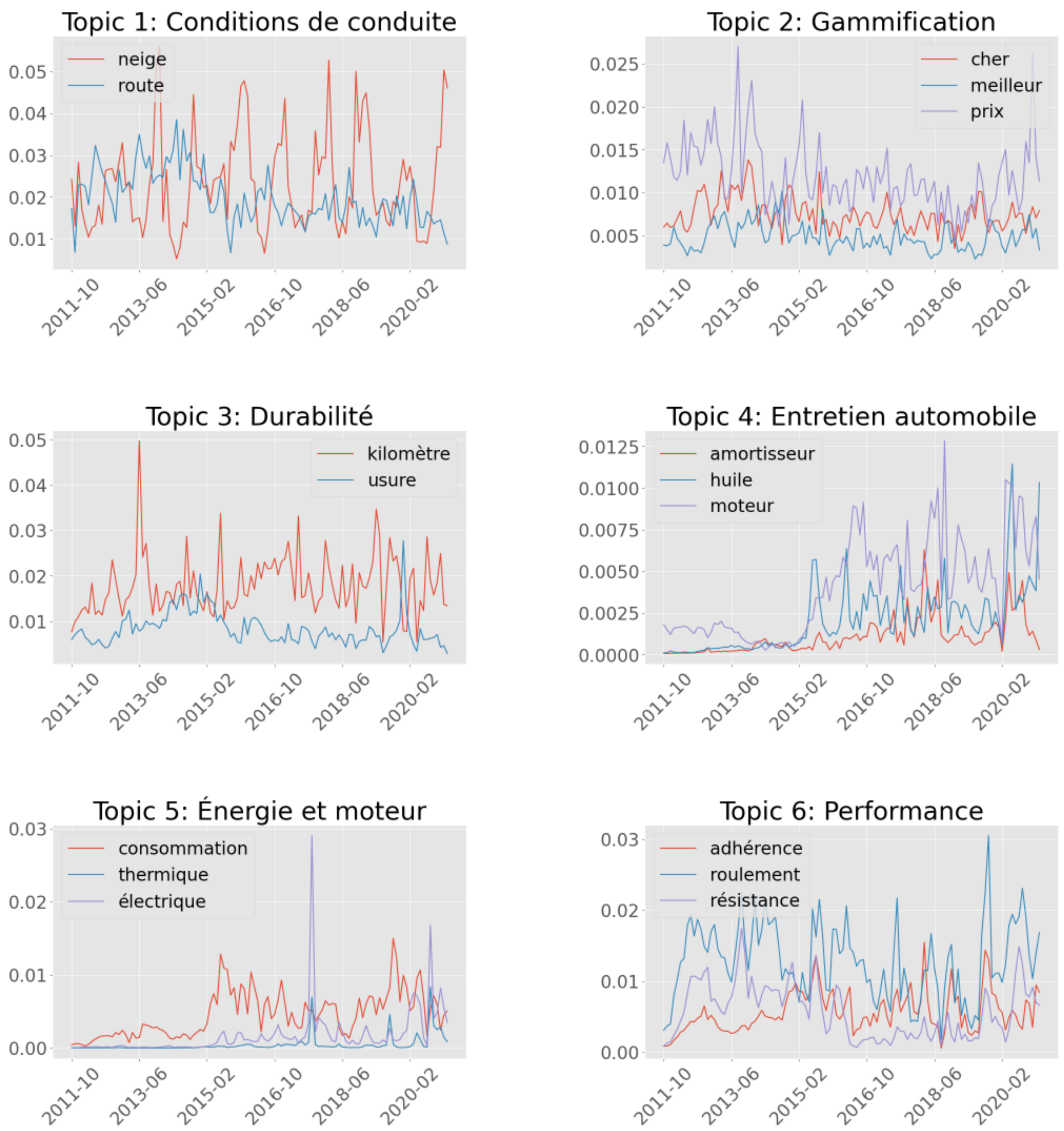

Figure 4.12 – Evolution of word probability for the topics extracted by the D-EDP on the FR dataset

| Reference word | Nearest neighbors |
|---|---|
| *pneu* | qu, bon, route, michelin, bien, hiver, monter, savoir, usure, gomme, monte, changer, très, faire, voir, falloir, sec, pluie, mettre |
| *crossclimate* | crossclimat, saison, hiver, neige, weatherproof, climat, pneu, alpin, climate, vector, gen, pluie, sec, nokian, cross, condition, hivernal, roule, mouiller |
| *performance* | sport, meilleur, sportif, gamme, nouveau, pilot, offrir, marque, dimension, performant, rapport, grip, choix, version, également, modèle, conduite, sec, grand |
| *carbone* | fibre, aileron, capot, diffuseur, émission, aérodynamique, piston, alcantara, carrosserie, m2, aluminium, échappement, forger, calandre, cylindre, jupe, toit, exemplaire, m4 |
| *technologie* | développer, innovation, innovant, matériau, développement, concept, technologique, mondial, mobilité, optimiser, actif, unique, grâce, doter, intégrer, automobiliste, prototype, offrir, automobile |
| *adhérence* | virage, route, pneu, sol, mouiller, gomme, dimension, meilleur, humide, performance, tenue, sec, profil, caractéristique, résistance, pluie, conduite, hiver, condition |
| *électrique* | batterie, autonomie, recharge, véhicule, rechargeable, recharger, moteur, thermique, énergie, système, automobile, tesla, matériau, kwh, conducteur, bord, grâce, affiche, caméra |

Table 4.10 – Word embeddings extracted by the D-EDP on the FR dataset

## 4.4 General conclusion

This Chapter presents the theoretical foundations for the D-EDP, a nonparametric time-aware topic model. This model can capture the number of topics, content, and dynamic and dense representations for topics and words. We tested our model on benchmark datasets and industrial datasets. Our results show that the D-EDP achieves a slightly similar perplexity to its parametric counterpart, the D-ETM, with the latter having as many components as the number of topics detected by the D-EDP. This result is similar to that of [Teh+06]. Both models are globally close regarding topic quality. However, the D-EDP emphasizes the quantitative importance of topics more than its parametric counterpart, while the D-ETM achieves finer granularity than the D-EDP. The impact in terms of practical value is debatable, strongly depends on the use case, and is well beyond this piece of research scope. While the phenomenon enables questioning the adequacy of the mainstream metrics for a topic model's evaluation, and while there is evidence of automatic evaluation being "broken" [Hoy+21], the question is extensible to other fields, including that of supervised learning.

# Chapter 5

# Work review and conclusion

The Variational Autoencoder framework is helpful in several regards. It enables the relatively fast and modular development of scalable algorithms while combining the good properties of Bayesian probabilistic modeling and Deep Learning. In other words, it allows for uncertainty modeling and flexibility. Its application to the task of topic modeling also enables one to account for rich contextual elements.

In this research, we have developed three topic models: the Embedded Dirichlet Process, the Embedded Hierarchical Dirichlet Process, and the Dynamic Embedded Dirichlet Process. Apart from being fully nonparametric regarding the number of topics in a given corpus, these topic models present the particularity of determining dense representations for topics and words. These dense representations, or embeddings, form a log-linear model that allows for simultaneous representation in the same space, thus providing practitioners with additional analytical levels compared with classic topic models. We present two tests for each: one on benchmark datasets and one on industrial datasets. The benchmark datasets are a proof of concept, while our industrial datasets represent our use case. Our models proved successful in both cases, thus demonstrating their performance being on par and even superior to state-of-the-art models. Despite these results, we must find a way to address the deficiencies of current metrics for topic model evaluation. These otherwise already identified deficiencies also exist in our real-world use case. This issue goes well beyond the sole field of topic modeling, and solving it would address many other issues in Natural Language Processing and Understanding. Our industrial context would also benefit from improving existing metrics or designing new indicators.

As stated in our preliminary remarks, this work is part of an industrial framework whose primary objective is discovering customer insight in social media without a business expert. Therefore, the model development priorities we recommend depend strictly on this industrial context.

**Zooming on topic contents** We observed the granularity differential between models and between models and expert annotation throughout our research. Consequently, we consider capturing a hierarchical structure of topics, thus distinguishing topics and subtopics. In other words, we would like to capture a *topical tree*. The idea is not new and appeared in previous research ([WB09]). We believe that a neural adaptation should involve mathematical structures (e.g., neural networks) capable of capturing the hierarchical structure of topics at both the latent variable and the embeddings' levels. The task, however, is not easy. *As far as we know*, in [Goy+17], the authors try to include the nested Chinese Restaurant Process in their VAE-based algorithms. They effectively apply their algorithm to video data instead of text. The setting, however, is fully Gaussian. As demonstrated and despite its flexibility, a Gaussian lacks the desirable properties of Dirichlet-related distributions for text modeling. Additionally, we express concerns regarding its scalability.

**Document-wise word distribution customization** Topic models focus on describing a document with topics that are, in turn, described by words. Document-wise topics' distribution varies depending on the document, but the topic-wise word distributions do not vary. To illustrate our point, let A and B be two separate documents exhibiting identical topic distributions. Suppose these topics are customer reviews about a given product, with customer A being pleased and customer B being utterly unsatisfied. It is implausible that the two customers' way of talking about the same topic is the same, considering their experience. Despite our example being sentiment analysis-related, we focus on advanced document filtering. Such a feature could help customer service and target customers more effectively, but it could also reveal different customer experiences, thus making the model more informative to analysts. We see at least one challenge in adding document-wise information: increasing complexity linearly as the number of documents increases too. We consider the issue a mild hindrance. On the one hand, the amount of documents to consider is foreseeably massive due to the Big Data industrial context. On the other hand, the VAE framework, whether implying alternate optimization or not, uses batch learning. Machine Learning, in general, is about induction and inference from population samples and has never been about deduction from the whole statistical population. However, adding a pseudorandom variable for document indexes can put the generative capabilities and the flexibility of a model at stake.

**Semi-supervised extensions** When developing our models, our starting point was complete ignorance about a given corpus. In practice, this ignorance-is-bliss stance is only partially realistic. Our zero-amount-of-knowledge assumption is a simplifying one that aims at avoiding as much practitioner bias as possible. Whether in a statistical sense or not, an incorrect prior can lead to a wrong inference and subsequent conclusions - including the posterior distribution. As scientific literature demonstrates, human-in-the-loop machine learning is optional to draw contradictory conclusions. These points explain why we precluded human-in-the-loop approaches: practitioners will only sometimes have precise domain experts to check the results [1], can disagree with each other, and make decisions that do not reflect the actual data contents, hence possibly preventing the model from performing proper inference. Bias is the greatest obstacle in analytical projects that involve unsupervised learning in entirely unknown settings. However, knowledge bases will eventually build following iterations based on essential elements or domain knowledge. Such based knowledge is precious. Apart from transfer learning using word embeddings or topic embeddings, it is possible to inject this knowledge through semi-supervised learning to back models' inference process.

1. In our case, this implies having marketing experts for every market who is preferably a native speaker of the language to handle ambiguities and address specificities.

# Appendix A

# Formulas

The following Sections present the Evidence Lower BOunds for all the models. We tried to present each symbol by appearing order and by coherence. We also include and detail each model-specific element. We repeat some elements for convenience.

## A.1 Evidence Lower Bound of the Embedded Topic Model

$$\begin{aligned}
\mathrm{ELBO}(\nu) &= \mathbb{E}_q \left[\log \Pr\left(\mathbf{w} \mid \delta, \rho, \alpha\right)\right] - \mathrm{KLD}\left(q\left(\delta \mid \mathbf{w}, \nu\right) \parallel \Pr\left(\delta\right)\right) \\
&= \frac{1}{S} \sum_{d=1}^{D} \sum_{n=1}^{N_d} \sum_{s=1}^{S} \log \Pr\left(w_{dn} \mid \delta_d^{(s)}, \rho, \alpha\right) \\
&\quad - \sum_{d=1}^{D} \left(\frac{1}{2} \left\{\mathrm{tr}\left(\Sigma_d\right) + \mu_d^{\top} \mu_d - \log\det\left(\Sigma_d\right) - K\right\}\right)
\end{aligned} \tag{A.1}$$

### Notation

— $\nu$ are the inference network's - or encoder's - weights

— $\mathbf{w}$ is the document set and $w_{dn}$ is the $n^{\text{th}}$ word from document $d$

— $\delta$ is the latent variable

— $\rho$ is the word embedding matrix

— $\alpha$ is the topic embedding matrix

— $S$ is the number of samples required to form Monte Carlo estimates of the data log-likelihood (in practice, $S = 1$ has proven enough)

— $D$ is the number of documents and $N_d$ is the number of words in document $d$

— $\mu$ and $\Sigma$ are the variational distribution's parameters (the prior is a standard Gaussian)

— $K$ is the number of topics

## A.2 Evidence Lower Bound of the Stick-Breaking Variational Autoencoder

$$
\begin{aligned}
\mathrm{ELBO}(\nu) &= \mathbb{E}_q\left[\log \Pr\left(\mathbf{w} \mid \pi\right)\right] - \mathrm{KLD}\left(q\left(\pi \mid \mathbf{w}, \nu\right) \parallel \Pr\left(\pi\right)\right) \\
&= \frac{1}{S}\sum_{d=1}^{D}\sum_{n=1}^{N_d}\sum_{s=1}^{S} \Pr\left(w_{dn} \mid \pi_{(s)}^{(d)}\right) \\
&\quad - \left(\begin{aligned} &\frac{a-1}{a}\left(-\gamma - \Psi(b) - \frac{1}{b}\right) + \log ab + \log \mathrm{B}(1, \beta) \\ &\qquad - \frac{b-1}{b} + (\beta - 1) b \sum_{m=1}^{\infty} \frac{1}{m + ab} \mathrm{B}\left(\frac{m}{a}, b\right) \end{aligned}\right)
\end{aligned}
\tag{A.2}
$$

**Notation**

— $\nu$ are the inference network's - or encoder's - weights

— $S$ is the number of samples required to form Monte Carlo estimates of the data log-likelihood (in practice, $S = 1$ has proven enough)

— $D$ is the number of documents and $N_d$ is the number of words in document $d$

— $w_{dn}$ is the $n^{\text{th}}$ word from document $d$

— $\pi$ is the latent variable obtained after applying a stick-breaking process on a sample from the Kumaraswamy variational

— $a$ and $b$ are the Kumaraswamy variational's parameters

— $\beta$ is the GEM prior's parameter (this is the same as $\mathrm{Beta}(1, \beta)$)

— $\gamma$ is Euler's constant

— $\Psi$ is the digamma function

— B is the Beta function

— $m$ is a term for a Taylor expansion (hence the infinite sum)

**Sampling from a Kumaraswamy distribution** Let $x \sim \text{Kumaraswamy}(a, b)$. This is equivalent to:

$$x \sim \left(1 - u^{\frac{1}{b}}\right)^{\frac{1}{a}} \text{ where } u \sim \text{Uniform}(0, 1) \tag{A.3}$$

For convenience, several Python packages include functions that enable sampling from a Kumaraswamy (e.g., TensorFlow Probability [1] and Sympy [2]).

**Stick-breaking process on a Kumaraswamy** Let $v \sim \text{Kumaraswamy}(a, b)$. The stick-breaking process is defined as follows:

$$\pi = \begin{cases} v_1 \text{ if } k = 1 \\ v_k \Pi_{j<k} (1 - v_j) \text{ for } k > 1 \end{cases} \tag{A.4}$$

where $k = 1, \ldots, K$ and $K$ is the truncature level on the number of topics.

## A.3 Evidence Lower Bound of the Embedded Dirichlet Process

$$\begin{aligned} \text{ELBO}(\nu) &= \mathbb{E}_q \left[\log \Pr(\mathbf{w} \mid \pi, \xi)\right] - \text{KLD}\left(q(\pi \mid \mathbf{w}, \nu) \parallel \Pr(\pi)\right) \\ &= \frac{1}{S} \sum_{d=1}^{D} \sum_{n=1}^{N_d} \sum_{s=1}^{S} \Pr\left(w_{dn} \mid \pi_{(s)}^{(d)}, \xi\right) \\ &\quad - \begin{pmatrix} \log \dfrac{\text{B}(a, b)}{\text{B}(1, \beta)} - (a-1)\Psi(1) - (b-\beta)\Psi(\beta) \\ + (a - 1 + b - \beta)\Psi(1 + \beta) \end{pmatrix} \end{aligned} \tag{A.5}$$

1. https://www.tensorflow.org/probability/api_docs/python/tfp/distributions/Kumaraswamy
2. https://docs.sympy.org/latest/modules/stats.html

**Notation**

— $\nu$ are the inference network's - or encoder's - weights

— $S$ is the number of samples required to form Monte Carlo estimates of the data log-likelihood (in practice, $S = 1$ has proven enough)

— $D$ is the number of documents and $N_d$ is the number of words in document $d$

— $\mathbf{w}$ is the document set

— $\xi$ is the topic-word matrix

— $\beta$ is the GEM prior's parameter (this is the same as $\text{Beta}(1, \beta)$)

— $a$ and $b$ are the Beta variational's parameters

— $w_{dn}$ is the $n^{\text{th}}$ word from document $d$

— $\pi$ is the latent variable obtained after applying a stick-breaking process on a sample from the Beta variational

— $\Psi$ is the digamma function

— $B$ is the Beta function

**Sampling from a Beta distribution** In the Embedded Dirichlet Process context, the model implies drawing from a Beta distribution using the implicit reparameterization gradients method. This research presents the implicit reparameterization gradients technique in Section 3.1.3. For convenience, we inform the reader that the implicit reparameterization gradients technique is implemented in Tensorflow Probability [3].

**Stick-breaking process on a Beta** Let $v \sim \text{Beta}(\alpha, \beta)$. The stick-breaking process is defined as follows:

$$\pi = \begin{cases} v_1 \text{ if } k = 1 \\ v_k \Pi_{j<k} \left(1 - v_j\right) \text{ for } k > 1 \end{cases} \tag{A.6}$$

where $k = 1, \ldots, K$ and $K$ is the truncature level on the number of topics.

3. `https://www.tensorflow.org/probability/api_docs/python/tfp/distributions/Beta`

## A.4 Evidence Lower Bound of the Embedded Hierarchical Dirichlet Process

$$
\begin{aligned}
\mathrm{ELBO}(\nu) &= \mathbb{E}_q \left[\log \Pr \left(\mathbf{w} \mid \pi, \xi\right)\right] \\
&\quad + \mathbb{E}_q \left[\log \Pr \left(\nu \mid \beta\right)\right] \\
&\quad - \mathbb{E}_q \left[\log q \left(\nu \mid \mathbf{w}\right)\right] \\
&\quad - \mathrm{KLD}\left(q\left(\beta \mid g_1, g_2\right) \parallel \Pr\left(\beta \mid \gamma_1, \gamma_2\right)\right) \\
&= \frac{1}{S} \sum_{d=1}^{D} \sum_{n=1}^{N_d} \sum_{s=1}^{S} \Pr\left(w_{dn} \mid \pi_{(s)}^{(d)}, \xi\right) \\
&\quad + (K-1)\left(\Psi\left(g_1\right) - \log(g_2)\right) \\
&\quad - \mathrm{KLD}\left(q(a,b) \parallel \Pr(1,\beta)\right) \\
&\quad + \log\left(\mathrm{B}(1,\beta)\right) \\
&\quad + 2\log\left(ab\right) \\
&\quad - \mathrm{KLD}\left(q(g_1,g_2) \parallel \Pr(\gamma_1,\gamma_2)\right)
\end{aligned} \tag{A.7}
$$

**Notation**

— $\nu$ are the inference network's - or encoder's - weights

— $\mathbf{w}$ is the document set

— $\pi$ is the latent variable obtained after applying a stick-breaking process on a sample from the Beta variational

— $\beta$ is the GEM's prior concentration parameter, achieved through $\beta = g_1/g_2$

— $g_1$ and $g_2$ are the Gamma variational's parameters

— $\gamma_1$ and $\gamma_2$ are the Gamma hyperprior's parameters

— $S$ is the number of samples required to form Monte Carlo estimates of the data log-likelihood (in practice, $S = 1$ has proven enough)

— $D$ is the number of documents and $N_d$ is the number of words in document $d$

— $w_{dn}$ is the $n^{\text{th}}$ word from document $d$

— $\xi$ is the topic-word matrix

— $K$ is the truncature level for the Dirichlet Process

— $\Psi$ is the digamma function

— $a$ and $b$ are the Beta variational's parameters

— $\mathrm{B}$ is the Beta function

**KL-Divergence between two Beta distributions** In Eqn. A.7,

$$\mathrm{KLD}\left(q(a,b) \parallel \Pr(1,\beta)\right) \tag{A.8}$$

is the KL-Divergence between two Beta distributions.
Let Pr be a Beta$(a, b)$ and $q$ be a Beta$(c, d)$. The general formula for a KLD between two Beta distributions is the following:

$$\mathrm{KLD}\left(\Pr \parallel q\right) = \log \frac{\mathrm{B}\left(c,d\right)}{\mathrm{B}\left(a,b\right)} - (c-a)\,\Psi\left(a\right) - (d-b)\,\Psi\left(b\right) + (c-a+d-b)\,\Psi\left(a+b\right) \tag{A.9}$$

where B is the Beta function and $\Psi$ is the digamma function. Consequently,

$$\mathrm{KLD}\left(q(a,b)\parallel \Pr(1,\beta)\right) = \log\frac{\mathrm{B}\left(1,\beta\right)}{\mathrm{B}\left(a,b\right)} - (1-a)\,\Psi\left(a\right) - (\beta-b)\,\Psi\left(b\right) + (c-1+\beta-b)\,\Psi\left(a+b\right) \tag{A.10}$$

**KL-Divergence between two Gamma distributions** In Eqn. A.7,

$$\mathrm{KLD}\left(q(g_1,g_2)\parallel \Pr(\gamma_1,\gamma_2)\right) \tag{A.11}$$

is the KL-Divergence between two Gamma distributions.
Let Pr be a Gamma($b_{Pr}, c_{Pr}$) and $q$ be a Gamma($b_q, c_q$). The general formula for a KLD between two Gamma distributions is the following:

$$\begin{aligned}\mathrm{KLD}\left(\Pr\parallel q\right) &= (c_{Pr}-1)\,\Psi\left(c_{Pr}\right) - \log b_{Pr} - c_{Pr} - \log\Gamma\left(c_{Pr}\right)\\ &+ \log\Gamma\left(c_q\right) + c_q\log b_q - (c_q-1)\left(\Psi\left(c_{Pr}\right)+\log b_{Pr}\right) + \frac{b_{Pr}c_{Pr}}{b_q}\end{aligned} \tag{A.12}$$

where $\Gamma$ is the Gamma function and $\Psi$ is the digamma function. Consequently,

$$\begin{aligned}\mathrm{KLD}\left(q(g_1,g_2)\parallel \Pr(\gamma_1,\gamma_2)\right) &= (g_2-1)\,\Psi\left(g_2\right) - \log g_1 - g_2 - \log\Gamma\left(g_2\right)\\ &+ \log\Gamma\left(\gamma_2\right) + \gamma_2\log\gamma_1 - (\gamma_2-1)\left(\Psi\left(g_2\right)+\log g_1\right) + \frac{g_1g_2}{\gamma_1}\end{aligned} \tag{A.13}$$

**Stick-breaking process on a Beta** Let $v \sim \mathrm{Beta}(\alpha,\beta)$. The stick-breaking process is defined as follows:

$$\pi = \begin{cases} v_1 \text{ if } k=1\\ v_k\Pi_{j<k}\left(1-v_j\right) \text{ for } k>1\end{cases} \tag{A.14}$$

where $k = 1,\ldots,K$ and $K$ is the truncature level on the number of topics.

## A.5 Evidence Lower Bound of the Dynamic Embedded Topic Model and the Dynamic Embedded Dirichlet Process

$$
\begin{aligned}
\mathrm{ELBO}(\nu) &= \mathbb{E}_q\left[\log \Pr(\mathbf{w}, \theta, \eta, \alpha) - \log q_\nu(\theta, \eta, \alpha)\right] \\
&= \frac{1}{S}\sum_{d=1}^{D}\sum_{n=1}^{N_d}\sum_{s=1}^{S} \log \Pr\left(w_{dn} \mid \theta_d^{(s)}, \eta, \alpha\right) \\
&\quad - \begin{pmatrix} \sum_d \mathrm{KLD}\left(q\left(\theta_d \mid \eta_{t_d}, \mathbf{w}_d\right) \parallel \mathcal{N}(\eta_{t_d-1}, a^2 I)\right) \\ + \sum_t \mathrm{KLD}\left(q\left(\eta_t \mid \eta_{1:t-1}, \widetilde{\mathbf{w}}_t\right) \parallel \mathcal{N}(\eta_{t-1}, \delta^2 I)\right) \\ + \sum_k \sum_t \mathrm{KLD}\left(q\left(\alpha_k^{(t)}\right) \parallel \mathcal{N}(\alpha_k^{(t-1)}, \gamma^2 I)\right) \end{pmatrix}
\end{aligned} \tag{A.15}
$$

**Notation**

— $\nu$ are the inference network's - or encoder's - weights

— $S$ is the number of samples required to form Monte Carlo estimates of the data log-likelihood (in practice, $S = 1$ has proven enough)

— $D$ is the number of documents and $N_d$ is the number of words in document $d$

— $\mathbf{w}$ is the document set and $\widetilde{\mathbf{w}}$ is the time slice-wise document set

— $w_{dn}$ is the $n^{\text{th}}$ word from document $d$

— $\theta_d$ is the latent variable obtained after applying a softmax normalization on a Gaussian sample in the D-ETM context or a stick-breaking process on a Gaussian sample in a D-EDP context

— $I$ is the identity matrix

— $\eta_d$ and $a^2 I$ are a Gaussian variational's location and scale parameters where $a$ is a hyperparameter

— $\eta_d$ and $\delta^2 I$ are a Gaussian variational's location and scale parameters where $\delta$ is a hyperparameter

— $\alpha_k^{(t)}$ and $\gamma^2 I$ are the Gaussian variational's location and scale parameters where $\gamma$ is a hyperparameter. The $\alpha$ tensor also corresponds to the time-dynamic topic embeddings

**KL-Divergence between two Gaussians** In Eqn. A.15, all the KLD terms correspond to KL-Divergences between two Gaussians.
Let Pr be a $\mathcal{N}(\boldsymbol{\mu}_{\text{Pr}}, \Sigma_{\text{Pr}})$ and $q$ be a $\mathcal{N}(\boldsymbol{\mu}_q, \Sigma_q)$. The general formula for a KLD between two Gaussians is the following:

$$\text{KLD}\left(\text{Pr} \parallel q\right) = \frac{1}{2}\left[\log\frac{|\Sigma_q|}{|\Sigma_{\text{Pr}}|} - k + \left(\boldsymbol{\mu}_{\text{Pr}} - \boldsymbol{\mu}_q\right)^T \Sigma_q^{-1}\left(\boldsymbol{\mu}_{\text{Pr}} - \boldsymbol{\mu}_q\right) + \text{tr}\left\{\Sigma_q^{-1}\Sigma_{\text{Pr}}\right\}\right] \quad \text{(A.16)}$$

**Stick-breaking process on a Gaussian** In the D-EDP's context, the stick-breaking process' weights are transformed samples from the variational distribution drawn using the reparameterization trick [KW14].
Let $z$ be a sample from the latent variable:

$$z = \mu + \sigma \otimes \epsilon \quad \text{and} \quad \epsilon \sim \mathcal{N}(0, \mathbf{I}) \quad \text{(A.17)}$$

In Eqn. A.17, $\otimes$ stands for the Hadamard product. These samples are then squashed to the simplex thanks to a logistic function:

$$v = 1/\left(1 + e^{-z}\right) \quad \text{(A.18)}$$

Finally:

$$\pi = \begin{cases} v_1 \text{ if } k = 1 \\ v_k \Pi_{j<k}\left(1 - v_j\right) \text{ for } k > 1 \end{cases} \quad \text{(A.19)}$$

where $k = 1, \ldots, K$ and $K$ is the truncature level on the number of topics.

# Appendix B

# Additional results on time dynamics modeling

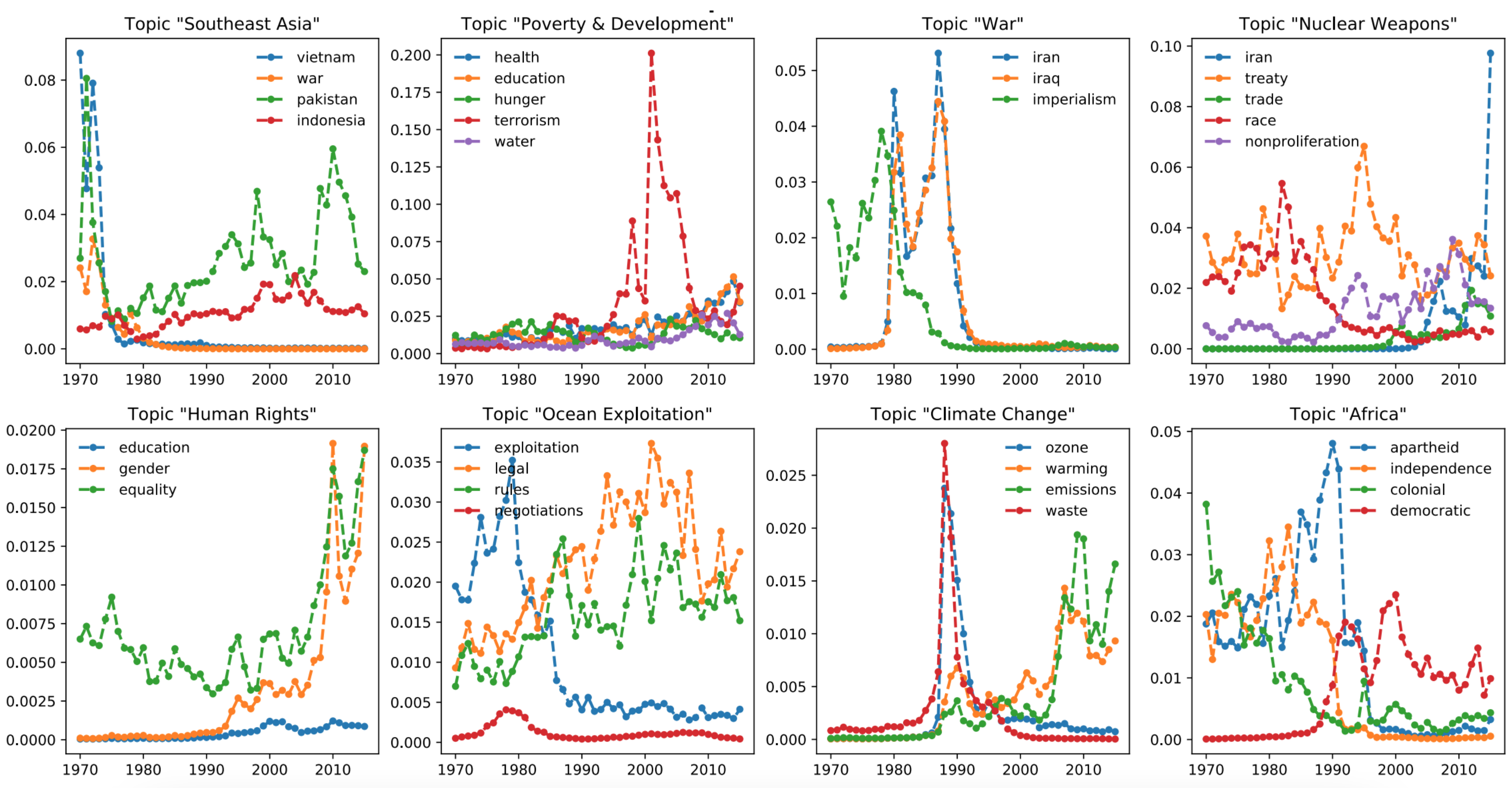

Figure B.1 – Evolution of word probability for eight topics extracted by the D-ETM on the UN dataset according to Dieng & al. [DRB19]

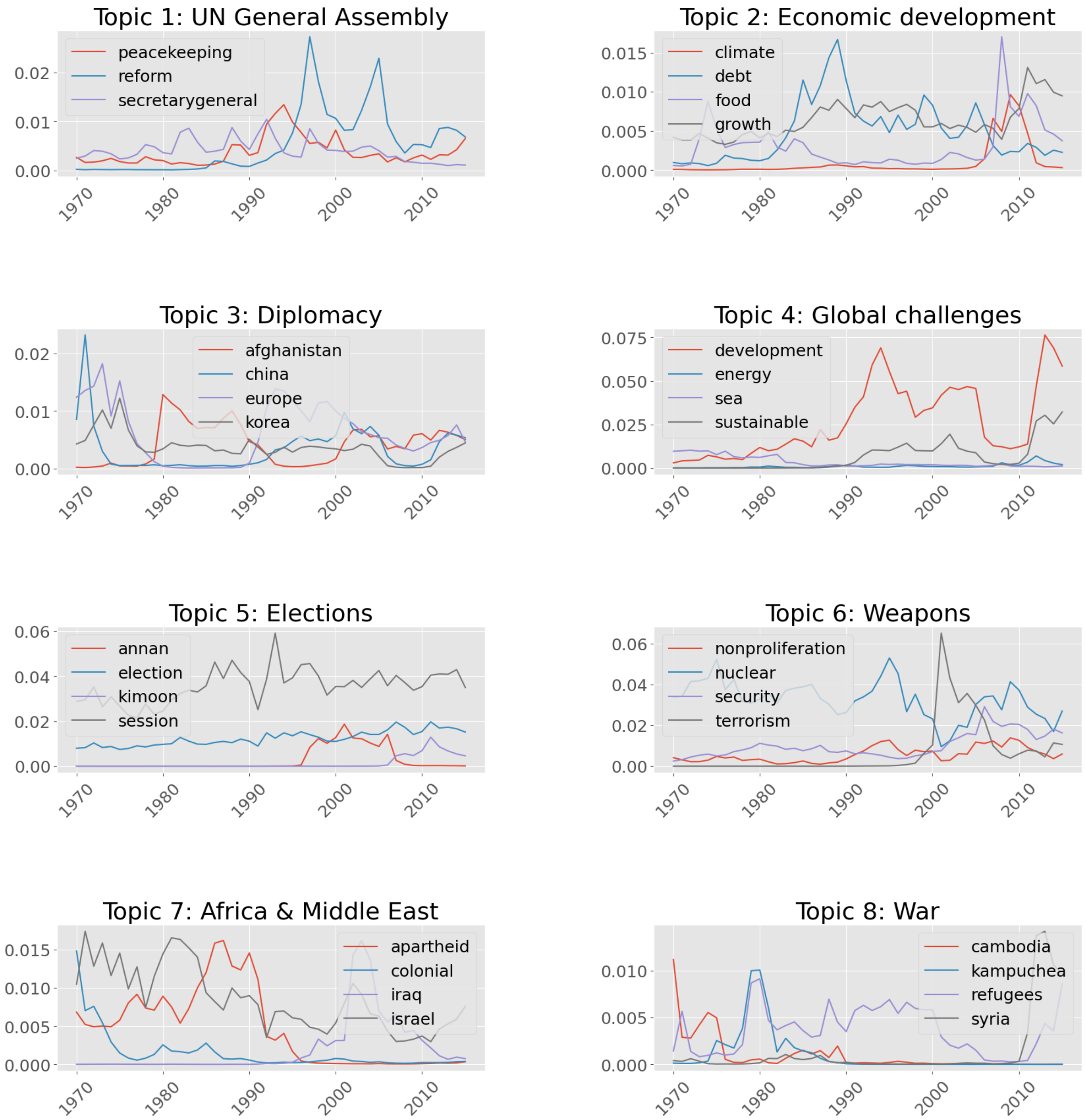

Figure B.2 – Evolution of word probability for eight topics extracted by the D-ETM on the UN dataset

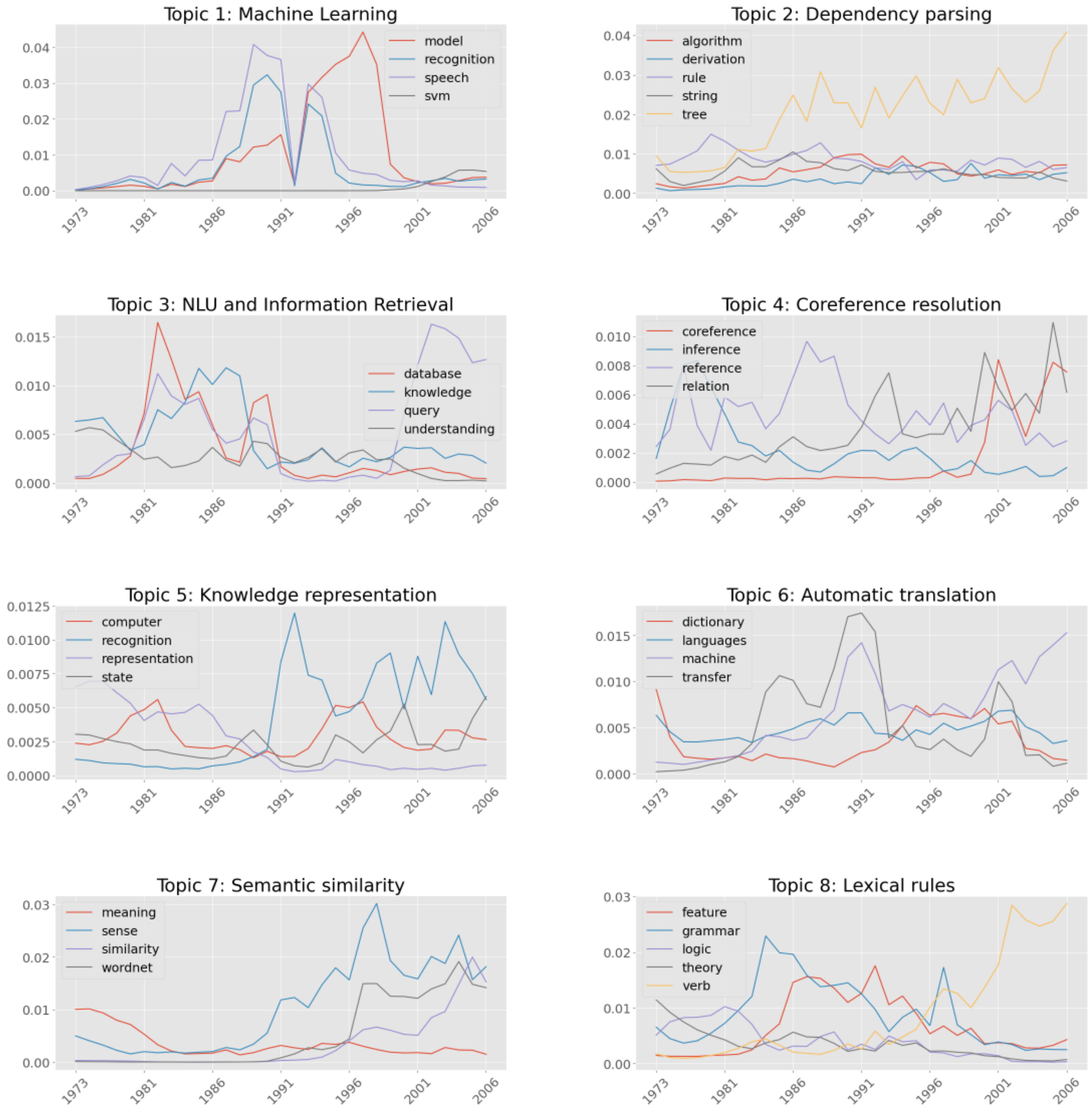

Figure B.3 – Evolution of word probability for eight topics extracted by the D-EDP on the ACL dataset

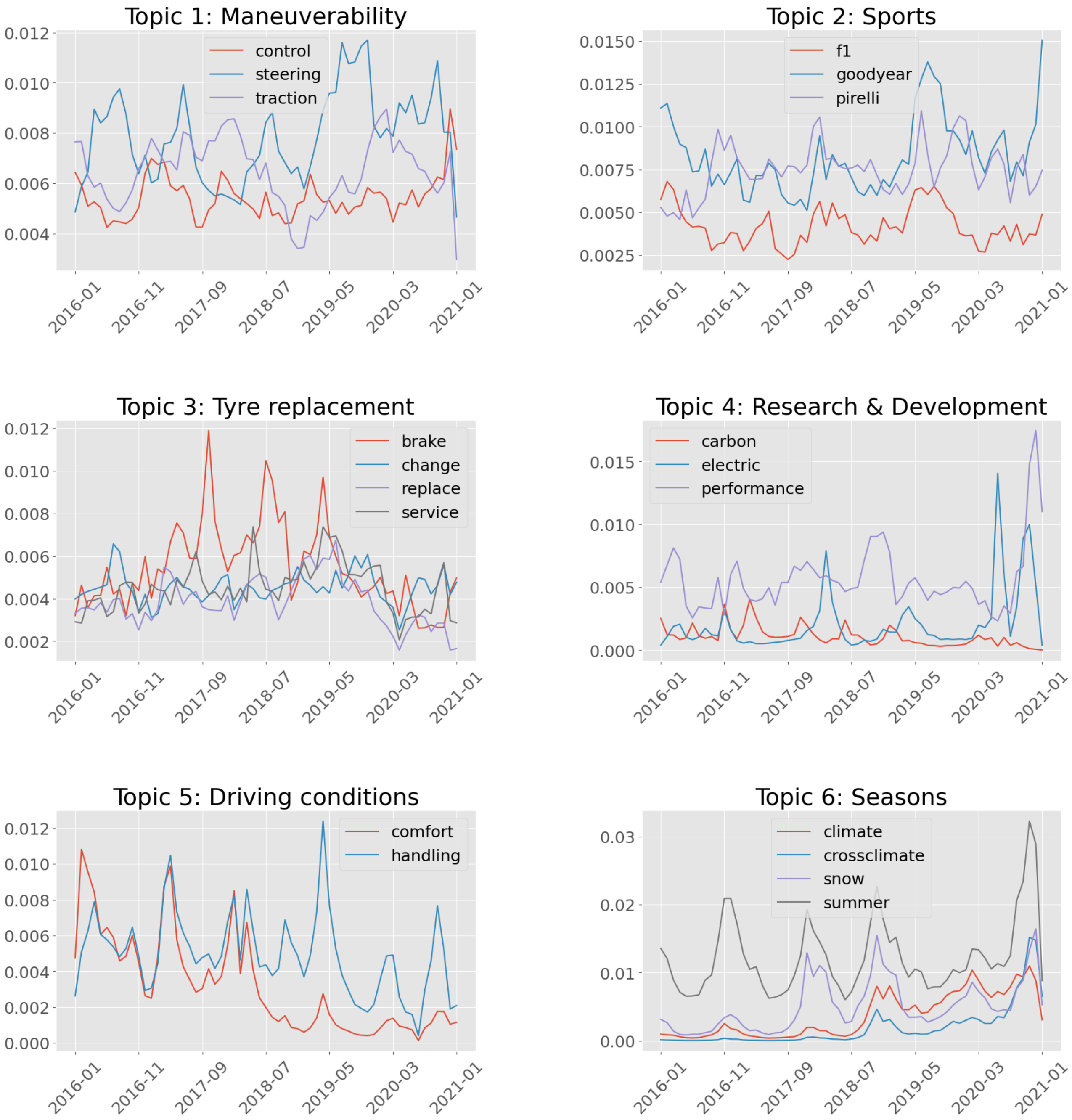

Figure B.4 – Evolution of word probability for six topics extracted by the D-ETM on the EN dataset

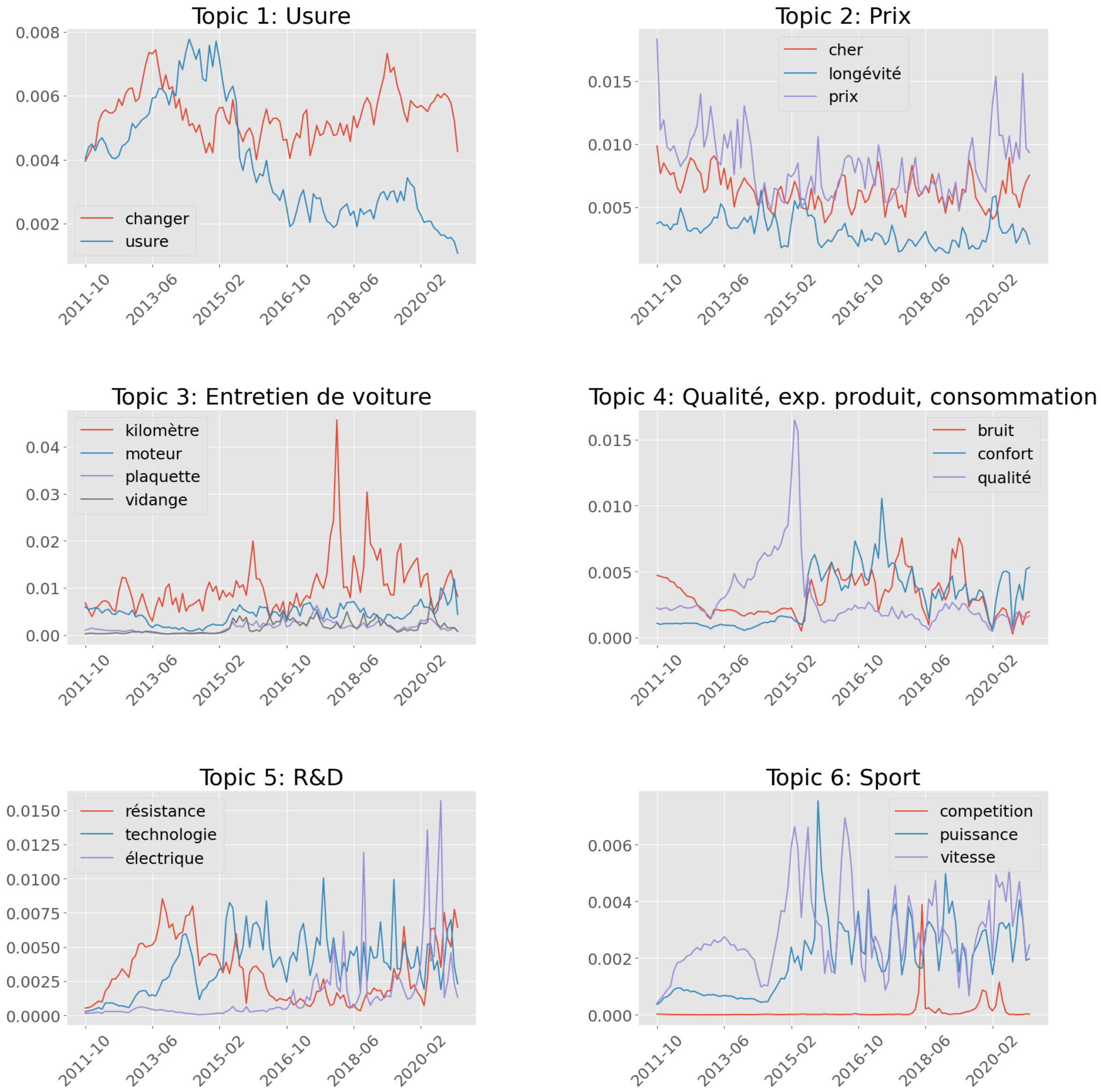

Figure B.5 – Evolution of word probability for six topics extracted by the D-ETM on the FR dataset

| Reference word | Nearest neighbors |
|---|---|
| *economic* | development, social, countries, developing, political, cooperation, system, order, economy, institutions, developed, international, level, resources, relations, based, world, common, progress |
| *assembly* | general, session, delegation, work, nations, united, secretarygeneral, organization, members, member, mr, express, deliberations, important, behalf, success, president, opportunity, extend |
| *security* | peace, council, united, states, force, conflict, military, nations, international, continue, organization, peaceful, time, conflicts, members, end, forces, part, efforts |
| *management* | reform, efficiency, developing, fund, resources, decisionmaking, institutional, growth, efficient, revitalization, financial, developed, structural, reforms, reforming, sustainable, resource, budget, innovative |
| *debt* | income, investment, debts, servicing, earnings, indebtedness, rates, growth, products, markets, rate, commodity, product, capita, creditors, exports, economies, debtservicing, prices |
| *rights* | human, freedoms, rule, fundamental, dignity, constitution, discrimination, inalienable, democracy, violations, freedom, law, selfdetermination, minority, respect, constitutional, protection, society, values |
| *africa* | african, continent, africans, south, southern, assistance, support, namibia, mozambique, zimbabwe, apartheid, liberation, angola, people, colonial, struggle, eradication, continues, continue |

Table B.1 – Word embeddings extracted by the D-ETM on the UN dataset

| Reference word | Nearest neighbors |
|---|---|
| *machine* | translation, english, methods, examples, problems, source, human, method, anguages, important, input, large, process, made, conference, features, similar, make, shown |
| *database* | databases, queries, query, management, developed, developing, applications, area, interface, real, domain, specific, including, provide, addition, issues, technology, users, future |
| *parsing* | parser, parse, parsers, grammar, grammars, parses, np, head, tree, parsed, trees, constituent, vp, structures, left, syntactic, constituents, rules, rule |
| *tree* | trees, node, nodes, grammar, parsing, grammars, left, free, structures, np, parse, called, rules, syntactic, rule, parser, algorithm, hand, context |
| *rule* | rules, left, applied, result, grammar, called, hand, np, cases, section, apply, parsing, simple, fact, context, process, forms, applying, tree |
| *clustering* | clusters, cluster, clustered, cosine, similarity, centroid, agglomerative, distributions, estimates, vectors, wordnet, estimation, vector, smoothing, entropy, estimate, tishby, idf, latent |
| *probability* | probabilities, estimation, estimate, likelihood, estimated, probabilistic, statistical, estimates, trigram, jelinek, entropy, maximum, estimating, stochastic, distribution, distributions, markov, training, models |

Table B.2 – Word embeddings extracted by the D-EDP on the ACL dataset

| Reference word | Nearest neighbors |
|---|---|
| *tire* | product, location, rate, highway, driving, style, city, review, spirited, combine, vehicle, average, mile, condition, purchase, drive, pa, traction, toyota |
| *handling* | handle, noise, traction, dry, review, wet, comfort, continental, resistance, balance, low, distance, point, straight, high, test, surface, road, comfortable |
| *performance* | sport, michelin, high, good, well, wet, grip, set, perform, pirelli, pilot, condition, new, come, summer, car, price, like, winter |
| *noise* | quiet, wear, ride, buy, mile, low, handling, handle, review, noisy, continental, tread, recommend, bad, road, pirelli, long, good, little |
| *braking* | resistance, brake, distance, aquaplane, steering, meter, rolling, lateral, handling, stability, test, dry, negative, th, relatively, control, aquaplaning, overall, ability |
| *michelin* | sport, pilot, good, pirelli, well, set, car, new, continental, come, price, tyre, like, year, long, buy, size, wear, get |
| *electric* | battery, emission, renault, co2, camera, charge, plug, leather, peugeot, powertrain, seat, spoke, instrument, mobility, concept, psa, hurl, combustion, adoption |

Table B.3 – Word embeddings extracted by the D-ETM on the EN dataset

| Reference word | Nearest neighbors |
|---|---|
| *pneus* | monter, user, bonjour, usure, pneu, montage, hiver, salut, neuf, changer, témoin, oui, écrire, monte, cher, perso, dunlop, savoir, ca |
| *crossclimate* | crossclimat, climat, saison, hiver, climate, neige, alpin, weatherproof, vector, cross, pneu, sec, michelin, monte, cc, gen, grip, sculpture, saver |
| *performance* | sport, meilleur, offrir, gamme, nouveau, pilot, également, sportif, test, haut, rapport, modèle, freinage, dimension, marque, version, choix, disponible, michelin |
| *carbone* | fibre, diffuseur, échappement, capot, émission, alcantara, aluminium, aileron, exclusif, système, optimiser, grâce, baquet, couleur, carrosserie, associer, co2, bouclier, innovant |
| *technologie* | développer, matériau, innovant, développement, innovation, technologique, doter, intégrer, mondial, optimiser, marché, composer, croissance, réduire, grâce, offrir, matière, présenter, automobile |
| *michelin* | pneu, qu, marque, mettre, prix, faire, venir, bon, monte, prendre, savoir, route, bien, dernier, continental, meilleur, voiture, hiver, monter |
| *électrique* | batterie, autonomie, recharge, thermique, moteur, kwh, système, rechargeable, pare, recharger, grâce, énergie, concept, automobile, puissance, hybride, dévoiler, présenter, kw |

Table B.4 – Word embeddings extracted by the D-ETM on the FR dataset

# Appendix C

# Software engineering part

This Section presents the implementation of our work. Similarly to statistical work, software engineering requires modeling. These two fields have corresponding frameworks and methodologies that are particularly useful in industrial contexts. Relying on our curated scientific literature and industrial experience, we leveraged state-of-the-art knowledge in both fields to build a toolkit. We first introduce the underlying principles of this part of our research before presenting our technological choices and subsequent decisions.

## C.1 Modeling principles

Our work aims at *industrial* topic extraction for the recall. Consequently, scalability and fast development are critical issues. Achieving these goals requires bridging the gap between business practices in statistical modeling, data mining, and software development. These fields work iteratively in business contexts. Statistical modeling and data mining, on the one hand, essentially iterate through three steps: the modeling phase, the parameter inference step, and the evaluation step [1]. Famous industrial project methodologies include Six Sigma's DMAIC and DMADV [2] [DB04], and CRISP-DM [She00] and its subsequent extension ASUM-DM [Ang+18]. The essential bottleneck is the inference step once a statistical model is defined. The most advanced and complex techniques generally have complex in-

1. G. Box's loop provides a simplified view of these three steps. Whether adopting a frequentist or a bayesian philosophy, these steps hold.
2. These methodologies take direct inspiration from W. E. Deming's Plan-Do-Study-Act Cycle.

ference algorithms due to computational intractability. These algorithms are often iterative and imply deriving update formulas for every parameter. Developing such algorithms requires significant time, and subsequent model modification requires deriving a new inference algorithm. While existing parts of a model - such as probability distributions in a graphical model - are theoretically reusable, recycling these parts in practice is difficult.
On the other hand, reusability is paramount in software engineering, which is involved in model implementation. The object-oriented paradigm gained much popularity in the industry in the 1990s. Its goal is to modularize code blocks, manage them more effectively and enhance reusability. Software project development relies heavily on the ability to build on the existing blocks, especially to comply with business needs. The Agile methodologies (e.g., Scrum and extreme programming) made this reliance central to collaborating with customers and end users, similar to a statistical or data analysis project. Software engineering, however, is not a field that considers data and needs an external driver.
To summarize our point, statistical modeling and software engineering are two distinct fields that share the same iterative approach to increasing performance and meeting industrial and modeling needs. They meet and intertwine in the sense that software implements statistical models are implemented, but it is the modeling that drives the implementation. The bottleneck of inference, however, hinders moving smoothly between the necessary steps to performance improvement.

The Auto-Encoding Variational Bayes (AEVB) framework and its Stochastic Gradient Variational Bayes (SGVB) [KW14] estimator address the issues of parameter inference on large datasets. In their seminal paper, Kingma & Welling introduce the Variational AutoEncoder (VAE) and present it as an application of the AEVB. To illustrate their point, the authors propose using MultiLayer Perceptrons as encoders and decoders with Gaussians and Bernouilli distributions, respectively. These neural networks and distributions are only examples and provide compliance with the AEVB requirements and a good illustration. Still, other neural networks and distributions are usable. The first benefit of using neural networks for parameter inference is that their training does not involve deriving update formulas for every parameter. Instead, training a neural network involves (stochastic) gradient descent and backpropagation. The only requirement besides the modeling step is to define a loss function - the Evidence Lower BOund (ELBO) - that depends on the data and the varia-tional parameters. In other words, a practitioner can add as many parameters as desired as long as it is possible to derive a loss function, thus alleviating the bottleneck of inference. The second benefit of using neural networks is that they allow for including the latest achievements. For instance, we leverage word embeddings in our contributions as we work with textual data, but we could also import knowledge and techniques from other fields, such as Computer Vision. The last benefit is the gain in terms of flexibility. The components of the neural networks are left for the practitioner to choose, thus making it possible to build adequate architectures. This flexibility is also philosophically attractive. When building a statistical, graphical model, one considers distributions as elementary building blocks. Each distribution can capture distinct elements (e.g., sparsity). However, their combination is limited in that some distributions are conjugate while others are not. The combination issue leads to intellectually restricting the modeling to conjugacy settings as it is easier to analyze a posterior. Leaving the data complexity to neural networks [3] enables adding effects in a relatively easy way and using distributions as interpretability proxies. In other words, the VAE framework leverages the benefits of thinking in terms of architecture and interpretability behind the concept of a probability distribution. The VAE for probabilistic graphical modeling framework also has impacts in terms of pure software engineering.

Combining probabilistic graphical modeling with VAEs is an efficient way of iterating through data analytics projects and statistical modeling. On the one hand, probabilistic graphical modeling relies on combining distributions to achieve a simple or hierarchical model and encode uncertainty. On the other hand, in its simplest form, the VAE framework has three distinct components (Fig. 2.3):

1. The encoder captures the data's underlying structure.
2. The "stochastic layer," i.e., the variational distribution and its normalization, allow for interpretability by explicitly modeling the latent variable of interest.
3. The decoder reconstructs the data using a sample from the latent variable.

The VAE's architecture is very modular, thus making it compatible with object-oriented programming (OOP). Additionally, each component is customizable as needed. Consequently, it is possible to create a class - a VAE class - built by composition from three other classes or building blocks: an encoder, a stochastic layer, and a decoder. This modularity requires

3. One could also consider a complex neural block that yields a given effect as an "elementary" component

| Criterion | Statistical modeling and data mining | Software engineering |
|---|---|---|
| *Development cycles* | Iterative | Frequently iterative |
| *Development driver* | Data-driven and business-driven | Business-driven |
| *Building blocks* | Probability distributions | Code |
| *Reusable components* | Distributions, priors, embeddings, and other coefficients | Code blocks (e.g., objects, functions, procedures) |
| *Key performance indicators* | Statistical indicators and business-defined indicators | Runtime, latency, resource usage, business-defined needs fulfillment |
| *Bottlenecks* | Distribution combination and parameter inference, lack of business knowledge | Scalability, code clutter, lack of business knowledge, lack or misuse of resources |

Table C.1 – Project management practice comparison

working on a standard set of interfaces. The ability to replace components modularly is also crucial to avoiding code duplication. Still, it enables the creation of new models by simply coding the missing component(s) or mixing and matching two pre-existent models to create a new one. For instance, one could replace the Bernouilli MLP decoder described in [KW14] with a product of experts [Hin02] or replace the stochastic layer to form a new model (e.g., setting a Beta distribution instead of a Gaussian).

Last but not least, interface and mixins enable extending the classes as needed, depending on the extension. For instance, one could be interested in getting word embeddings in a model that embarks them. While the corresponding functions make sense for these models, they do not serve any purpose for a model without word embeddings. The following section shows how we implement modeling principles in our industrial context.

## C.2 Implementation

At the time our work began, and before starting the software development part, we tried to enumerate the existing solutions , we found no software that strictly complied with our requirements. As we use Deep Learning, we exclude libraries that do not build on Deep Learning frameworks, such as Stan [4] or PyMC3 [5]. The most famous and widely supported are TensorFlow [6] and PyTorch [7]. We preferred TensorFlow (version 2.4.1) over PyTorch for the following reasons:

— TensorFlow has a stable API and an enterprise-friendly ecosystem (e.g., TFX [8]), especially regarding model deployment.

— Similarly to PyTorch, TensorFlow is a high-performance toolkit that allows distributed computing with minimal code base changes [9] and uses various computing units (CPUs, GPUs, or TPUs).

— TensorFlow now includes Keras [10], thus reducing the number of project dependencies. This inclusion also allows a certain flexibility. On the one hand, using Keras enables fast development of neural networks as it is a high level of abstraction to practitioners. Thanks to these properties, they can focus on statistical modeling. On the other hand, developing new, custom components for Keras with TensorFlow is still possible on a lower level. Custom components, or layers, imply that the developer sticks with the Keras API [11] to make them as easy to use. In other words, we can make a toolkit for industrial practice and Research and Development.

Probabilistic modeling had also made its way to TensorFlow by the time we began our research. TensorFlow developers, indeed, have created a dedicated extension known as TensorFlow Probability [12] (TFP). This extension contains classes that stand for probability distributions. These classes implement a variety of methods that enable computing of fa-

4. https://mc-stan.org
5. https://www.pymc.io/welcome.html
6. https://www.tensorflow.org
7. https://pytorch.org
8. https://www.tensorflow.org/tfx
9. https://www.tensorflow.org/guide/distributed_training
10. https://keras.io/
11. https://keras.io/api/layers/base_layer/
12. https://www.tensorflow.org/probability

miliar KL-Divergences [13] (KLD) without having to code the whole formula, adding custom KL-Divergences as well (e.g., a KLD between a Beta and a Kumaraswamy distribution), and sampling. TFP is deeply rooted in scientific research. Consequently, it implements many techniques, including implicit reparameterization gradients [FMM18]. Last but not least, and similarly to TensorFlow, TFP is still under active development, thus making its deprecation unlikely in the following years.

**Disclaimer** Due to contractual restrictions, we cannot disclose all of the toolkit's blueprint or code and must restrain our presentation to the essentials. We refer the reader to Tensorflow's documentation for more details on its contents [14]. The following page shows a *voluntarily rudimentary* UML representation of our toolkit.

13. `https://www.tensorflow.org/probability/api_docs/python/tfp/distributions/kl_divergence`
14. `https://www.tensorflow.org/versions/r2.4/api_docs/python/tf`

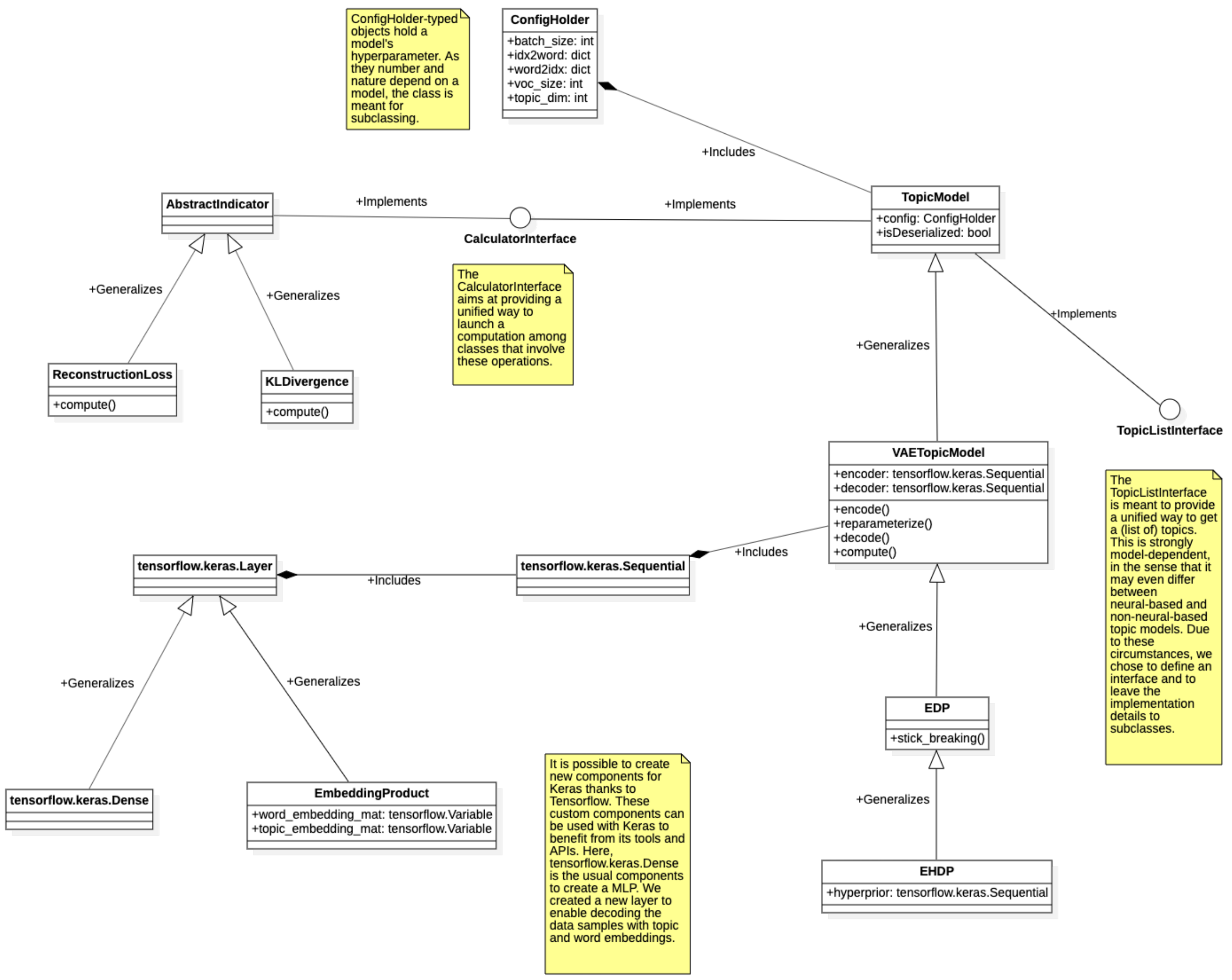
ConfigHolder-typed objects hold a model's hyperparameter. As they number and nature depend on a model, the class is meant for subclassing.
ConfigHolder
+batch_size: int
+idx2word: dict
+word2idx: dict
+voc_size: int
+topic_dim: int
+Includes
AbstractIndicator
+Implements
CalculatorInterface
+Implements
TopicModel
+config: ConfigHolder
+isDeserialized: bool
+Generalizes
+Generalizes
The CalculatorInterface aims at providing a unified way to launch a computation among classes that involve these operations.
+Implements
+Generalizes
ReconstructionLoss
+compute()
KLDivergence
+compute()
TopicListInterface
VAETopicModel
+encoder: tensorflow.keras.Sequential
+decoder: tensorflow.keras.Sequential
+encode()
+reparameterize()
+decode()
+compute()
The TopicListInterface is meant to provide a unified way to get a (list of) topics. This is strongly model-dependent, in the sense that it may even differ between neural-based and non-neural-based topic models. Due to these circumstances, we chose to define an interface and to leave the implementation details to subclasses.
tensorflow.keras.Layer
+Includes
tensorflow.keras.Sequential
+Includes
+Generalizes
+Generalizes
+Generalizes
EDP
+stick_breaking()
tensorflow.keras.Dense
EmbeddingProduct
+word_embedding_mat: tensorflow.Variable
+topic_embedding_mat: tensorflow.Variable
It is possible to create new components for Keras thanks to Tensorflow. These custom components can be used with Keras to benefit from its tools and APIs. Here, tensorflow.keras.Dense is the usual components to create a MLP. We created a new layer to enable decoding the data samples with topic and word embeddings.
+Generalizes
EHDP
+hyperprior: tensorflow.keras.Sequential

# Appendix D

# Publication list

All the conferences listed below imply a peer-reviewing process.

## D.1 International conferences

**Palencia-Olivar, M.**, Bonnevay, S., Aussem, A., Canitia, B. (2022). Nonparametric neural topic modeling for customer insight extraction about the tire industry. International Joint Conference on Neural Networks (IJCNN), 2022, pp. 01-09, doi: 10.1109/IJCNN55064.2022.9892577.

**Palencia-Olivar, M.** (2022). A Topical Approach to Capturing Customer Insight Dynamics in Social Media. In: , et al. Advances in Information Retrieval. ECIR 2022. Lecture Notes in Computer Science, vol 13186. Springer, Cham. https://doi.org/10.1007/978-3-030-99739-7_64.

**Palencia-Olivar, M.**, Bonnevay, S., Aussem, A., Canitia, B. (2021). Neural Embedded Dirichlet Processes for Topic Modeling. In: Torra, V., Narukawa, Y. (eds) Modeling Decisions for Artificial Intelligence. MDAI 2021. Lecture Notes in Computer Science, vol 12898. Springer, Cham. https://doi.org/10.1007/978-3-030-85529-1_24.

## D.2 Local conferences

**Palencia-Olivar, M.**, Bonnevay, S., Aussem, A., Canitia, B. (2023). Topic modeling neuronal non-paramétrique pour l'extraction d'insight client : une application à l'industrie du pneumatique. Conférence francophone sur l'Extraction et la Gestion des Connaissances (EGC), Lyon (France), January, 2023 (to appear).

**Palencia-Olivar, M.**, Bonnevay, S., Aussem, A., Canitia, B. (2021). Processus de Dirichlet profonds pour le topic modeling. Conférence francophone sur l'Extraction et la Gestion des Connaissances (EGC), vol. RNTI-E-38, pages 355-362, Blois (France), January, 2022.

# References

[Agr19] Alan Agresti. *An Introduction to Categorical Data Analysis*. July 2019. ISBN: 978-1-119-62342-7.

[AX10] Amr Ahmed and Eric P. Xing. "Timeline: A Dynamic Hierarchical Dirichlet Process Model for Recovering Birth/Death and Evolution of Topics in Text Stream". In: *Proceedings of the Twenty-Sixth Conference on Uncertainty in Artificial Intelligence*. UAI'10. Catalina Island, CA: AUAI Press, 2010, pp. 20–29.

[AC01] Jim Albert and Siddhartha Chib. "Sequential Ordinal Modeling with Applications to Survival Data". In: *Biometrics* 57 (Oct. 2001), pp. 829–36. DOI: 10.1111/j.0006-341X.2001.00829.x.

[AS13] Nikolaos Aletras and Mark Stevenson. "Evaluating Topic Coherence Using Distributional Semantics". In: *Proceedings of the 10th International Conference on Computational Semantics (IWCS 2013) – Long Papers*. Potsdam, Germany: Association for Computational Linguistics, Mar. 2013, pp. 13–22. URL: https://aclanthology.org/W13-0102.

[Ang+18] Santiago Angée et al. "Towards an Improved ASUM-DM Process Methodology for Cross-Disciplinary Multi-organization Big Data Analytics Projects: 13th International Conference, KMO 2018, Žilina, Slovakia, August 6–10, 2018, Proceedings". In: July 2018, pp. 613–624. ISBN: 978-3-319-95203-1. DOI: 10.1007/978-3-319-95204-8_51.

[Arm+21] Maxime D. Armstrong et al. "Topic Modeling in Embedding Spaces for Depression Assessment". en. In: *Proceedings of the Canadian Conference on Artificial*

*Intelligence* (June 2021). DOI: 10.21428/594757db.9e67a9f0. (Visited on 12/01/2021).

[Bal+16] Georgios Balikas et al. "Modeling topic dependencies in semantically coherent text spans with copulas". In: *Proceedings of COLING 2016, the 26th International Conference on Computational Linguistics: Technical Papers*. Osaka, Japan: The COLING 2016 Organizing Committee, Dec. 2016, pp. 1767–1776. URL: https://aclanthology.org/C16-1166.

[Bat+16] Kayhan Batmanghelich et al. "Nonparametric Spherical Topic Modeling with Word Embeddings". In: *Proceedings of the 54th Annual Meeting of the Association for Computational Linguistics (Volume 2: Short Papers)*. Berlin, Germany: Association for Computational Linguistics, Aug. 2016, pp. 537–542. DOI: 10.18653/v1/P16-2087. URL: https://aclanthology.org/P16-2087.

[BDM17] Alexander Baturo, Niheer Dasandi, and Slava J. Mikhaylov. "Understanding state preferences with text as data: Introducing the UN General Debate corpus". In: *Research & Politics* 4.2 (2017), p. 2053168017712821. URL: https://doi.org/10.1177/2053168017712821.

[BMT21] Andrew Bennett, Dipendra Misra, and Nga Than. "Have you tried Neural Topic Models? Comparative Analysis of Neural and Non-Neural Topic Models with Application to COVID-19 Twitter Data". In: *CoRR* abs/2105.10165 (2021). arXiv: 2105.10165. URL: https://arxiv.org/abs/2105.10165.

[BTH21] Federico Bianchi, Silvia Terragni, and Dirk Hovy. "Pre-training is a Hot Topic: Contextualized Document Embeddings Improve Topic Coherence". In: *Proceedings of the 59th Annual Meeting of the Association for Computational Linguistics and the 11th International Joint Conference on Natural Language Processing (Volume 2: Short Papers)*. Online: Association for Computational Linguistics, Aug. 2021, pp. 759–766. DOI: 10.18653/v1/2021.acl-short.96. URL: https://aclanthology.org/2021.acl-short.96.

[Bia+21] Federico Bianchi et al. "Cross-lingual Contextualized Topic Models with Zero-shot Learning". In: *Proceedings of the 16th Conference of the European Chapter of the Association for Computational Linguistics: Main Volume*. Online: Association for Computational Linguistics, Apr. 2021, pp. 1676–1683. DOI: 10.18653/

v1/2021.eacl-main.143. URL: https://aclanthology.org/2021.eacl-main.143.

[Bir+08] Steven Bird et al. “The ACL Anthology Reference Corpus: A Reference Dataset for Bibliographic Research in Computational Linguistics”. In: *Proceedings of the Sixth International Conference on Language Resources and Evaluation (LREC'08).* Marrakech, Morocco: European Language Resources Association (ELRA), May 2008. URL: http://www.lrec-conf.org/proceedings/lrec2008/pdf/445_paper.pdf.

[BKM16] David Blei, Alp Kucukelbir, and Jon McAuliffe. “Variational Inference: A Review for Statisticians”. In: *Journal of the American Statistical Association* 112 (Jan. 2016). DOI: 10.1080/01621459.2017.1285773.

[BL06] David M. Blei and John D. Lafferty. “Dynamic Topic Models”. In: *Proceedings of the 23rd International Conference on Machine Learning.* ICML '06. Pittsburgh, Pennsylvania, USA: Association for Computing Machinery, 2006, pp. 113–120. ISBN: 1595933832. DOI: 10.1145/1143844.1143859. URL: https://doi.org/10.1145/1143844.1143859.

[BNJ03] David M. Blei, Andrew Y. Ng, and Michael I. Jordan. “Latent dirichlet allocation”. In: *J. Mach. Learn. Res.* 3 (2003), pp. 993–1022. ISSN: 1532-4435. DOI: http://dx.doi.org/10.1162/jmlr.2003.3.4-5.993. URL: http://portal.acm.org/citation.cfm?id=944937.

[Ble+03] David M. Blei et al. “Hierarchical Topic Models and the Nested Chinese Restaurant Process”. In: *Proceedings of the 16th International Conference on Neural Information Processing Systems.* NIPS'03. Whistler, British Columbia, Canada: MIT Press, 2003, pp. 17–24.

[Bow+16] Samuel R. Bowman et al. “Generating Sentences from a Continuous Space”. In: *Proceedings of the 20th SIGNLL Conference on Computational Natural Language Learning.* Berlin, Germany: Association for Computational Linguistics, Aug. 2016, pp. 10–21. DOI: 10.18653/v1/K16-1002. URL: https://aclanthology.org/K16-1002.

[Box76] George E. P. Box. "Science and Statistics". In: *Journal of the American Statistical Association* 71.356 (1976), pp. 791–799. DOI: 10.1080/01621459.1976.10480949. eprint: https://www.tandfonline.com/doi/pdf/10.1080/01621459.1976.10480949. URL: https://www.tandfonline.com/doi/abs/10.1080/01621459.1976.10480949.

[BHM17] Jordan Boyd-Graber, Yuening Hu, and David Mimno. *Applications of Topic Models*. Jan. 2017. ISBN: 9781680833096. DOI: 10.1561/9781680833096.

[Cha+09] Jonathan Chang et al. "Reading Tea Leaves: How Humans Interpret Topic Models". In: *Advances in Neural Information Processing Systems*. Ed. by Y. Bengio et al. Vol. 22. Curran Associates, Inc., 2009. URL: https://proceedings.neurips.cc/paper/2009/file/f92586a25bb3145facd64ab20fd554ff-Paper.pdf.

[Che+17] Xi Chen et al. "Variational Lossy Autoencoder". In: *5th International Conference on Learning Representations, ICLR 2017, Toulon, France, April 24-26, 2017, Conference Track Proceedings*. OpenReview.net, 2017. URL: https://openreview.net/forum?id=BysvGP5ee.

[Con72] W. J. Conover. "A Kolmogorov Goodness-of-Fit Test for Discontinuous Distributions". In: *Journal of the American Statistical Association* 67.339 (1972), pp. 591–596. DOI: 10.1080/01621459.1972.10481254. eprint: https://www.tandfonline.com/doi/pdf/10.1080/01621459.1972.10481254. URL: https://www.tandfonline.com/doi/abs/10.1080/01621459.1972.10481254.

[DZD15] Rajarshi Das, Manzil Zaheer, and Chris Dyer. "Gaussian LDA for Topic Models with Word Embeddings". In: *Proceedings of the 53rd Annual Meeting of the Association for Computational Linguistics and the 7th International Joint Conference on Natural Language Processing (Volume 1: Long Papers)*. Beijing, China: Association for Computational Linguistics, July 2015, pp. 795–804. DOI: 10.3115/v1/P15-1077. URL: https://aclanthology.org/P15-1077.

[DB04] Joseph A. De Feo and William Barnard. *Juran Institute's six sigma : breakthrough and beyond : quality performance breakthrough methods*. McGraw-Hill, 2004. ISBN: 9780071422277.

[Dee+90] Scott Deerwester et al. "Indexing by latent semantic analysis". In: *Journal of the American Society for Information Science* 41.6 (1990), pp. 391–407. URL: shorturl.at/CFSTY.

[Dev86] Luc Devroye. "Sample-Based Non-Uniform Random Variate Generation". In: *Proceedings of the 18th Conference on Winter Simulation*. WSC '86. Washington, D.C., USA: Association for Computing Machinery, 1986, pp. 260–265. DOI: 10.1145/318242.318443. URL: https://doi.org/10.1145/318242.318443.

[Die21] Adji B. Dieng. *Deep Probabilistic Graphical Modeling*. 2021. DOI: 10.48550/ARXIV.2104.12053. URL: https://arxiv.org/abs/2104.12053.

[DRB19] Adji B. Dieng, Francisco J. R. Ruiz, and David M. Blei. *The Dynamic Embedded Topic Model*. 2019. DOI: 10.48550/ARXIV.1907.05545. URL: https://arxiv.org/abs/1907.05545.

[DRB20] Adji B. Dieng, Francisco J. R. Ruiz, and David M. Blei. "Topic Modeling in Embedding Spaces". In: *Transactions of the Association for Computational Linguistics* 8 (2020), pp. 439–453. DOI: 10.1162/tacl_a_00325. URL: https://aclanthology.org/2020.tacl-1.29.

[Die+17] Adji B. Dieng et al. "TopicRNN: A Recurrent Neural Network with Long-Range Semantic Dependency". In: *International Conference on Learning Representations*. 2017. URL: https://openreview.net/forum?id=rJbbOLcex.

[DBJ10] Lan Du, Wray Lindsay Buntine, and Huidong Jin. "Sequential Latent Dirichlet Allocation: Discover Underlying Topic Structures within a Document". In: *2010 IEEE International Conference on Data Mining*. 2010, pp. 148–157. DOI: 10.1109/ICDM.2010.51.

[Dub+12] Avinava Dubey et al. *A non-parametric mixture model for topic modeling over time*. 2012. DOI: 10.48550/ARXIV.1208.4411. URL: https://arxiv.org/abs/1208.4411.

[DB16] Christophe Dupuy and Francis R. Bach. "Online but Accurate Inference for Latent Variable Models with Local Gibbs Sampling". In: *CoRR* abs/1603.02644 (2016). URL: http://arxiv.org/abs/1603.02644.

[El +21] Manal El Akrouchi et al. “End-to-end LDA-based automatic weak signal detection in web news”. en. In: *Knowledge-Based Systems* 212 (Jan. 2021), p. 106650. ISSN: 09507051. DOI: 10.1016/j.knosys.2020.106650. (Visited on 12/01/2021).

[EW95] Michael D. Escobar and Mike West. “Bayesian Density Estimation and Inference Using Mixtures”. In: *Journal of the American Statistical Association* 90.430 (1995), pp. 577–588. DOI: 10.1080/01621459.1995.10476550.

[FMM18] Michael Figurnov, Shakir Mohamed, and Andriy Mnih. “Implicit Reparameterization Gradients”. In: *Proceedings of the 32nd International Conference on Neural Information Processing Systems*. NIPS’18. Montréal, Canada: Curran Associates Inc., 2018, pp. 439–450.

[GS21] S. N. Geethalakshmi and S. Shaambavi. “Opinion Mining With Hotel Review using Latent Dirichlet Allocation-Fuzzy C-Means Clustering (LDA-FCM)”. en. In: *Turkish Journal of Computer and Mathematics Education Vol.12 No. 11 (2021), 3001- 3007 Research Article* 11 (2021), p. 7.

[GG14] Samuel J. Gershman and Noah D. Goodman. “Amortized Inference in Probabilistic Reasoning”. In: *Cognitive Science* 36 (2014).

[Goy+17] Prasoon Goyal et al. *Nonparametric Variational Auto-encoders for Hierarchical Representation Learning*. 2017. DOI: 10.48550/ARXIV.1703.07027. URL: https://arxiv.org/abs/1703.07027.

[Gri+04] Thomas Griffiths et al. “Integrating Topics and Syntax”. In: *Advances in Neural Information Processing Systems*. Ed. by L. Saul, Y. Weiss, and L. Bottou. Vol. 17. MIT Press, 2004. URL: https://proceedings.neurips.cc/paper/2004/file/ef0917ea498b1665ad6c701057155abe-Paper.pdf.

[HP21] Stefan Helmstetter and Heiko Paulheim. “Collecting a Large Scale Dataset for Classifying Fake News Tweets Using Weak Supervision”. en. In: *Future Internet* 13.5 (Apr. 2021), p. 114. ISSN: 1999-5903. DOI: 10.3390/fi13050114. (Visited on 12/01/2021).

[HS06] G. E. Hinton and R. R. Salakhutdinov. "Reducing the Dimensionality of Data with Neural Networks". In: *Science* 313.5786 (2006), pp. 504–507. DOI: 10.1126/science.1127647. URL: https://www.science.org/doi/abs/10.1126/science.1127647.

[Hin02] Geoffrey E. Hinton. "Training Products of Experts by Minimizing Contrastive Divergence". In: *Neural Comput.* 14.8 (Aug. 2002), pp. 1771–1800. ISSN: 0899-7667. DOI: 10.1162/089976602760128018. URL: https://doi.org/10.1162/089976602760128018.

[HBB10] Matthew D. Hoffman, David M. Blei, and Francis Bach. "Online Learning for Latent Dirichlet Allocation". In: *Proceedings of the 23rd International Conference on Neural Information Processing Systems - Volume 1.* NIPS'10. Vancouver, British Columbia, Canada: Curran Associates Inc., 2010, pp. 856–864.

[Hof+13] Matthew D. Hoffman et al. "Stochastic Variational Inference". In: *Journal of Machine Learning Research* 14.40 (2013), pp. 1303–1347. URL: http://jmlr.org/papers/v14/hoffman13a.html.

[Hof99] Thomas Hofmann. "Probabilistic Latent Semantic Indexing". In: *Proceedings of the 22nd Annual International ACM SIGIR Conference on Research and Development in Information Retrieval.* SIGIR '99. Berkeley, California, USA: Association for Computing Machinery, 1999, pp. 50–57. ISBN: 1581130961. DOI: 10.1145/312624.312649. URL: https://doi.org/10.1145/312624.312649.

[Hon+19] Matthew Honnibal et al. *explosion/spaCy: v2.1.7: Improved evaluation, better language factories and bug fixes.* Version v2.1.7. Aug. 2019. DOI: 10.5281/zenodo.3358113. URL: https://doi.org/10.5281/zenodo.3358113.

[HSW89] Kurt Hornik, Maxwell Stinchcombe, and Halbert White. "Multilayer feedforward networks are universal approximators". In: *Neural Networks* 2.5 (1989), pp. 359–366. ISSN: 0893-6080. DOI: https://doi.org/10.1016/0893-6080(89)90020-8. URL: shorturl.at/hPRU2.

[Hor17] Ghraham V. Horwood. "reliefweb_corpus_raw_20160331.json". In: *Humanitarian Assistance and Disaster Relief (HA/DR) Articles and Lexicon.* Version Number: V1. Harvard Dataverse, 2017.

[Hoy+21] Alexander Hoyle et al. "Is Automated Topic Model Evaluation Broken? The Incoherence of Coherence". In: *Advances in Neural Information Processing Systems*. Ed. by M. Ranzato et al. Vol. 34. Curran Associates, Inc., 2021, pp. 2018–2033. URL: `shorturl.at/cnsS3`.

[HBS11] Yuening Hu, Jordan Boyd-Graber, and Brianna Satinoff. "Interactive Topic Modeling". In: *Proceedings of the 49th Annual Meeting of the Association for Computational Linguistics: Human Language Technologies*. Portland, Oregon, USA: Association for Computational Linguistics, June 2011, pp. 248–257. URL: `https://aclanthology.org/P11-1026`.

[IJ01] Hemant Ishwaran and Lancelot F James. "Gibbs Sampling Methods for Stick-Breaking Priors". In: *Journal of the American Statistical Association* 96.453 (2001), pp. 161–173. URL: `https://doi.org/10.1198/016214501750332758`.

[Jel+19] Hamed Jelodar et al. "Latent Dirichlet Allocation (LDA) and Topic Modeling: Models, Applications, a Survey". In: *Multimedia Tools Appl.* 78.11 (June 2019), pp. 15169–15211. ISSN: 1380-7501. DOI: `10.1007/s11042-018-6894-4`. URL: `https://doi.org/10.1007/s11042-018-6894-4`.

[KW14] Diederik P. Kingma and Max Welling. "Auto-Encoding Variational Bayes". In: *2nd International Conference on Learning Representations, ICLR 2014, Banff, AB, Canada, April 14-16, 2014, Conference Track Proceedings*. Ed. by Yoshua Bengio and Yann LeCun. 2014. URL: `http://arxiv.org/abs/1312.6114`.

[Kin+16] Durk P Kingma et al. "Improved Variational Inference with Inverse Autoregressive Flow". In: *Advances in Neural Information Processing Systems*. Ed. by D. Lee et al. Vol. 29. Curran Associates, Inc., 2016. URL: `https://proceedings.neurips.cc/paper/2016/file/ddeebdeefdb7e7e7a697e1c3e3d8ef54-Paper.pdf`.

[Kum80] P. Kumaraswamy. "A generalized probability density function for double-bounded random processes". In: *Journal of Hydrology* 46.1 (1980), pp. 79–88. ISSN: 0022-1694. DOI: `https://doi.org/10.1016/0022-1694(80)90036-0`. URL: `https://www.sciencedirect.com/science/article/pii/0022169480900360`.

[LB05] John Lafferty and David Blei. "Correlated Topic Models". In: *Advances in Neural Information Processing Systems*. Ed. by Y. Weiss, B. Schölkopf, and J. Platt. Vol. 18. MIT Press, 2005. URL: https://proceedings.neurips.cc/paper/2005/file/9e82757e9a1c12cb710ad680db11f6f1-Paper.pdf.

[LNB14] Jey Han Lau, David Newman, and Timothy Baldwin. "Machine Reading Tea Leaves: Automatically Evaluating Topic Coherence and Topic Model Quality". In: *Proceedings of the 14th Conference of the European Chapter of the Association for Computational Linguistics*. Gothenburg, Sweden: Association for Computational Linguistics, Apr. 2014, pp. 530–539. DOI: 10.3115/v1/E14-1056. URL: https://aclanthology.org/E14-1056.

[Li+20] Linhong Li et al. "Neural Topic Models with Survival Supervision: Jointly Predicting Time-to-Event Outcomes and Learning How Clinical Features Relate". In: *CoRR* abs/2007.07796 (2020). arXiv: 2007.07796. URL: https://arxiv.org/abs/2007.07796.

[Mau+21] Diego Maupomé et al. "Early Detection of Signs of Pathological Gambling, Self-Harm and Depression through Topic Extraction and Neural Networks". In: *Proceedings of the Working Notes of CLEF 2021 - Conference and Labs of the Evaluation Forum, Bucharest, Romania, September 21st - to - 24th, 2021*. Ed. by Guglielmo Faggioli et al. Vol. 2936. CEUR Workshop Proceedings. CEUR-WS.org, 2021, pp. 1031–1045.

[MBJ06] Jon D. McAuliffe, David M. Blei, and Michael I. Jordan. "Nonparametric Empirical Bayes for the Dirichlet Process Mixture Model". In: *Statistics and Computing* 16.1 (Mar. 2006), pp. 5–14. ISSN: 0960-3174. DOI: 10.1007/s11222-006-5196-2.

[MYB16] Yishu Miao, Lei Yu, and Phil Blunsom. "Neural Variational Inference for Text Processing". In: *Proceedings of The 33rd International Conference on Machine Learning*. Ed. by Maria Florina Balcan and Kilian Q. Weinberger. Vol. 48. Proceedings of Machine Learning Research. New York, New York, USA: PMLR, 20–22 Jun 2016, pp. 1727–1736. URL: https://proceedings.mlr.press/v48/miao16.html.

[Mik+13] Tomas Mikolov et al. *Efficient Estimation of Word Representations in Vector Space*. 2013. DOI: 10.48550/ARXIV.1301.3781. URL: https://arxiv.org/abs/1301.3781.

[Mim+11] David Mimno et al. "Optimizing Semantic Coherence in Topic Models". In: *Proceedings of the Conference on Empirical Methods in Natural Language Processing*. EMNLP '11. Edinburgh, United Kingdom: Association for Computational Linguistics, 2011, pp. 262–272. ISBN: 9781937284114.

[NS17] Eric Nalisnick and Padhraic Smyth. "Stick-Breaking Variational Autoencoders". In: *International Conference on Learning Representations*. 2017. URL: https://openreview.net/forum?id=S1jmAotxg.

[New+10] David Newman et al. "Automatic Evaluation of Topic Coherence." In: Jan. 2010, pp. 100–108.

[Nin+20] Xuefei Ning et al. "Nonparametric Topic Modeling with Neural Inference". In: *Neurocomputing* 399 (2020), pp. 296–306. ISSN: 0925-2312. DOI: https://doi.org/10.1016/j.neucom.2019.12.128. URL: https://www.sciencedirect.com/science/article/pii/S0925231220303015.

[Qia+22] Jipeng Qiang et al. "Short Text Topic Modeling Techniques, Applications, and Performance: A Survey". In: *IEEE Transactions on Knowledge and Data Engineering* 34.3 (2022), pp. 1427–1445. DOI: 10.1109/TKDE.2020.2992485.

[RDC08] Lu Ren, David B. Dunson, and Lawrence Carin. "The Dynamic Hierarchical Dirichlet Process". In: *Proceedings of the 25th International Conference on Machine Learning*. ICML '08. Helsinki, Finland: Association for Computing Machinery, 2008, pp. 824–831. ISBN: 9781605582054. DOI: 10.1145/1390156.1390260. URL: https://doi.org/10.1145/1390156.1390260.

[Ren+11] Lu Ren et al. "Logistic Stick-Breaking Process". In: *J. Mach. Learn. Res.* 12 (Feb. 2011), pp. 203–239. ISSN: 1532-4435.

[RD11] Abel Rodriguez and David Dunson. "Nonparametric Bayesian models through probit stick-breaking processes". In: *Bayesian Analysis* 6 (Mar. 2011), pp. 145–178. DOI: 10.1214/11-BA605.

[RTB16] Francisco J. R. Ruiz, Michalis K. Titsias, and David M. Blei. "The Generalized Reparameterization Gradient". In: *Proceedings of the 30th International Conference on Neural Information Processing Systems*. NIPS'16. Barcelona, Spain: Curran Associates Inc., 2016, pp. 460–468. ISBN: 9781510838819.

[RHW86] David E. Rumelhart, Geoffrey E. Hinton, and Ronald J. Williams. "Learning representations by back-propagating errors". In: *Nature* 323 (1986), pp. 533–536.

[Set94] Jayaram Sethuraman. "A Constructive Definition Of Dirichlet Priors". In: *Statistica Sinica* 4.2 (1994), pp. 639–650. ISSN: 10170405, 19968507.

[She00] Colin Shearer. "The CRISP-DM Model: The New Blueprint for Data Mining". In: *Journal of Data Warehousing* 5 (2000), pp. 13–22.

[Søn+17] Casper Kaae Sønderby et al. "Amortised MAP Inference for Image Super-resolution". In: *5th International Conference on Learning Representations, ICLR 2017, Toulon, France, April 24-26, 2017, Conference Track Proceedings*. OpenReview.net, 2017. URL: `https://openreview.net/forum?id=S1RP6GLle`.

[SS17] Akash Srivastava and Charles Sutton. "Autoencoding Variational Inference For Topic Models". In: *International Conference on Learning Representations*. 2017. URL: `https://openreview.net/forum?id=BybtVK9lg`.

[Teh+06] Yee Whye Teh et al. "Hierarchical Dirichlet Processes". In: *Journal of the American Statistical Association* 101.476 (2006), pp. 1566–1581. URL: `https://doi.org/10.1198/016214506000000302`.

[VJ21] Vidya Vukanti and Anu Jose. "Business Analytics: A case-study approach using LDA topic modelling". In: *2021 5th International Conference on Computing Methodologies and Communication (ICCMC)*. 2021, pp. 1818–1823. DOI: `10.1109/ICCMC51019.2021.9418344`.

[Vul+15] Ivan Vulić et al. "Probabilistic topic modeling in multilingual settings: An overview of its methodology and applications". In: *Information Processing Management* 51.1 (2015), pp. 111–147. ISSN: 0306-4573.

[Wal+09] Hanna M. Wallach et al. “Evaluation Methods for Topic Models”. In: *Proceedings of the 26th Annual International Conference on Machine Learning*. ICML ’09. Montreal, Quebec, Canada: Association for Computing Machinery, 2009, pp. 1105–1112. ISBN: 9781605585161. DOI: 10.1145/1553374.1553515. URL: https://doi.org/10.1145/1553374.1553515.

[WB09] Chong Wang and David Blei. “Variational Inference for the Nested Chinese Restaurant Process”. In: *Advances in Neural Information Processing Systems*. Ed. by Y. Bengio et al. Vol. 22. Curran Associates, Inc., 2009. URL: https://proceedings.neurips.cc/paper/2009/file/ca46c1b9512a7a8315fa3c5a946e8265-Paper.pdf.

[WBH08] Chong Wang, David Blei, and David Heckerman. “Continuous Time Dynamic Topic Models”. In: *Proceedings of the Twenty-Fourth Conference on Uncertainty in Artificial Intelligence*. UAI’08. Helsinki, Finland: AUAI Press, 2008, pp. 579–586. ISBN: 0974903949.

[WB13] Chong Wang and David M. Blei. “Variational Inference in Nonconjugate Models”. In: *J. Mach. Learn. Res.* 14.1 (Apr. 2013), pp. 1005–1031. ISSN: 1532-4435.

[WPB11a] Chong Wang, John Paisley, and David M. Blei. “Online Variational Inference for the Hierarchical Dirichlet Process”. In: *Proceedings of the Fourteenth International Conference on Artificial Intelligence and Statistics*. Ed. by Geoffrey Gordon, David Dunson, and Miroslav Dudík. Vol. 15. Proceedings of Machine Learning Research. Fort Lauderdale, FL, USA: PMLR, Nov. 2011, pp. 752–760. URL: https://proceedings.mlr.press/v15/wang11a.html.

[WPB11b] Chong Wang, John Paisley, and David M. Blei. “Online Variational Inference for the Hierarchical Dirichlet Process”. In: *Proceedings of the Fourteenth International Conference on Artificial Intelligence and Statistics*. Ed. by Geoffrey Gordon, David Dunson, and Miroslav Dudík. Vol. 15. Proceedings of Machine Learning Research. Fort Lauderdale, FL, USA: PMLR, Nov. 2011, pp. 752–760. URL: https://proceedings.mlr.press/v15/wang11a.html.

[WM06] Xuerui Wang and Andrew McCallum. “Topics over Time: A Non-Markov Continuous Time Model of Topical Trends”. In: *Proceedings of the 12th ACM*

*SIGKDD International Conference on Knowledge Discovery and Data Mining*. KDD '06. Philadelphia, PA, USA: Association for Computing Machinery, 2006, pp. 424–433. ISBN: 1595933395. DOI: 10.1145/1150402.1150450. URL: https://doi.org/10.1145/1150402.1150450.

[WSW07] Xing Wei, Jimeng Sun, and Xuerui Wang. "Dynamic Mixture Models for Multiple Time-Series". In: *IJCAI 2007, Proceedings of the 20th International Joint Conference on Artificial Intelligence, Hyderabad, India, January 6-12, 2007*. Ed. by Manuela M. Veloso. 2007, pp. 2909–2914. URL: http://ijcai.org/Proceedings/07/Papers/468.pdf.

[Xio+18] Shufeng Xiong et al. "A short text sentiment-topic model for product reviews". In: *Neurocomputing* 297 (2018), pp. 94–102. ISSN: 0925-2312. DOI: https://doi.org/10.1016/j.neucom.2018.02.034. URL: https://www.sciencedirect.com/science/article/pii/S0925231218301693.

[Xun+17] Guangxu Xun et al. "A Correlated Topic Model Using Word Embeddings". In: *Proceedings of the Twenty-Sixth International Joint Conference on Artificial Intelligence, IJCAI-17*. 2017, pp. 4207–4213. DOI: 10.24963/ijcai.2017/588. URL: https://doi.org/10.24963/ijcai.2017/588.

[Yan+13] Xiaohui Yan et al. "A Biterm Topic Model for Short Texts". In: *Proceedings of the 22nd International Conference on World Wide Web*. WWW '13. Rio de Janeiro, Brazil: Association for Computing Machinery, 2013, pp. 1445–1456. DOI: 10.1145/2488388.2488514. URL: https://doi.org/10.1145/2488388.2488514.

[Yan+19] Weiwei Yang et al. "A Multilingual Topic Model for Learning Weighted Topic Links Across Corpora with Low Comparability". In: *Proceedings of the 2019 Conference on Empirical Methods in Natural Language Processing and the 9th International Joint Conference on Natural Language Processing (EMNLP-IJCNLP)*. Hong Kong, China: Association for Computational Linguistics, Nov. 2019, pp. 1243–1248.

[Yeu+17] Serena Yeung et al. "Tackling Over-pruning in Variational Autoencoders". In: *CoRR* abs/1706.03643 (2017). arXiv: 1706.03643. URL: http://arxiv.org/abs/1706.03643.

[ZSE17] Shengjia Zhao, Jiaming Song, and Stefano Ermon. "Towards Deeper Understanding of Variational Autoencoding Models". In: *CoRR* abs/1702.08658 (2017). arXiv: `1702.08658`. URL: `http://arxiv.org/abs/1702.08658`.

[ZBL06] Xiaojin Zhu, David Blei, and John Lafferty. "TagLDA: Bringing document structure knowledge into topic models". In: (Dec. 2006). URL: `shorturl.at/cfISY`.